\documentclass[preprint,12pt]{elsarticle}

\usepackage{amssymb}
\usepackage{amsmath}
\usepackage{lineno}

\usepackage{pifont}
\usepackage{multirow}
\newcommand{\cmark}{\ding{51}}  % checkmark
\newcommand{\xmark}{\ding{55}}  % crossmark

\journal{arXiv}
\begin{document}

\begin{frontmatter}



\title{Training deep physics-informed Kolmogorov--Arnold networks}
%\title{Deep Residual Gated Physics-Informed Kolmogorov-Arnold Networks}


\author[label1]{Spyros Rigas\corref{cor1}}
\ead{spyrigas@uoa.gr}
\author[label2]{Fotios Anagnostopoulos}
\ead{fotisanagn@uop.gr}
\author[label3]{Michalis Papachristou}
\ead{mixpap@phys.uoa.gr}
\author[label1]{Georgios Alexandridis}
\ead{gealexandri@uoa.gr}

\affiliation[label1]{organization={Department of Digital Industry Technologies, School of Science, National and Kapodistrian University of Athens},%Department and Organization
            %addressline={}, 
            %city={},
            postcode={34400}, 
            %state={},
            country={Greece}}

\affiliation[label2]{organization={Department of Informatics \& Telecommunications, University of the Peloponnese},%Department and Organization
            %addressline={}, 
            %city={Athens},
            postcode={22131}, 
            %state={},
            country={Greece}}
        
\affiliation[label3]{organization={Department of Physics, School of Science, National and Kapodistrian University of Athens},%Department and Organization
    %addressline={}, 
    %city={Athens},
    postcode={15784}, 
    %state={},
    country={Greece}}

\cortext[cor1]{Corresponding author}

%% Abstract
\begin{abstract}

Since their introduction, Kolmogorov--Arnold Networks (KANs) have been successfully applied across several domains, with physics-informed machine learning (PIML) emerging as one of the areas where they have thrived. In the PIML setting, Chebyshev-based physics-informed KANs (cPIKANs) have become the standard due to their computational efficiency. However, like their multilayer perceptron-based counterparts, cPIKANs face significant challenges when scaled to depth, leading to training instabilities that limit their applicability to several PDE problems. To address this, we propose a basis-agnostic, Glorot-like initialization scheme that preserves activation variance and yields substantial improvements in stability and accuracy over the default initialization of cPIKANs. Inspired by the PirateNet architecture, we further introduce Residual-Gated Adaptive KANs (RGA KANs), designed to mitigate divergence in deep cPIKANs where initialization alone is not sufficient. Through empirical tests and information bottleneck analysis, we show that RGA KANs successfully traverse all training phases, unlike baseline cPIKANs, which stagnate in the diffusion phase in specific PDE settings. Evaluations on nine standard forward PDE benchmarks under a fixed training pipeline with adaptive components demonstrate that RGA KANs consistently outperform parameter-matched cPIKANs and PirateNets -- often by several orders of magnitude -- while remaining stable in settings where the others diverge.

\end{abstract}

%%%\begin{graphicalabstract}
%%%\end{graphicalabstract}

%%%\begin{highlights}
%%%\item Fading of a script alone does not foster domain-general strategy knowledge
%%%\item Performance of the strategy declines during the fading of a script
%%%\item Monitoring by a peer keeps performance of the strategy up during script fading
%%%\item Performance of a strategy after fading fosters domain-general strategy knowledge
%%%\item Fading and monitoring by a peer combined foster domain-general strategy knowledge
%%%\end{highlights}

%% Keywords

\end{frontmatter}

%\linenumbers

\section{Introduction} \label{sec1}

The widespread adoption and integration of machine learning into computational science has profoundly influenced the way complex physical phenomena are modeled and analyzed. One of the most striking advances is the Physics‑Informed Machine Learning (PIML) framework \cite{NaturePIML, ReviewPIML}, which offers a compelling alternative to traditional discretization‑based solvers for both forward and inverse problems involving partial differential equations (PDEs). Within the PIML framework, the governing equations, alongside boundary and initial conditions, plus any observational data, are embedded into a differentiable loss function, while a neural network parametrizes the unknown solution field. Leveraging automatic differentiation \cite{AD} to evaluate differential operators exactly, PIML eliminates the need for mesh generation and yields continuous, high‑fidelity predictions with reduced computational cost. As a result, it has found success across a broad spectrum of scientific and engineering disciplines, from fluid mechanics \cite{fluid1, fluid2, fluid3, fluid4} and materials science \cite{mat1, mat2} to medicine \cite{med1, med2, med3} and chemistry \cite{chem1, chem2}.

While a variety of neural architectures have been explored within the PIML paradigm, including convolutional neural networks (CNNs) \cite{archcnn}, generative adversarial networks (GANs) \cite{archgan}, and long short-term memory (LSTM) networks \cite{archlstm}, the fully connected multilayer perceptron (MLP) is the predominant backbone. When an MLP parametrizes the solution field, the variant is conventionally termed a Physics-Informed Neural Network (PINN), which is also the original formulation of PIML \cite{Raissi}. Despite their widespread use, PINNs exhibit several well‑documented shortcomings, including spectral bias toward low‑frequency modes \cite{SpectralBias}, restricted interpretability and limited scalability with depth, among other challenges in their training dynamics. To address these issues, numerous mitigation strategies have been proposed, ranging from architectural modifications \cite{Wang1, Wang2, RWF, PirateNets}, to adaptive training techniques \cite{WangNTK, Causal, RBA}. A complementary approach is to forgo the MLP backbone altogether in favor of alternative architectures that mitigate several of these issues by design.

One such emerging alternative is the Kolmogorov--Arnold Network (KAN) \cite{KAN1}. Whereas the expressivity of MLPs is supported by the universal approximation theorem \cite{UAT}, KANs are grounded in the Kolmogorov--Arnold representation theorem \cite{KART}. In practice, a KAN layer replaces fixed nonlinear activations with a learnable linear combination of basis functions; the original implementation employs B-splines, but other basis functions such as radial basis functions \cite{rbfKAN}, Chebyshev polynomials \cite{chebyKAN} and Rectified Linear Unit (ReLU)-based functions \cite{reluKAN} have also been explored. This design offers several benefits, most notably in terms of enhanced interpretability \cite{KAN2} and robustness against spectral bias \cite{KAN3}. These have motivated the development of Physics-Informed Kolmogorov--Arnold Networks (PIKANs) \cite{FAIR, KINNs}, where the MLP backbone is substituted by a KAN in the PINN framework. Initial studies have demonstrated that PIKANs can attain higher accuracy on benchmark PDEs, or comparable accuracy with considerably smaller network architectures than their MLP-based counterparts \cite{KAN1, FAIR, Rigas}. Consequently, they have already seen practical deployment in a variety of scientific and engineering contexts \cite{pikanapp1, pikanapp2, pikanapp3}.

Despite their promising early results, PIKANs present their own challenges. Computational overhead is the most immediate: evaluating and differentiating the B‑spline basis in the original KAN formulation quickly becomes a bottleneck, leading most physics-informed implementations to adopt the more efficient Chebyshev variant (cPIKAN) \cite{FAIR}. Scalability is another concern; empirical studies report training instabilities as the number of network parameters increases beyond a point \cite{KANlimit, KKANs}, limiting applicability in deep learning regimes. Similar issues are observed in deep PINNs, although the PirateNet architecture appears to mitigate them \cite{PirateNets}. Methodological gaps also persist. Weight initialization schemes are still largely ad hoc: each basis family (B‑spline, radial‑basis, Chebyshev, etc.) provides its own default, yet no analogue to the well-studied Glorot initialization for PINNs \cite{Glorot} has been empirically or theoretically analyzed. Systematic experimentation with initialization schemes is at a nascent stage and has so far concentrated exclusively on B-spline-based KANs \cite{MOSS}. The picture is similar for adaptive training strategies. While several PINN-oriented techniques have been ported to PIKANs \cite{Rigas, AAH1, AAH2}, a unified training pipeline, comparable to the one codified for PINNs \cite{ExpertsGuide}, has yet to be established.

Motivated by these gaps, we concentrate on cPIKANs, which are a fitting choice for physics-informed applications in terms of their computational efficiency and accuracy. We first observe that the reported depth-related instabilities may be closely tied to weight initialization: an initialization that preserves activation variance can prevent vanishing or exploding gradients, just as the Glorot scheme does in MLPs. Accordingly, we derive a ``Glorot‑like'' initialization for KANs that makes no assumptions about the specific basis and is therefore applicable to any KAN variant. On a series of function-fitting and PDE benchmarks, we show that this initialization improves optimization stability and yields significantly more accurate solutions than the default initialization of Chebyshev-based KANs and cPIKANs.

Building on this foundation, we address the depth-scaling issue. Because each KAN layer carries multiple learnable basis coefficients, a KAN layer of the same width as an MLP layer is substantially more parameter-heavy; real-world tasks that need larger capacity must therefore rely on greater depth, which in turn demands stable training. To this end, we introduce a Residual-Gated Adaptive KAN (RGA KAN) architecture. We analyze its training dynamics through the lens of the Information Bottleneck (IB) framework \cite{IBTheory} and empirically demonstrate that RGA KANs remain stable and train effectively at depths where baseline cPIKANs diverge.

Finally, using our proposed initialization scheme and the RGA KAN architecture, we conduct extensive experiments on a suite of forward PDE problems. RGA KANs are compared with parameter-matched PirateNets and baseline cPIKANs under a standardized training pipeline that incorporates adaptive techniques drawn from PINN best practices. Ablation studies are also performed to quantify the influence of each adaptive component of the training pipeline, establishing a first set of depth-scalable benchmarks for cPIKANs and demonstrating that our contributions jointly close much of the performance and stability gap identified in earlier work.

In summary, the key contributions of this work are the following:

\begin{itemize}
    \item We derive a basis-agnostic, Glorot-like initialization scheme that improves the accuracy of the studied KANs on both function-fitting and PDE-solving tasks.

    \item We introduce RGA KANs, designed to address the degradation in performance observed during the training of deep cPIKANs. We further analyze their training dynamics through the lens of IB theory.

    \item We benchmark RGA KANs against baseline cPIKANs and PirateNets on a suite of forward PDE problems, using identical adaptive training techniques across all models.

    \item Through ablation studies, we quantify the individual contributions of each adaptive technique to the overall performance of our proposed architecture.
\end{itemize}

The remainder of this paper is structured as follows. Section \ref{sec2} reviews the theoretical foundations of our study, covering the PIML framework and KANs. Section \ref{sec3} presents the proposed basis-agnostic Glorot-like initialization and demonstrates its clear advantage over the default cPIKAN initialization through small-scale function-fitting and PDE benchmarks. Section \ref{sec4} addresses the depth-scaling limitations of cPIKANs by introducing the RGA KAN architecture; we analyze its training dynamics via the IB theory and show that it remains stable where cPIKANs diverge. Section \ref{sec5} delivers a comprehensive empirical comparison among RGA KANs, baseline cPIKANs, and PirateNets on a suite of forward PDE problems, supplemented by ablation studies that isolate the contribution of each adaptive training component. Finally, Section \ref{sec6} summarizes our principal findings and outlines promising directions for future research.

\section{Theoretical Background} \label{sec2}

\subsection{Problem Formulation} \label{sec2.1}

Without loss of generality, we consider PDEs of the form 

\begin{equation}
    \mathcal{F}\left[u\left(t,\mathbf{x}\right)\right] = f\left(t,\mathbf{x}\right), ~ t \in \left[0,T\right],~ \mathbf{x} \in \Omega, \label{eq1}
\end{equation}

\noindent defined over a bounded $d$-dimensional spatial domain $\Omega \subset \mathbb{R}^d$ with boundary $\partial \Omega$ and a temporal domain $\left[0,T\right]$, and subject to initial and boundary conditions

\begin{align}
    u\left(0,\mathbf{x}\right) &= g\left(\mathbf{x}\right),~ \mathbf{x} \in \Omega, \label{eq2} \\
    \mathcal{R}_\text{bc}\left[u\left(t,\mathbf{x}\right)\right] &= 0,~ t \in \left[0,T\right],~ \mathbf{x} \in \partial \Omega. \label{eq3}
\end{align}

\noindent In the above expressions, $\mathcal{F}$ corresponds to an abstract differential operator, $\mathcal{R}_\text{bc}$ is a boundary operator that imposes Dirichlet, Neumann, Robin, or periodic boundary conditions, while $u\left(t,\mathbf{x}\right)$ represents the solution of the PDE. Additionally, $f\left(t,\mathbf{x}\right)$ and $g\left(\mathbf{x}\right)$ are known functions corresponding to the PDE's source term and initial condition, respectively.

The core idea behind PIML is to approximate the unknown solution by a neural network $u\left(t,\mathbf{x};\boldsymbol{\theta}\right)$, where $\boldsymbol{\theta}$ denotes all trainable parameters of the network. To this end, we define the PDE's residuals as

\begin{equation}
    \mathcal{R}_\text{pde}\left[u\left(t,\mathbf{x};\boldsymbol{\theta}\right)\right] = \mathcal{F}\left[u\left(t,\mathbf{x};\boldsymbol{\theta}\right)\right] - f\left(t,\mathbf{x}\right), \label{eq4}
\end{equation}

\noindent and the initial condition's residuals as

\begin{equation}
    \mathcal{R}_\text{ic}\left[u\left(t,\mathbf{x};\boldsymbol{\theta}\right)\right] = u\left(0,\mathbf{x};\boldsymbol{\theta}\right) - g\left(\mathbf{x}\right). \label{eq5}
\end{equation}

\noindent Then, the neural network is trained by minimizing the composite loss function

\begin{equation}
    \mathcal{L}\left(\boldsymbol{\theta}\right) = \lambda_\text{pde} \mathcal{L}_\text{pde}\left(\boldsymbol{\theta}\right) + \lambda_\text{ic} \mathcal{L}_\text{ic}\left(\boldsymbol{\theta}\right) + \lambda_\text{bc} \mathcal{L}_\text{bc}\left(\boldsymbol{\theta}\right), \label{eq6}
\end{equation}

\noindent where $\lambda_\text{pde}$, $\lambda_\text{ic}$, $\lambda_\text{bc}$ are hyperparameters that allow the assignment of different weights to each individual term of the composite loss function and

\begin{align}
    \mathcal{L}_\text{pde}\left(\boldsymbol{\theta}\right) &= \frac{1}{N_\text{pde}}\sum_{i=1}^{N_\text{pde}}\left \lVert \mathcal{R}_\text{pde}\left[u\left(t_\text{pde}^i,\mathbf{x}_\text{pde}^i;\boldsymbol{\theta}\right)\right] \right \rVert_2^2, \label{eq7} \\
    \mathcal{L}_\text{ic}\left(\boldsymbol{\theta}\right) &= \frac{1}{N_\text{ic}}\sum_{i=1}^{N_\text{ic}}\left \lVert \mathcal{R}_\text{ic}\left[u\left(0,\mathbf{x}_\text{ic}^i;\boldsymbol{\theta}\right)\right] \right \rVert_2^2, \label{eq8} \\
    \mathcal{L}_\text{bc}\left(\boldsymbol{\theta}\right) &= \frac{1}{N_\text{bc}}\sum_{i=1}^{N_\text{bc}}\left \lVert \mathcal{R}_\text{bc}\left[u\left(t_\text{bc}^i,\mathbf{x}_\text{bc}^i;\boldsymbol{\theta}\right)\right] \right \rVert_2^2, \label{eq9}
\end{align}

\noindent where $\lVert \cdot \rVert_2$ denotes the $L^2$ norm and $\left\{\left(t_\xi^i,\mathbf{x}_\xi^i\right)\right\}_{i=1}^{N_\xi}$, with $\xi$ being either ``pde'', ``ic'' or ``bc'', correspond to collocation points used to calculate the PDE's, initial condition's and boundary conditions' residuals, respectively.

%We remark that the global $\lambda$-weights can either be defined based on domain knowledge (e.g., \cite{FAIR}), or adjusted dynamically during the network's training \cite{WangNTK, Causal, KKANs, AAHWeights}. Moreover, we note that the set of collocation points used to calculate the PDE's residuals can be sampled once from a fixed grid or adaptively re-sampled throughout training \cite{COLLOCS1, RAD, COLLOCS2, COLLOCS3}. In this work, we incorporate a specific suite of such adaptive strategies, namely Residual-Based Attention (RBA), Residual-Based Adaptive Distribution (RAD), causal training and learning-rate annealing (LRA). Details on these methods can be found in \ref{app:adaptive}.

While the standard formulation described above relies on fixed weights and static sets of collocation points, many practical applications require adaptive training strategies. The framework can thus be extended by introducing dynamic global or local weights, as well as adaptive resampling schemes. In this work, we specifically incorporate a suite of such adaptive strategies, namely residual-based attention (RBA), residual-based adaptive distribution (RAD), causal training and learning-rate annealing (LRA). A detailed description of these methods is provided in \ref{app:adaptive}.

\subsection{Kolmogorov--Arnold Networks}  \label{sec2.2}

Until recently, the vast majority of the neural network architectures which were chosen to approximate the PDE's solution utilized MLPs as their backbone. In an MLP, the output of the $l$-th layer is recursively defined in terms of the output of the $(l-1)$-th layer as follows:

\begin{equation}
    u_j^{(l)}\left(t,\mathbf{x};\boldsymbol{\theta}\right) = \sigma\left(\sum_{i=1}^{d_{l-1}} w_{ji}^{(l)}\, u_i^{(l-1)}\left(t,\mathbf{x};\boldsymbol{\theta}\right) + b_j^{(l)}\right), \label{eq10}
\end{equation}

\noindent where $w_{ji}^{(l)}, b_j^{(l)}$ represent the weights and biases of the $l$-th layer, $d_{l-1}$ is the output dimension of the $(l-1)$-th layer and $\sigma$ is a non-linear activation function -- typically the hyperbolic tangent for PINNs. For the recursion to be consistent, the first layer is assumed to perform an identity operation, i.e.,

\begin{equation}
    u_j^{(0)} = %\left(t,\mathbf{x};\boldsymbol{\theta}\right) =
    \begin{cases}
        t, & j = 0,\\[6pt]
        x_j, & j \in \{1,\dots,d\}.
    \end{cases} \label{eq11}
\end{equation}

\noindent For an MLP with an input layer of dimension $d_\text{I}$, $L$ hidden layers each of dimension $d_\text{H}$ and an output layer of dimension $d_\text{O}$, the cardinality of the set of the network's parameters is given by

\begin{equation}
    \left|\boldsymbol{\theta}\right| = d_\text{H}\left[d_\text{I} + \left(L-1\right)d_\text{H} + L + d_\text{O}\right] + d_\text{O} = \mathcal{O}\left(d_\text{H}^2 L\right). \label{eq12}
\end{equation}

Inspired by the Kolmogorov--Arnold representation theorem \cite{KART}, the authors of \cite{KAN1} introduced KANs, a novel class of neural networks that have since been adopted as an alternative to MLPs. The formulation corresponding to Eq. \eqref{eq10} for the original implementation of KANs, known as ``vanilla'' KANs, is given by

\begin{equation}
    \begin{aligned}
        u_j^{(l)}\left(t,\mathbf{x};\boldsymbol{\theta}\right) = \sum_{i=1}^{d_{l-1}} 
        \Bigg( r_{ji}^{(l)} &\, R\left[u_i^{(l-1)}\left(t,\mathbf{x};\boldsymbol{\theta}\right)\right] \\
        & + c_{ji}^{(l)} \sum_{m=1}^{D} w_{jim}^{(l)} \, B_m^{(l)}
        \left[u_i^{(l-1)}\left(t,\mathbf{x};\boldsymbol{\theta}\right)\right] \Bigg),
    \end{aligned}
    \label{eq13}
\end{equation}

\noindent where $r_{ji}^{(l)}$, $c_{ji}^{(l)}$ and $w_{jim}^{(l)}$ are the trainable parameters of the $l$-th layer,

\begin{equation}
    R(x) = \frac{x}{1 + \exp\left(-x\right)} \label{eq14}
\end{equation}

\noindent is a residual function and $B_m^{(l)}\left(\cdot\right)$ are univariate spline basis functions. The superscript $(l)$ reflects the inherent dependency of the spline basis functions on a layer-specific grid, while the subscript $m$ runs from 1 to $D = G+k$, where $G$ is the number of grid intervals and $k$ is the order of the basis functions. Comparing Eq. \eqref{eq13} to Eq. \eqref{eq10}, a fundamental distinction arises between MLPs and KANs in terms of their functional representation: in MLPs, nonlinearity is introduced through fixed activation functions,  while only the linear transformations between layers are trainable; in contrast, KANs replace these static activation functions with learnable ones, an aspect that highlights their potential to create more expressive architectures \cite{KAN1, KAN3}.

While vanilla KANs have demonstrated promising results in solving forward PDE problems \cite{KINNs, Rigas}, their training is burdened by the expensive computation of spline basis functions. Additionally, their dependency on a grid, although beneficial in certain applications \cite{KAN1, KAN2, Rigas}, becomes redundant when the grid remains fixed throughout training. To mitigate these inefficiencies, some approaches have introduced optimization strategies \cite{jaxKAN, AAH3}, while others have explored alternative, more computationally efficient basis functions \cite{rbfKAN, chebyKAN, ActNet}. In this work, we adopt the latter approach and use Chebyshev-based KANs, due to their proven success in PIML \cite{FAIR, pikanapp3, KKANs, BPIKANs}. We therefore modify the expression of Eq. \eqref{eq13} to

\begin{equation}
    u_j^{(l)}\left(t,\mathbf{x};\boldsymbol{\theta}\right) = \sum_{i=1}^{d_{l-1}} 
        \sum_{m=1}^{D} w_{jim}^{(l)} \, B_m
        \left[u_i^{(l-1)}\left(t,\mathbf{x};\boldsymbol{\theta}\right)\right] + b_j^{(l)} ,
    \label{eq15}
\end{equation}

\noindent where the residual term has now been removed, the $c_{ji}$ weights have been absorbed by $w_{jim}$, an additional bias term, $b_j$, has been introduced and $B_m$ are now grid-independent basis functions with $m = 1, \dots, D$. These basis functions are given by

\begin{equation}
    B_m\left(x\right) = T_m\left(\tanh\left(x\right)\right), \label{eq16}
\end{equation}

\noindent where $T_m\left(\cdot\right)$ are Chebyshev polynomials of the first kind. The hyperbolic tangent in the argument of $T_m$ maps the unbounded input $x$ to the canonical range $[-1,1]$, which is the orthogonal domain of the Chebyshev polynomials. For the purposes of the present work, the Chebyshev polynomials are explicitly defined as functions up to order $D$ to maximize computational efficiency.

Assuming a KAN architecture equivalent to the previously discussed MLP, with input dimension $d_\text{I}$, $L$ hidden layers of dimension $d_\text{H}$ and output dimension $d_\text{O}$, the total number of trainable parameters is given by

\begin{equation}
    \left|\boldsymbol{\theta}\right| = d_\text{H}\left[d_\text{I} D + D\left(L-1\right)d_\text{H} + L + d_\text{O}D\right] + d_\text{O} = \mathcal{O}\left(d_\text{H}^2 DL\right). \label{eq17}
\end{equation}

\noindent Comparing Eq. \eqref{eq17} to Eq. \eqref{eq12}, a fundamental limitation of KANs becomes evident: for architectures with the same depth and width, a KAN contains approximately $D$ times more parameters than its MLP counterpart. Consequently, the primary challenge in training KANs is to achieve equal or superior performance to MLPs while maintaining a comparable total number of parameters.

\section{KAN Initialization} \label{sec3}

Initialization plays a critical role in the training dynamics of deep neural networks. In the case of MLPs, extensive theoretical and empirical work has led to well-established initialization schemes tailored to different architectures and activation functions. Within the PINN framework, architectures are most often initialized using the Glorot scheme \cite{ExpertsGuide}, which aims to preserve the variance of activations and gradients across layers to avoid vanishing or exploding signals, thereby enabling stable deep learning.

For KANs -- and, by extension, PIKANs -- the choice of weight initialization remains largely ad hoc, with each variant in the literature adopting its own heuristic procedure. In contrast to the extensive body of work on MLP initialization, systematic studies for KANs are scarce and have thus far been limited to the vanilla formulation with B-spline bases \cite{MOSS}, where power-law and LeCun-like schemes were proposed. In the specific case of Chebyshev-based KANs, the only initialization strategy reported in the literature \cite{chebyKAN,KKANs} involves drawing the basis coefficients $w_{jim}$ of Eq. \eqref{eq15} from a truncated normal distribution with zero mean and variance defined by:

\begin{equation}
	\sigma^2_{\text{default}} = \frac{1}{d_\text{I}(D+1)}, \label{eq_default_init}
\end{equation}

\noindent where $D$ is the number of basis functions and $d_\text{I}$ is the layer's input dimension. Building on these early efforts, we develop a Glorot-like initialization derived from a variance-preservation analysis and formulated to be agnostic to the choice of basis functions. In the present work, we assess its performance in Chebyshev-based KANs and cPIKANs, demonstrating that it offers improved training stability and accuracy over the initialization scheme of Eq. \eqref{eq_default_init}.

\subsection{Proposed Scheme} \label{sec3.1}

Consider a single KAN layer with input $\mathbf{x} \in \mathbb{R}^{d_\text{I}}$ and output $\mathbf{y} \in \mathbb{R}^{d_\text{O}}$. Based on Eq. \eqref{eq15}, the $j$-th output component is given by

\begin{equation}
	y_j = \sum_{i=1}^{d_\text{I}} \sum_{m=1}^{D} w_{jim} \, B_m\left(x_i\right) + b_j ,
	\label{eq26}
\end{equation}

\noindent where $B_m(\cdot)$ denotes the $m$-th basis function. Throughout this derivation we set all biases $b_j$ to zero at initialization and focus on the initialization of the coefficients $w_{jim}$.

In the MLP setting, weight initialization is typically modeled by assuming i.i.d. Gaussian entries drawn from $\mathcal{N}\left(0,\,\sigma^2\right)$, and selecting $\sigma$ according to a chosen criterion. Here, the presence of an additional basis index, $m$, motivates the more general assumption

\begin{equation}
	w_{jim} \ \sim\ \mathcal{N}\!\left(0,\,\sigma_m^2\right),
	\label{eq27}
\end{equation}

\noindent where $\sigma_m$ is a basis-term–dependent standard deviation to be determined. For the inputs $x_i$, we assume i.i.d. samples with zero mean and unit variance, consistent with common deep learning practices.

A central principle in initialization design, which is also the core idea behind the LeCun initialization \cite{lecun}, is to preserve the variance of the signal during the forward pass. Applying this condition to Eq. \eqref{eq26} gives

\begin{equation}
	1 \ =\ d_\text{I} \sum_{m=1}^{D} \sigma_m^2 \, \mu_m^{(0)} ,
	\label{eq28}
\end{equation}

\noindent where

\begin{equation}
	\mu_m^{(0)} \ := \ \mathbb{E}\left[ B_m\!\left(x\right)^2 \right],
	\label{eq29}
\end{equation}

\noindent and the expectation is taken with respect to the distribution of $x$.

In his original work, Glorot \cite{Glorot} further required variance preservation during the backward pass, so that gradients propagate across layers without amplification or attenuation. Applying the same reasoning to the gradient signal through Eq. \eqref{eq26} leads to

\begin{equation}
	1 \ =\ d_\text{O} \sum_{m=1}^{D} \sigma_m^2 \, \mu_m^{(1)} ,
	\label{eq30}
\end{equation}

\noindent where

\begin{equation}
	\mu_m^{(1)} \ :=\ \mathbb{E}\!\left[ B_m^\prime\!\left(x\right)^2 \right],
	\label{eq31}
\end{equation}

\noindent with $B_m^\prime(\cdot)$ denoting the derivative of the $m$-th basis function. Eqs. \eqref{eq28} and \eqref{eq30} thus impose the forward- and backward-variance constraints. Balancing them in the spirit of Glorot leads to

\begin{equation}
	\sigma_m^2 \ =\ \frac{1}{D} \cdot \frac{2}{\,d_\text{I}\,\mu_m^{(0)} \ +\ d_\text{O}\,\mu_m^{(1)}\,} .
	\label{eq32}
\end{equation}

\noindent This expression defines a basis-agnostic Glorot-like initialization rule: $\mu_m^{(0)}$ and $\mu_m^{(1)}$ capture the effect of the chosen basis functions on variance, while $d_\text{I}$ and $d_\text{O}$ play the same role as in the original Glorot scheme. The additional factor $1/D$ accounts for the contribution of the $D$ basis terms associated with each input dimension. In the special case $\mu_m^{(0)} = \mu_m^{(1)} = 1$ and $D=1$, corresponding to the MLP setting, which can be interpreted as using a single basis function, Eq. \eqref{eq32} reduces exactly to the standard Glorot initialization. Detailed derivations of Eqs. \eqref{eq28} and \eqref{eq30} are provided in \ref{app:derivs1}.

For practical applications, it is often convenient to introduce a multiplicative gain factor, following common practice in initialization utilities provided by deep learning frameworks such as \texttt{PyTorch} \cite{pytorch}. This gain term allows for empirical correction in cases where the input distribution deviates from the unit-variance assumption. Incorporating this factor, the final initialization rule becomes

\begin{equation}
	\sigma_m \ =\ \text{gain} \, \sqrt{\frac{1}{D} \cdot \frac{2}{\,d_\text{I}\,\mu_m^{(0)} \ +\ d_\text{O}\,\mu_m^{(1)}\,}} ,
	\label{eq33}
\end{equation}

\noindent which recovers the standard Glorot initialization when $\mu_m^{(0)} = \mu_m^{(1)} = 1$, $D=1$ and $\text{gain}=1$.

\subsection{Small-Scale Benchmarks} \label{sec3.2}

To evaluate the effectiveness of the proposed initialization scheme, we compare it against the standard practice for Chebyshev-based KANs. To this end, we conduct experiments on two small-scale benchmarks: (i) function fitting, and (ii) forward PDE problems. For the remainder of this work, we refer to the standard initialization strategy defined in Eq. \eqref{eq_default_init} as the ``default'' initialization.

\subsubsection{Function Fitting} \label{sec3.2.1}

We first assess initialization performance on five function-fitting tasks of increasing dimensionality: (i) a one-dimensional oscillatory function, $f_1(x)$, (ii) the two-dimensional product function, $f_2(x_1,x_2)$, (iii) a more challenging two-dimensional function borrowed from \cite{MOSS}, $f_3(x_1,x_2)$, (iv) the three-dimensional Hartmann function, $f_4(x_1,x_2,x_3)$, which is a common benchmark in function approximation, and (v) the five-dimensional Sobol g-function, $f_5(x_1,\dots,x_5)$, widely used in global sensitivity analysis. The analytic definitions of all functions are provided in \ref{app:ff}.

\begin{figure}[t!]
	\centering
	\includegraphics[width=\linewidth]{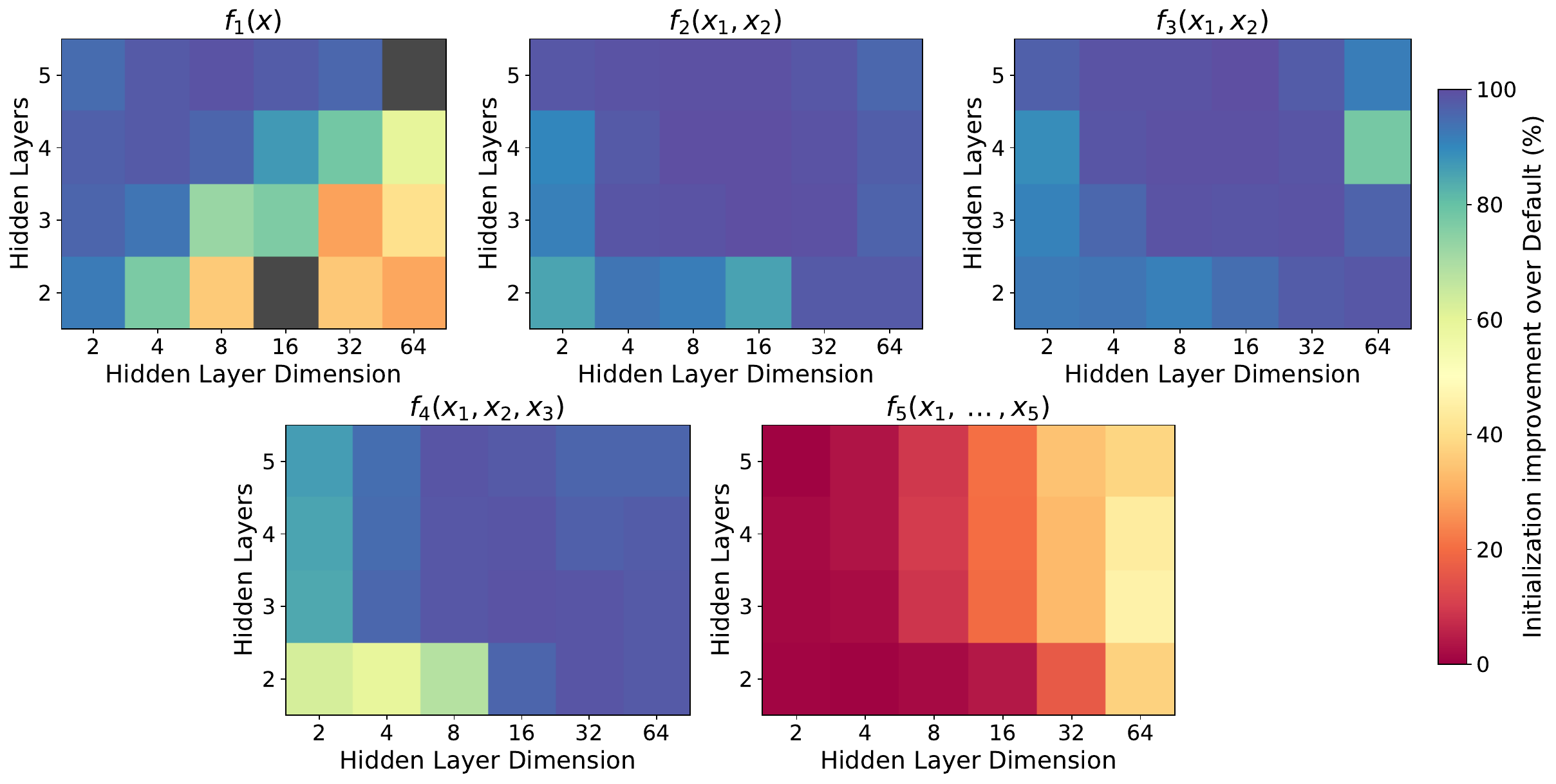}
	\caption{Relative comparison of proposed and default initialization across the five benchmark functions. Each subplot corresponds to one function, with the color scale indicating the percentage improvement of the proposed initialization over the default in terms of the final $L^2$ error. Black cells denote configurations where the default initialization attains lower error.}
	\label{fig1}
\end{figure}

We train Chebyshev-based KANs with polynomial order $D=8$ using both the default and the proposed Glorot-like initialization schemes. We conduct a sweep over architectures of varying width (hidden layer dimensions of 2, 4, 8, 16, 32, 64) and depth (2 to 5 hidden layers) and evaluate performance in terms of the relative $L^2$ error with respect to the reference solution. To ensure statistical significance, each configuration is repeated with five random seeds. Full implementation details regarding data generation and optimization settings are provided in \ref{app:impl1}. The results of these experiments are summarized in Figure \ref{fig1}. Each heatmap corresponds to one benchmark function, with hidden layer dimension on the horizontal axis and number of hidden layers on the vertical axis. The color scale quantifies the relative improvement achieved by the proposed initialization over the default scheme, computed as

\begin{equation*}
	\frac{\mathcal{E}_\text{default} - \mathcal{E}_\text{proposed}}{\mathcal{E}_\text{default}} \times 100\% ,
\end{equation*}

\noindent where $\mathcal{E}$ denotes the relative $L^2$ error. Values are clipped to the range $0$--$100\%$, with black cells indicating cases where the default initialization outperforms the proposed method.

Evidently, the impact of the proposed initialization is substantial in most benchmarks. For the two-dimensional functions ($f_2$ and $f_3$) and the three-dimensional Hartmann function ($f_4$), improvements approach 100\% across nearly all architectures, indicating that the proposed initialization reduces the final relative $L^2$ error by up to several orders of magnitude compared to the default scheme. The one-dimensional oscillatory function ($f_1$) also shows clear gains in the majority of cases, with only two isolated configurations where the default (marginally) outperforms. The five-dimensional Sobol g-function ($f_5$) exhibits improvements as well, though they are less pronounced, typically in the range of 5--50\%; in this setting, both initializations yield comparable overall accuracy, and therefore the difference in initialization impact is less striking.

\begin{figure}[t!]
	\centering
	\includegraphics[width=\linewidth]{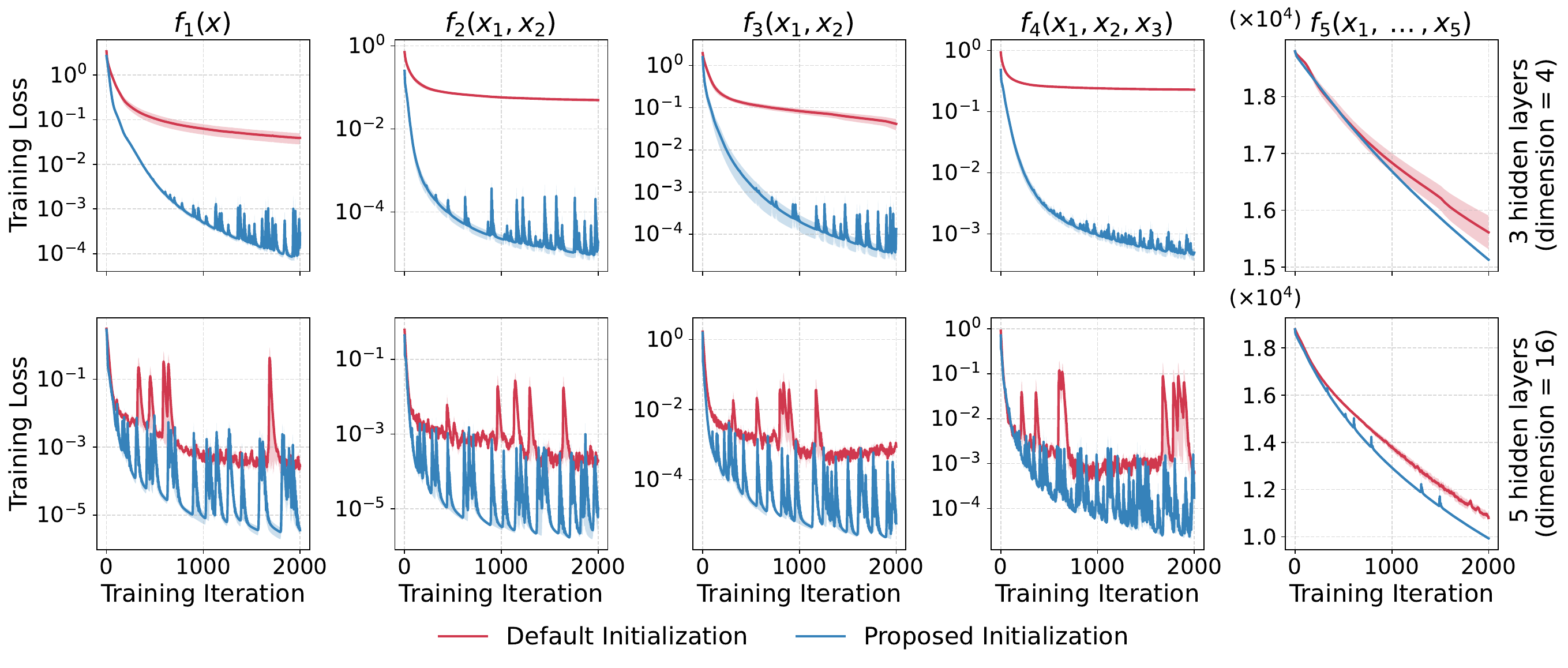}
	\caption{Loss throughout training for two representative architectures (top row: three 4-dimensional hidden layers; bottom row: five 16-dimensional hidden layers) across the five benchmark functions. Each subplot shows the mean training loss over five independent runs (solid lines) together with the SEM (shaded area). The final column, corresponding to $f_5$, is shown without logarithmic scaling on the $y$-axis, since the loss did not exhibit significant improvement during training.}
	\label{fig2}
\end{figure}

Apart from the final error metrics, it is also informative to examine the training loss evolution, in order to assess whether the proposed initialization leads to a more effective optimization of the loss function. To this end, we consider two representative architectures: a smaller network with width $4$ and depth $3$, and a larger network with width $16$ and depth $5$. Figure \ref{fig2} depicts the training loss curves for each of the five benchmark functions under both initialization schemes. Solid lines indicate the mean loss across five independent runs, while the shaded regions correspond to the standard error of the mean (SEM). The proposed initialization consistently accelerates convergence and achieves substantially lower training losses. For the oscillatory function ($f_1$) and the two-dimensional cases ($f_2$ and $f_3$), the difference spans more than two orders of magnitude, with low variability across seeds. Similar improvements are observed for the Hartmann function ($f_4$), where the default initialization stalls at higher loss values compared to the proposed scheme. For the Sobol $g$-function ($f_5$), both schemes exhibit nearly identical behavior, in line with the earlier observation that this benchmark is less sensitive to initialization. Importantly, these trends are visible in both architectures, demonstrating robustness across different model scales. To provide deeper insight into the stability improvements of the proposed initialization, we also present a Neural Tangent Kernel (NTK) analysis for these architectures in \ref{app:ntk1}.

\subsubsection{Forward PDE Problems} \label{sec3.2.2}

\begin{figure}[b!]
	\centering
	\includegraphics[width=\linewidth]{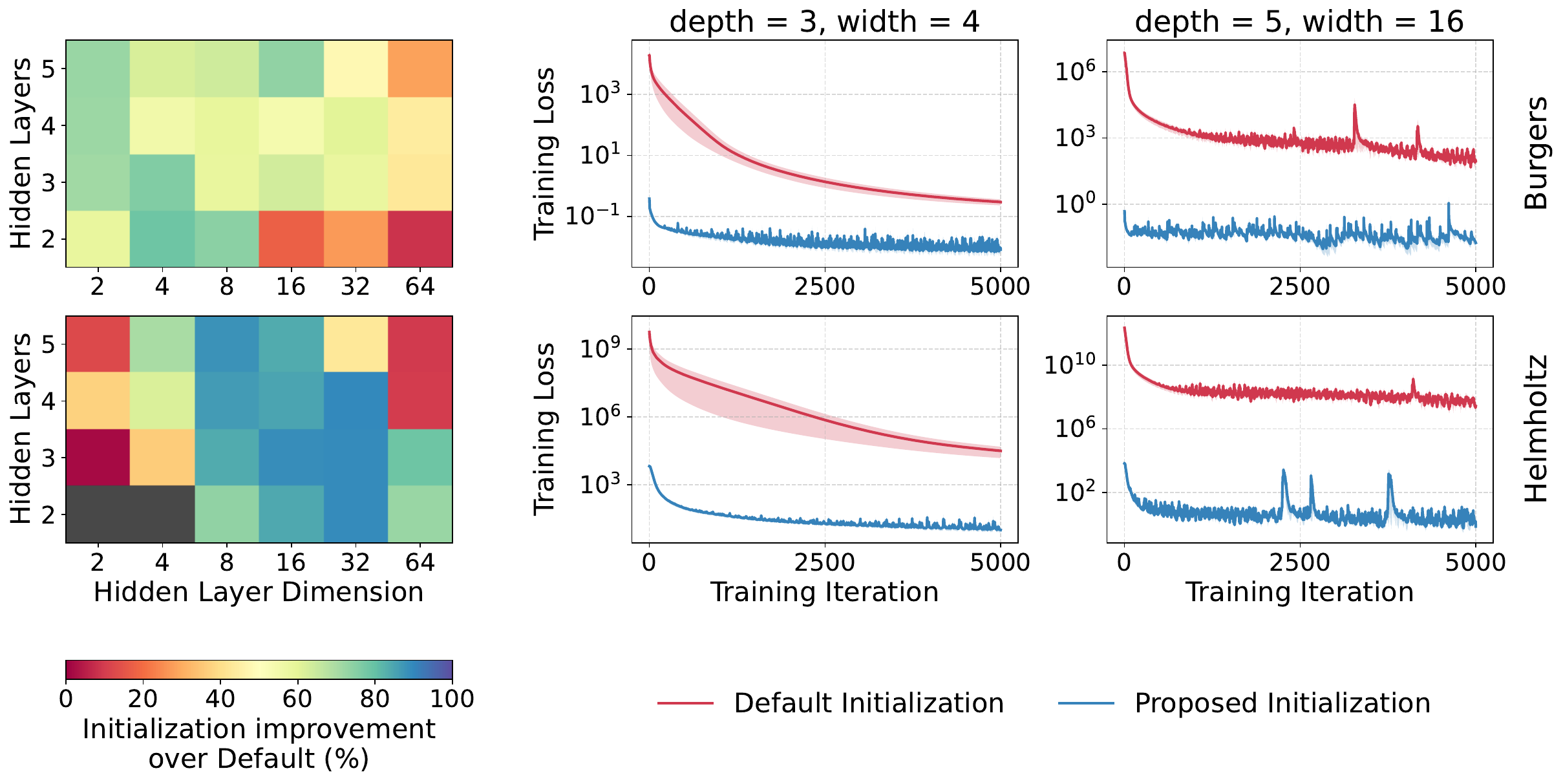}
	\caption{Comparison of default and proposed initialization schemes on Burgers' (top row) and Helmholtz (bottom row) equations. Left column: heatmaps of relative improvement in final $L^2$ error. Middle/Right column: training-loss curves per initialization scheme for a representative architecture of depth = 3/5 and width = 4/16, respectively. Shaded regions denote the SEM across five runs.}
	\label{fig3}
\end{figure}

We next assess the proposed initialization on forward PDE benchmarks using cPIKANs. Specifically, we consider Burgers' equation as well as the Helmholtz equation with $a_1=1$, $a_2=4$ (see \ref{app:pdes} for details), following the PIML framework introduced in Section \ref{sec2}, without incorporating any additional adaptive techniques. We use Chebyshev-based KANs of order $D=8$ and conduct a sweep over the same architectural configurations (widths and depths) as in the function-fitting experiments, repeating each experiment with five random seeds. Full details on collocation point distribution and optimization settings are provided in \ref{app:impl1}. The results of these experiments are summarized in Figure \ref{fig3}, which combines heatmaps of final relative $L^2$ error improvements with representative training-loss curves.

As in the function-fitting benchmarks, the proposed initialization consistently outperforms the default scheme in terms of final error, with the exception of only two small architectures where the default initialization yields a marginal advantage. The overall gains are somewhat less pronounced than in Section \ref{sec3.2.1}, which can be attributed to the relatively low number of training iterations and the absence of adaptive training techniques. Nevertheless, the training-loss curves reveal a striking effect of initialization. In particular, for the Helmholtz equation, where it is well known that fast convergence usually requires adaptive weighting of PDE and boundary condition terms (see, e.g., \cite{FAIR}), the proposed initialization achieves up to eight orders of magnitude lower training loss compared to the default scheme. This distinct gap highlights a critical stability advantage: while the default initialization leads to training divergence, the proposed scheme ensures stable convergence toward the solution. This empirical finding is further substantiated by the NTK analysis in \ref{app:ntk2}, which links the proposed initialization to more favorable spectral properties.

\subsection{Training Divergence with Increasing Depth} \label{sec3.3}

The benchmarks considered so far involved relatively shallow networks, consistent with most current KAN applications, where architectures are typically limited to a small number of hidden layers. Within this setting, the proposed initialization was shown to improve accuracy and training stability across both function-fitting and forward PDE tasks. A natural question, however, is whether these gains extend to deeper architectures. To this end, we revisit Burgers' equation and additionally consider the Allen--Cahn equation (see \ref{app:pdes} for details), training cPIKANs of increasing depth to examine how network depth influences training stability under both initialization schemes.

\begin{figure}[t!]
	\centering
	\includegraphics[width=\linewidth]{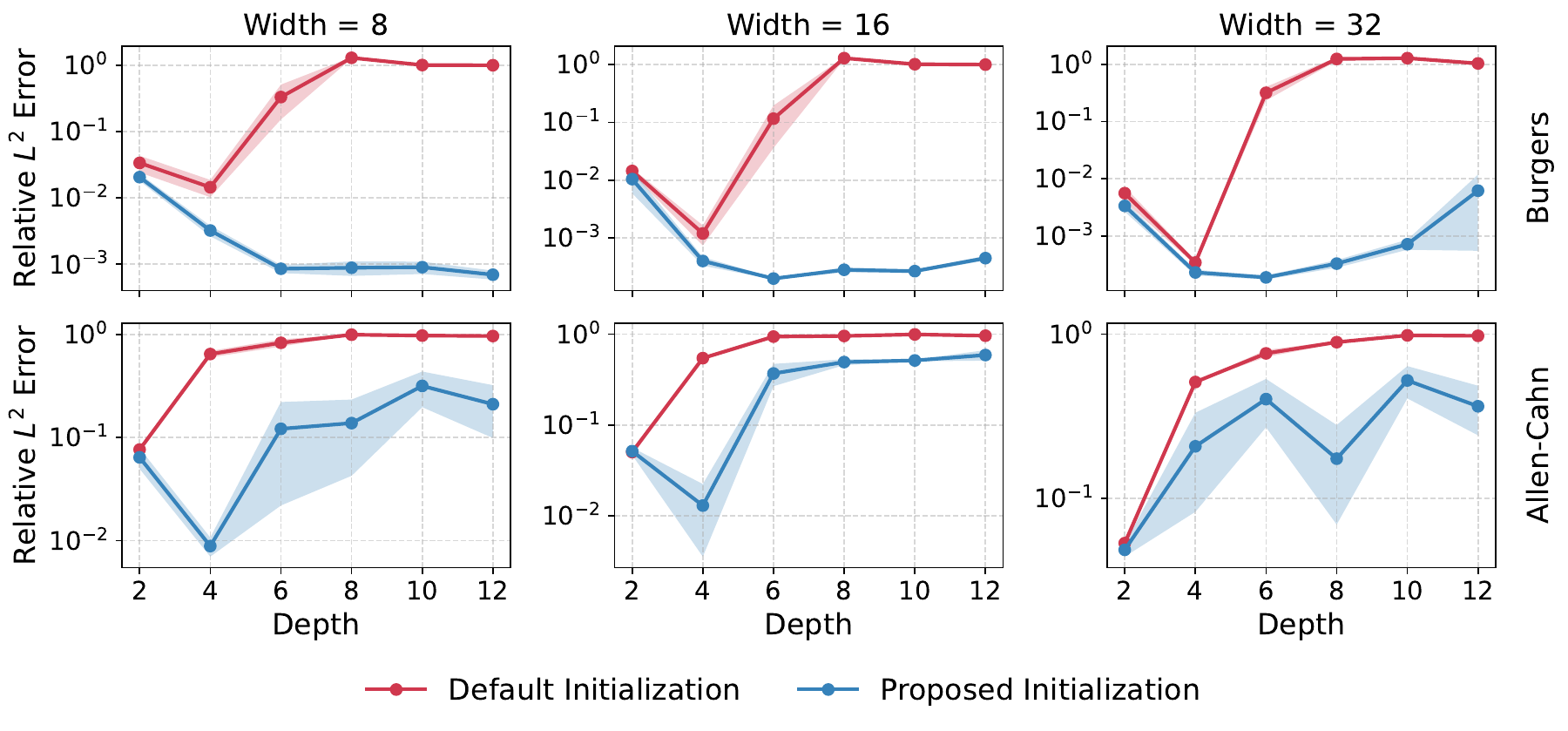}
	\caption{Relative $L^2$ error across increasing network depths for Burgers' (top row) and Allen--Cahn (bottom row) equations, under both default and proposed initialization schemes. Each column corresponds to a different network width (8, 16, 32). Solid lines show mean values over five random seeds, while shaded areas represent the SEM.}
	\label{fig4}
\end{figure}

In contrast to the previous benchmarks, here we adopt a standardized adaptive training pipeline, the configuration and hyperparameter settings of which are provided in \ref{app:impl2}, since both PDEs are trained for a large number of iterations and the Allen--Cahn equation in particular fails to yield accurate results without adaptive strategies \cite{Causal, RBA}. We consider Chebyshev-based KANs with polynomial order $D=5$, initialized under both the default and the proposed scheme, with architectures of varying widths $\{8, 16, 32\}$ and depths ranging from 2 to 12 hidden layers. Performance is evaluated in terms of the relative $L^2$ error with respect to the reference solution, with each experiment repeated five times using different random seeds. The results of these experiments are summarized in Figure \ref{fig4}, which presents the relative $L^2$ error of each network configuration under both initialization schemes for the two studied PDEs. Each column corresponds to a specific network width, with the horizontal axis of each subplot representing the network depth. Solid lines denote the mean relative $L^2$ error over the five independent runs, and the shaded regions indicate the corresponding SEM.

The results highlight the consistent advantage of the proposed initialization across all architectures and both PDEs, often yielding relative $L^2$ errors lower by several orders of magnitude compared to the default scheme. For the Burgers' equation, the effect is particularly striking: beyond a depth of four hidden layers, models initialized with the default scheme exhibit complete divergence, with relative errors of $\mathcal{O}(1)$, whereas the proposed initialization maintains stable training and achieves errors as low as $\mathcal{O}(10^{-3})$--$\mathcal{O}(10^{-4})$. Although some degradation in accuracy is observed at larger depths for the widest networks (width $32$), no divergence occurs, and the relative errors remain well below those of the default scheme. In contrast, for the Allen--Cahn equation, training instability persists for deeper networks under both initialization schemes. While the proposed initialization consistently improves performance compared to the default, the relative error grows rapidly with depth and reaches $\mathcal{O}(1)$ for the deepest configurations. The lowest errors are typically achieved at shallow depths (i.e., depth $2$ for width $32$ and depth $4$ for the other two cases), after which performance degrades, albeit not monotonically for all cases. Overall, while the proposed scheme substantially improves training stability and accuracy, it does not, by itself, guarantee stable convergence in all cases as network depth increases.

\section{Residual-Gated Adaptive KANs} \label{sec4}

While the proposed Glorot-like initialization improved training stability and accuracy across all benchmarks, its effectiveness in mitigating depth-related issues proved case-dependent. In particular, for Burgers' equation, cPIKANs initialized with the proposed scheme remained stable and convergent even at larger depths, whereas for the Allen--Cahn equation similar architectures exhibited divergence beyond a certain number of hidden layers. A similar phenomenon is observed in PINNs: when networks are initialized following the Glorot scheme, training tends to diverge as depth increases, regardless of the specific activation function among those commonly used in practice \cite{PirateNets}. However, before attempting to transfer to cPIKANs the remedies that have been proposed for this behavior in PINNs, it is first necessary to determine whether the underlying mechanisms are indeed analogous.

\subsection{KAN Derivatives at Initialization} \label{sec4.1}

In the case of MLP-based networks -- the backbone of PINNs -- \cite{PirateNets} demonstrated that training divergence with increasing depth originates from the behavior of the network's derivatives at initialization. To illustrate this, they considered a simplified MLP with scalar input and output, employing the hyperbolic tangent activation $\sigma(x)=\tanh(x)$, and focused on the first-order derivative with respect to the input. At initialization, the network operates in a near-linear regime where $\sigma(x) \approx x$, leading to the following approximation for the first-order derivative of Eq. \eqref{eq10} with respect to the input coordinate $x$:

\begin{equation}
	\frac{\partial u_j^{(l)}(x;\boldsymbol{\theta})}{\partial x}
	\;\approx\;
	\sum_{i=1}^{d_{l-1}} w_{ji}^{(l)} \,
	\frac{\partial u_i^{(l-1)}(x;\boldsymbol{\theta})}{\partial x}.
	\label{eq34}
\end{equation}

\noindent Assuming a network composed of $L$ hidden layers of width $d_\text{H}$, and noting that $\frac{\partial u^{(0)}\left(x;\boldsymbol{\theta}\right)}{\partial x} = 1$, the derivative of the network output can be recursively expressed as

\begin{equation}
	\frac{\partial u(x;\boldsymbol{\theta})}{\partial x}
	\;\approx\;
	\sum_{i=1}^{d_\text{H}}\sum_{n=1}^{d_\text{H}}\!\cdots\!\sum_{m=1}^{d_\text{H}}
	w_{1i}^{(L+1)} w_{in}^{(L)} \cdots w_{m1}^{(1)}.
	\label{eq35}
\end{equation}

\noindent This result shows that the derivative in Eq. \eqref{eq35} behaves as a deep linear network at initialization and, more importantly, is independent of the input $x$. This reveals a fundamental limitation in the expressivity of the derivative network and explains why, in the context of PIML -- where PDE residuals depend directly on network derivatives -- deep MLP-based architectures tend to diverge during training.

Under analogous simplifying assumptions for a KAN-based network, the derivative of Eq. \eqref{eq15} with respect to the input $x$ is given by

\begin{equation}
	\frac{\partial u_j^{(l)}(x;\boldsymbol{\theta})}{\partial x}
	\;=\;
	\sum_{i=1}^{d_{l-1}}\sum_{m=1}^{D}
	w_{jim}^{(l)} \,
	\frac{\partial B_m}{\partial x}\!\left[u_i^{(l-1)}(x;\boldsymbol{\theta})\right]
	\frac{\partial u_i^{(l-1)}(x;\boldsymbol{\theta})}{\partial x}.
	\label{eq36}
\end{equation}

\noindent For Chebyshev-based KANs, where the basis functions are given by Eq. \eqref{eq16}, it can be shown (see \ref{app:derivs2} for the detailed derivation) that, in the linear regime,

\begin{equation}
	\frac{\partial u_j^{(l)}(x;\boldsymbol{\theta})}{\partial x}
	\;\approx\;
	\sum_{i=1}^{d_{l-1}}
	\tilde{w}_{ji}^{(l)} \,
	\frac{\partial u_i^{(l-1)}(x;\boldsymbol{\theta})}{\partial x},
	\label{eq37}
\end{equation}

\noindent where 

\begin{equation}
	\tilde{w}_{ji}^{(l)} 
	\; = \;
	\sum_{m\,\mathrm{odd}}^{D}
	m\,w_{jim}^{(l)}\,(-1)^{\frac{m-1}{2}}.
	\label{eq38}
\end{equation}

\noindent Equation \eqref{eq37} is formally equivalent to Eq. \eqref{eq34}, indicating that, within the linear regime, the first-order derivative of a Chebyshev-based KAN behaves analogously to that of an MLP at initialization. This correspondence indicates that the observed training instabilities in deep cPIKANs arise from a similar mechanism identified in deep PINNs.

\subsection{Proposed Architecture} \label{sec4.2}

\begin{figure}[t!]
	\centering
	\includegraphics[width=\linewidth]{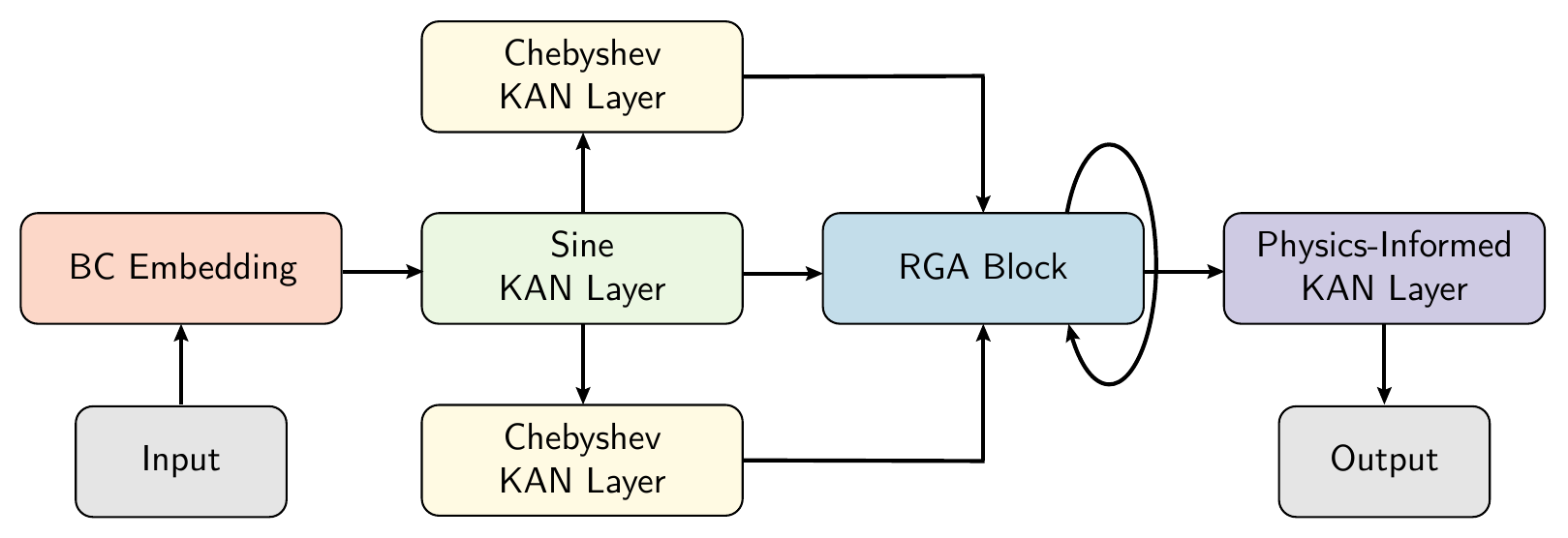}
	\caption{Schematic of the proposed RGA KAN architecture. Periodic boundary conditions, when present, are enforced directly through the BC Embedding layer. The embedded inputs are then passed through a sine-based KAN layer, whose outputs are split into three branches: two feeding Chebyshev-based KAN layers and one entering the first RGA block. Within each RGA block, the three signals are combined through gating operators and routed through adaptive skip connections, which dynamically modulate the effective network depth during training. Multiple RGA blocks can be stacked sequentially. The final output is produced by a physics-informed KAN layer, which incorporates prior information from the initial condition(s) when available.}
	\label{fig5}
\end{figure}

To address these instabilities in PINNs, \cite{PirateNets} introduced the PirateNet architecture (see \ref{app:pirate} for details). PirateNets incorporate several architectural components known to improve accuracy, such as random Fourier feature (RFF) embeddings \cite{RFF} and a physics-informed initialization of the final network layer. However, the key idea for resolving the depth-related issue is the introduction of an adaptive skip connection. This mechanism introduces a learnable gating parameter, $\alpha$, which dynamically modulates the network's effective depth during training, thereby stabilizing optimization and enabling convergence for deeper architectures. Inspired by this approach, we introduce \emph{Residual-Gated Adaptive Kolmogorov--Arnold Networks} (RGA KANs), the architecture of which is illustrated in Figure \ref{fig5}.

%This mechanism introduces a learnable gating parameter, $\alpha$, initialized at zero, which gradually increases during training, allowing the network's effective depth to grow progressively rather than being fixed from the outset. This gradual deepening stabilizes optimization and allows convergence for deeper networks. 

For a single input sample $\mathbf{x} \in \mathbb{R}^{d_\text{I}}$, where $d_\text{I}$ denotes the number of coordinates (including a possible temporal coordinate), periodic boundary conditions -- when present -- are enforced directly through the embedding layer. Specifically, if periodic boundary conditions apply to the $i$-th coordinate, the embedding is defined as

\begin{equation}
	\mathrm{Emb}(x_i) 
	= 
	\begin{bmatrix}
		\cos\left(\Omega_i x_i\right) \\[4pt]
		\sin\left(\Omega_i x_i\right)
	\end{bmatrix}
	\in \mathbb{R}^2,
	\qquad
	\Omega_i = \frac{2\pi}{L_i},
	\label{eq39}
\end{equation}

\noindent where $L_i$ is the length of the $i$-th coordinate's domain. In most cases considered in this work, where $x_i \in [-1,1]$, we have $\Omega_i = \pi$. After embedding, the input is mapped to $\tilde{\mathbf{x}} \in \mathbb{R}^{\tilde{d}_\text{I}}$, where $\tilde{d}_\text{I}$ is the new number of effective input coordinates.

The embedded input then passes through a sine-based KAN layer, whose output, $\mathbf{s} \in \mathbb{R}^{d_\text{H}}$, is computed as

\begin{equation}
	s_j 
	= 
	\sum_{i=1}^{\tilde{d}_\text{I}}\sum_{m=1}^{D_s}
	b^{s}_{jim}\,
	B^s_m\left(\tilde{x}_i\right)
	+ c^s_j, \quad 
	\label{eq40}
\end{equation}

\noindent where $c^s_j$ is a bias term and $B_m^s(\cdot)$, with $m = 1, \dots, D_s$, are sine-based basis functions defined by

\begin{equation}
	B^s_m(x) 
	= 
	\frac{\sin\left(\omega_m x + p_m\right) - \mu\left(\omega_m, p_m\right)}{\sigma\left(\omega_m, p_m\right)}.
	\label{eq41}
\end{equation}

\noindent Here, $\mu(\omega_m, p_m)$ and $\sigma(\omega_m, p_m)$ denote the mean and standard deviation of $\sin(\omega_m x + p_m)$, given by

\begin{align}
	\mu(\omega_m, p_m) &= \exp\left(-\frac{\omega_m^2}{2}\right)\sin\left(p_m\right),
	\label{eq42} \\[6pt]
	\sigma(\omega_m, p_m) &= \sqrt{\tfrac{1}{2} - \tfrac{1}{2}\exp\left(-2\omega_m^2\right)\cos\left(2p_m\right) - \mu(\omega_m, p_m)^2}.
	\label{eq43}
\end{align}

\noindent This basis function design is inspired by the ActLayer \cite{ActNet} and plays a similar role to RFF embeddings used in the PirateNet architecture. In preliminary experiments, RFF embeddings were found to degrade performance in our setting, whereas sine-based KAN layers preserved the benefits of trigonometric features, which have been shown to be particularly effective in many PDE problems \cite{ActNet}. In addition to the trainable coefficients $b^s_{jim}$ (initialized using the Glorot-like scheme proposed in this work), we also introduce trainable phase parameters $p_m$ (initialized at zero) and frequency parameters $\omega_m$ (initially sampled from a standard normal distribution), as in ActLayers.

At this stage, drawing inspiration from the Modified MLP architecture \cite{Wang1}, we define two gates using Chebyshev-based KAN layers:

\begin{equation}
	U_j 
	= 
	\sum_{i=1}^{d_\text{H}}\sum_{m=1}^{D}
	b^{u}_{jim}\,
	B_m\left(s_i\right)
	+ c^u_j,
	\qquad
	V_j 
	= 
	\sum_{i=1}^{d_\text{H}}\sum_{m=1}^{D}
	b^{v}_{jim}\,
	B_m\left(s_i\right)
	+ c^v_j,
	\label{eq44}
\end{equation}

\noindent where $\mathbf{U}, \mathbf{V} \in \mathbb{R}^{d_\text{H}}$. These gate outputs, together with the outputs of the sine-based KAN layer, form the inputs to the first RGA block. Considering a total of $N$ such RGA blocks and denoting the input to the $l$-th block by $\mathbf{x}^{(l)}$, with $l = 1, \dots, N$ and $\mathbf{x}^{(1)} = \mathbf{s}$, the forward pass through each block is defined recursively as follows:

\begin{align}
	f_j^{(l)} &= \sum_{i=1}^{d_\text{H}}\sum_{m=1}^{D} b^{(l)}_{jim}\, B_m\left(x_i^{(l)}\right) + c^{(l)}_j, \label{eq45} \\
	g_j^{(l)} &= f_j^{(l)}\, U_j + \left(1 - f_j^{(l)}\right) V_j, \label{eq46} \\
	z_j^{(l)} &= \beta\, g_j^{(l)} + \left(1-\beta\right)x_j^{(l)}, \label{eq47}
\end{align} 

\begin{align}
	\tilde{f}_j^{(l)} &= \sum_{i=1}^{d_\text{H}}\sum_{m=1}^{D} \tilde{b}^{(l)}_{jim}\, B_m\left(z_i^{(l)}\right) + \tilde{c}^{(l)}_j, \label{eq48} \\
	\tilde{g}_j^{(l)} &= \tilde{f}_j^{(l)}\, U_j + \left(1 - \tilde{f}_j^{(l)}\right) V_j, \label{eq49} \\
	x_j^{(l+1)} &= \alpha\, \tilde{g}_j^{(l)} + \left(1-\alpha\right)x_j^{(l)}, \label{eq50}
\end{align}

\noindent where the dimension of all intermediate and final outputs is $d_\text{H}$. All bias terms appearing in Eqs. \eqref{eq44}, \eqref{eq45} and \eqref{eq48} are initialized to zero, while the basis function coefficients are initialized using the proposed Glorot-like scheme of Eq. \eqref{eq33}.

%As in PirateNets, the skip connection governed by the parameter $\alpha$ is initialized at zero, ensuring that the contribution of each RGA block is initially suppressed and progressively activated during training. 

The skip connection governed by the parameter $\alpha$ is typically initialized either at zero, effectively suppressing the contribution of each RGA block at initialization, or at unity, enabling the network to start at its full intended depth while still allowing the effect of each block to be adaptively modulated during training. While PirateNets employ a three-layer block with a single adaptive skip connection, we adopt a two-layer design and introduce an additional adaptive parameter, $\beta$, after the first layer. When $\beta$ is initialized at $1$, the block behaves analogously to the original PirateNet block at initialization, whereas $\beta = 0$ corresponds to introducing an adaptive skip after each layer. Here, $\alpha$ controls the activation of the entire block, while $\beta$ affects only the first layer. This design was preferred over a direct three-layer port, as it proved more modular and yielded better results in preliminary tests. In terms of effective depth, an RGA KAN with $N$ blocks is equivalent to a conventional KAN with $L = 2N$ hidden layers.

The output of the final RGA block, $\mathbf{x}^{(N+1)} \in \mathbb{R}^{d_\text{H}}$, is mapped to the network output through a final Chebyshev-based KAN layer, defined as

\begin{equation}
	u_j 
	= 
	\sum_{i=1}^{d_\text{H}}\sum_{m=1}^{D} 
	b^{o}_{jim}\,
	B_m\left(x^{(N+1)}_i\right),
	\label{eq51}
\end{equation}

\noindent where no bias term is included. In the absence of additional information, the coefficients $b^{o}_{jim}$ are initialized using the same Glorot-like scheme proposed in Section~\ref{sec3.1}. However, if the PDE problem under consideration is equipped with an initial condition, this layer is instead physics-informed initialized. Specifically, the weights are chosen such that the network output approximates the initial condition at $t=0$ as accurately as possible over the entire temporal domain.

For the purposes of this physics-informed initialization, we re-index the pair $(i, m)$ into a single index $p = 1, \dots, d_\text{H}D$, yielding the equivalent form

\begin{equation}
	u_j 
	= 
	\sum_{p=1}^{d_\text{H}D} 
	\tilde{b}^o_{jp}\,
	\tilde{B}_p,
	\label{eq52}
\end{equation}

\noindent where $\tilde{B}_p$ denotes the activation of the $m$-th basis function evaluated at the $i$-th component of $\mathbf{x}^{(N+1)}$, with the composite index $p \leftrightarrow (i,m)$. The physics-informed initialization amounts to solving the least-squares problem

\begin{equation}
	\tilde{\mathbf{b}}^o 
	= 
	\arg\min_{\mathbf{b}}
	\left\| \mathbf{y}_0 - \mathbf{\tilde{B}}\,\mathbf{b} \right\|_2^2,
	\label{eq53}
\end{equation}

\noindent where $\mathbf{y}_0$ contains the initial-condition target values and $\mathbf{\tilde{B}}$ is the matrix of basis activations evaluated on the outputs of the final RGA block when the original inputs are set to the collocation points enforcing the initial condition. The vector $\tilde{\mathbf{b}}^o$ contains the optimal coefficients obtained by the least-squares fit and can be re-indexed back to the original $(i,m)$ indexing to yield the coefficients $b^{o}_{jim}$ of Eq. \eqref{eq51}. Note that, although here we focus on standard PDE benchmarks with initial conditions, this formulation can in principle incorporate arbitrary external data through $\mathbf{y}_0$, such as experimental measurements \cite{PirateNets}.

If non-periodic boundary conditions are present, they can be directly enforced after this final physics-informed layer by multiplying the network output by suitable boundary-shaping functions, thereby ensuring that the solution satisfies these constraints exactly \cite{ExactBCs}. Based on the above architectural components, for an RGA KAN model with $d_\text{O}$-dimensional output, the total number of trainable parameters can be explicitly determined as follows:

\begin{equation}
	\begin{aligned}
		\left|\boldsymbol{\theta}\right|
		&= \underbrace{d_\text{H}\left(\tilde{d}_\text{I}D_s + 1\right) + 2D_s}_{\text{Sine KAN Layer}}
		+ \overbrace{2d_\text{H}\left(d_\text{H}D + 1\right)}^{\text{Chebyshev KAN Gates}} \\[4pt]
		&\quad
		+ \underbrace{N\left[2d_\text{H}\left(d_\text{H}D + 1\right) + 2\right]}_{\text{RGA Blocks}}
		+ \overbrace{d_\text{O}d_\text{H} D}^{\text{Output Layer}} \\[6pt]
		&= 2d_\text{H}\left(d_\text{H}D + 1\right)\left(N+1\right) + 2N + 2D_s
		+ d_\text{H}\left(\tilde{d}_\text{I}D_s + d_\text{O}D + 1\right).
	\end{aligned}
	\label{eq54}
\end{equation}

\begin{figure}[b!]
	\centering
	\includegraphics[width=\linewidth]{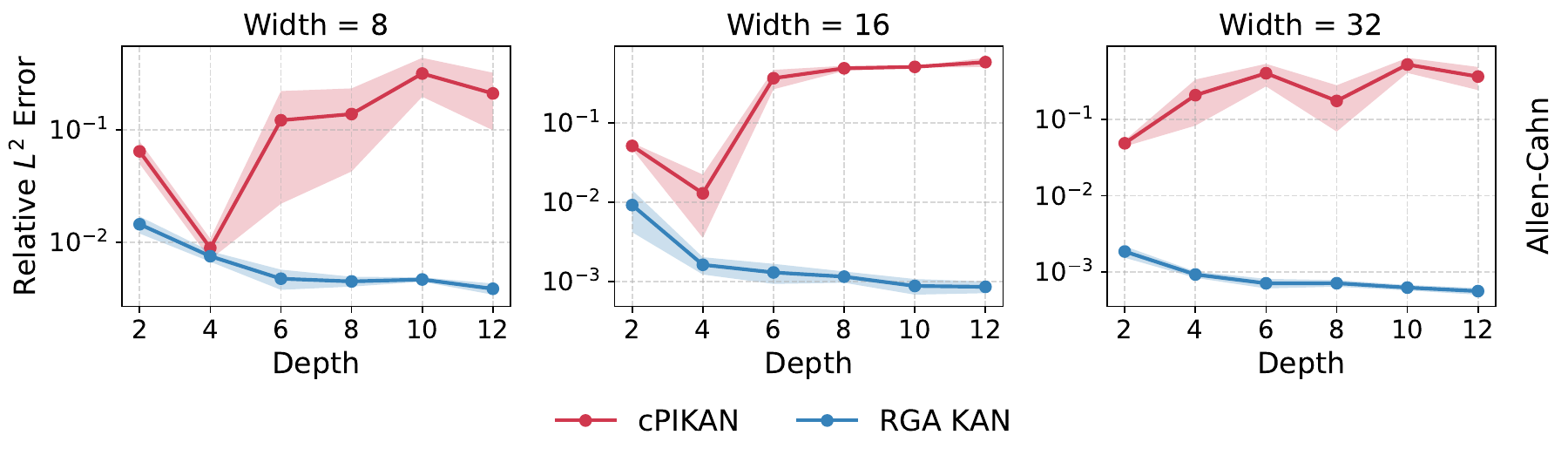}
	\caption{Relative $L^2$ error across increasing network depths for the Allen--Cahn equation, comparing RGA KANs and cPIKANs. Each column corresponds to a different network width (8, 16, 32) and the number of hidden layers for RGA KANs equals twice the number of RGA blocks. All results are averaged over five random seeds using the proposed Glorot-like initialization scheme. Solid lines indicate mean values and shaded areas denote the SEM.}
	\label{fig6}
\end{figure}

Using this architecture, we repeat the experiments of Section \ref{sec3.3} for the Allen--Cahn equation, where cPIKANs diverged with increasing depth. To ensure a fair comparison with the previous cPIKAN results, we employ identical parameter settings and use the same random seeds. Since the superiority of the proposed Glorot-like initialization scheme has already been established, we focus exclusively on this initialization strategy here. For parameters without a direct analogue in cPIKANs, we set $D_s = 5$ for the sine-based KAN layer and initialize both $\beta$ and $\alpha$ to zero in each RGA block. The corresponding results are depicted in Figure \ref{fig6}, where the reported number of hidden layers corresponds to twice the number of RGA blocks for the RGA KAN architecture.

The results demonstrate a significant improvement in stability and accuracy compared to the baseline cPIKANs. Across all widths and depths considered, RGA KANs maintain low relative $L^2$ errors without exhibiting divergence, even for the deepest networks. Moreover, a favorable scaling effect is observed: as the network width increases, the relative error decreases consistently, with the widest configuration (width $32$) yielding the best results across depths. More importantly, within each width setting, increasing the number of RGA blocks either preserves performance (plateau behavior) or further reduces the error -- a behavior that mirrors the improvements reported for PirateNets over regular MLPs \cite{PirateNets}.

As a closing remark, we note that we opted for $\alpha = 0$ and $\beta = 0$ at initialization, effectively initializing the network in a state that resembles a single-layer model and progressively increasing its effective depth during training. This choice follows a similar rationale to that of PirateNets, where gradual deepening contributes to stable optimization. However, in practice, the residual gates $\mathbf{U}$ and $\mathbf{V}$ can already act as stabilizing components, often preventing divergence even when the network is initialized at full depth. To investigate this further, in the next section -- where we present more extensive benchmarks across several PDEs -- we include ablation studies on the initialization values of $\alpha$ and $\beta$ to examine their effect on training dynamics and performance for each problem.

\subsection{The Lens of Information Bottleneck Theory} \label{sec4.3}

To better understand why the RGA KAN architecture not only achieves superior accuracy compared to baseline cPIKANs but also avoids performance degradation with increasing depth, we turn to Information Bottleneck (IB) theory to analyze the training dynamics of both architecture types. In supervised learning, neural networks aim to reproduce target outputs by progressively forming compressed internal representations of the inputs through their layer activations. According to IB theory, an optimally trained model preserves only the information relevant for reproducing the output while discarding irrelevant input details, effectively forming an ``information bottleneck'' \cite{IBIntro}. This learning process typically unfolds in two distinct phases, namely fitting and diffusion, separated by a phase transition \cite{IB1, IB2, IB3}; it is during the diffusion phase that the network develops its generalization capabilities. IB theory has also been applied to analyze the training dynamics of neural networks within the PIML framework \cite{IBTheory, RBA, FAIR, KKANs}, and has even been extended to incorporate a third phase within this context, termed diffusion equilibrium \cite{IBTheory} or total diffusion \cite{FAIR, KKANs}.

To detect phase transitions during training, two key indicators are typically monitored: the relative $L^2$ error and the batch-wise signal-to-noise ratio (SNR), defined as

\begin{equation}
	\text{SNR} = \frac{\left \lVert \mathbb{E}\left[ \nabla_{\boldsymbol{\theta}} \mathcal{L}_\text{batch}\left(\boldsymbol{\theta}\right) \right] \right \rVert_2}{\left \lVert \sqrt{ \mathbb{E}\left[ \left(\nabla_{\boldsymbol{\theta}} \mathcal{L}_\text{batch}\left(\boldsymbol{\theta}\right) - \mathbb{E}\left[ \nabla_{\boldsymbol{\theta}} \mathcal{L}_\text{batch}\left(\boldsymbol{\theta}\right) \right] \right)^2  \right] } \right \rVert_2},
	\label{eq55}
\end{equation}

\noindent where $\mathcal{L}_\text{batch}\left(\boldsymbol{\theta}\right)$ denotes the loss of Eq. \eqref{eq6} evaluated over a single batch of collocation points, and expectations are taken across all non-overlapping batches. Intuitively, the SNR measures the ratio between the mean gradient norm (signal) and its standard deviation (noise), reflecting the clarity of the learning signal during optimization.

In addition, recent studies have shown that the geometric complexity of the network can provide further insight into its training dynamics \cite{GeomCompl, KKANs}. This metric, defined via the discrete Dirichlet energy, is given by

\begin{equation}
	\text{Complexity} = \frac{1}{N}\sum_{i=1}^{N}{ \left \lVert \nabla_{t, \mathbf{x}} u\left(t^i, \mathbf{x}^i;\boldsymbol{\theta}\right) \right \rVert_F^2 },
	\label{eq56}
\end{equation}

\noindent where $\left\{\left(t^i, \mathbf{x}^i\right)\right\}_{i=1}^{N} = \left\{\left(t_\text{pde}^i,\mathbf{x}_\text{pde}^i\right)\right\}_{i=1}^{N_\text{pde}} \cup \left\{\left(0, \mathbf{x}_\text{ic}^i\right)\right\}_{i=1}^{N_\text{ic}}$ denotes the complete set of $N = N_\text{pde} + N_\text{ic}$ collocation points, and $\lVert \cdot \rVert_F$ is the Frobenius norm. Note that boundary condition points are not explicitly included here, as they are already enforced through the network architecture; otherwise, they would contribute to this set in the same manner.

\begin{figure}[t!]
	\centering
	\includegraphics[width=\linewidth]{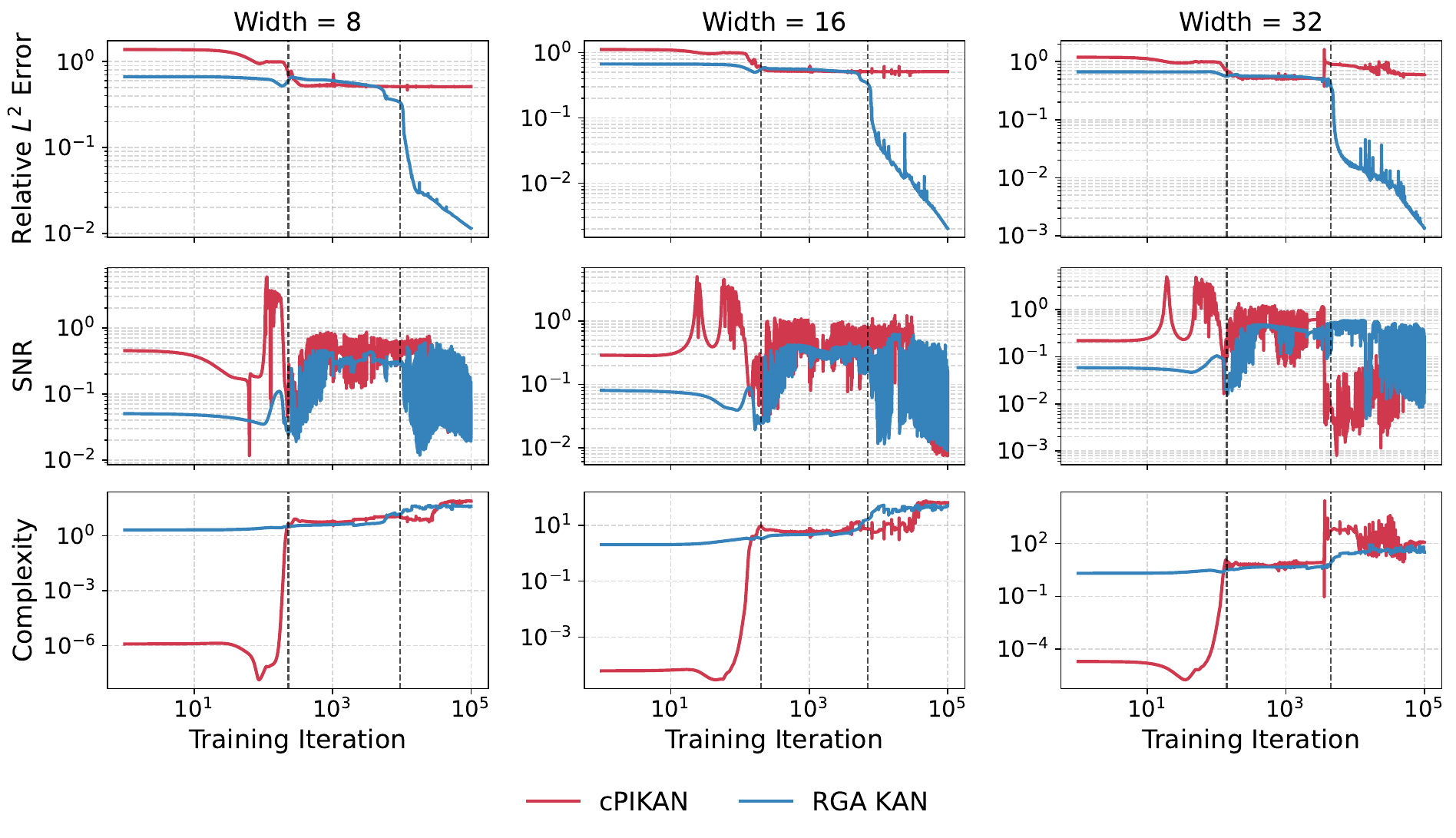}
	\caption{Training dynamics of cPIKAN and RGA KAN architectures of different widths in terms of relative $L^2$ error (top row), SNR (middle row) and geometric complexity (bottom row). Dashed vertical lines indicate the transitions between training phases (fitting, diffusion and diffusion equilibrium) for the RGA KAN models.}
	\label{fig7}
\end{figure}

To investigate the differences in training dynamics between cPIKANs and RGA KANs, we repeat the experiments for the Allen--Cahn equation using networks with $12$ hidden layers (equivalently, $6$ RGA blocks), and widths $8$, $16$ and $32$. All hyperparameters for RGA KANs are kept identical to those in Section \ref{sec4.2}, while cPIKANs use the same settings as in Section \ref{sec3.3}. During each training iteration, we record the relative $L^2$ error as well as the SNR and geometric complexity defined in Eqs. \eqref{eq55} and \eqref{eq56}, respectively. Among the five independently trained instances per architecture (each initialized with a different random seed), Figure \ref{fig7} reports the results corresponding, for each width, to the run with the highest final relative $L^2$ error for the RGA KAN architecture.

The training dynamics of the RGA KAN architecture, analyzed through the lens of IB theory as studied in the context of PIML \cite{IBTheory, FAIR, KKANs}, reveal a clear progression through all three learning phases. During the initial fitting phase, which spans roughly the first $200$ training iterations, the relative $L^2$ error remains nearly constant, while the geometric complexity increases steadily. Simultaneously, the SNR exhibits a brief oscillatory pattern before beginning to decline. In this stage, the network primarily memorizes the training data without significant generalization. A subsequent transition marks the onset of the diffusion phase, characterized by a steadily increasing and fluctuating SNR, a slight decrease in relative $L^2$ error and a continued increase in geometric complexity. This phase corresponds to an exploratory stage in which the model identifies more effective learning directions and begins to generalize. Finally, the network enters the diffusion equilibrium phase, during which the SNR reaches a stable, though still oscillatory, plateau, the geometric complexity continues to rise and then also plateaus, and the relative $L^2$ error drops sharply, indicating a rapid improvement in generalization and predictive accuracy. Remarkably, this qualitative behavior is consistent across all three network widths. The only systematic difference is the timing of the phase transitions, which occur earlier as the model width increases. This observation aligns with our previous findings: when combined with the proposed initialization, increasing the capacity of the RGA KAN architecture does not lead to divergence but instead improves model accuracy.

In contrast, the behavior of the cPIKAN models differs substantially. While they exhibit a fitting phase similar to that of the RGA KAN, the increase in geometric complexity is far more abrupt and several orders of magnitude larger -- an effect previously reported for cPIKANs in \cite{KKANs}. After transitioning to the diffusion phase, these models never reach the diffusion equilibrium phase. The geometric complexity plateaus prematurely, the SNR exhibits strong oscillations without converging to a stable plateau, and the relative $L^2$ error stagnates. As a result, the cPIKAN models fail to generalize, explaining the poor performance and divergence observed at large depths for the Allen--Cahn equation.

\begin{figure}[b!]
	\centering
	\includegraphics[width=\linewidth]{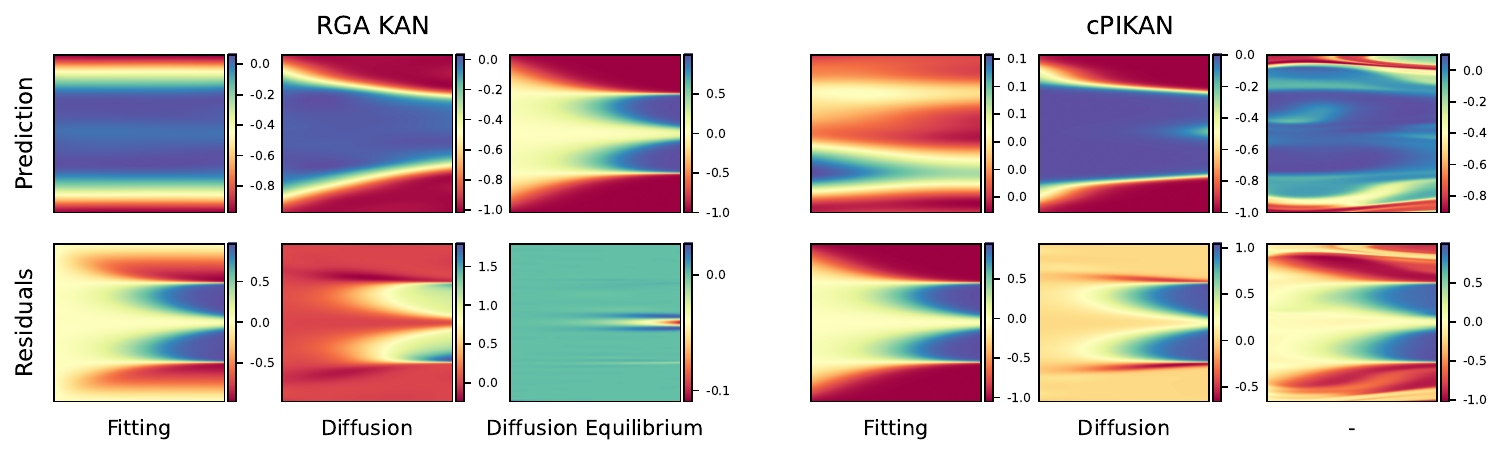}
	\caption{Evolution of the model predictions (top row) and residuals (bottom row) across the three IB training phases for RGA KAN (left) and cPIKAN (right) architectures, using a width-$16$ configuration. For the RGA KAN model, the predictions become progressively more structured and closely match the reference solution as training proceeds, while the residuals approach noise during the diffusion equilibrium phase. In contrast, the cPIKAN model fails to undergo a clear second transition, resulting in residuals that remain semi-ordered and predictions that deviate significantly from the reference solution.}
	\label{fig8}
\end{figure}

These results provide strong evidence, from the perspective of IB theory, for why the RGA KAN architecture maintains stability and achieves superior accuracy where cPIKANs fail. An additional layer of insight can be gained by examining the evolution of the models' predictions and residuals (the difference between the reference solution and the model output) across the identified phases. This is illustrated in Figure \ref{fig8}, where we show results for the width $16$ configuration for both architectures (RGA KAN on the left, cPIKAN on the right). As expected, during the fitting phase, when the model has not yet learned to generalize, the predictions are overly simplistic and the residuals structured, effectively mirroring the inverse of the reference solution. During the diffusion phase, the predictions gradually become more structured and the residuals more disordered. However, while the RGA KAN undergoes a clear second transition, leading to predictions closely matching the reference solution and residuals steadily approaching noise, the cPIKAN remains stuck in a semi-ordered state, never fully generalizing.

\section{Experimental Results} \label{sec5}

Having established that RGA KANs, when initialized with the proposed Glorot-like scheme, remain stable as depth increases and avoid the divergence observed in baseline cPIKANs -- both empirically (Section \ref{sec4.2}) and through the lens of IB theory (Section \ref{sec4.3}) -- we now turn to a series of forward PDE benchmarks and ablation studies. With the exception of the experiments presented in Sections \ref{sec5.8} and \ref{sec5.9}, we use the exact same hyperparameter and optimization settings as those used in Section \ref{sec3.3}. These settings, the details of which can be found in \ref{app:impl3}, are applied uniformly across all architectures considered. The goal of this section is not to perform extensive hyperparameter sweeps to obtain state-of-the-art performance for each PDE individually, but rather to demonstrate that RGA KANs already achieve strong results without any task-specific tuning, using a single, generic configuration.

For the main experiments, we use RGA KANs of width $16$ and depth $12$ (corresponding to $N=6$ RGA blocks). To provide a fair comparison, we also evaluate baseline cPIKANs initialized with the proposed Glorot-like scheme and PirateNets, which represent the current state of the art MLP-based architecture on several PDE benchmarks \cite{PirateNets}. To match parameter counts across architectures at the same depth, we adjust widths accordingly: for cPIKANs we use width $18$ and for PirateNets we use width $36$. All experiments are repeated with three random seeds for statistical significance. For RGA KANs, we additionally investigate four variants corresponding to different initializations of the adaptive skip parameters, with $(\alpha,\beta) \in \{0,1\} \times \{0,1\}$. After reporting the benchmark results for each architecture, we identify the $(\alpha,\beta)$ configuration that achieves the lowest error and use it as the reference in the subsequent ablation studies.

The ablation experiments aim to quantify the contribution of each adaptive training component to the overall performance of RGA KANs. To this end, we perform the following sequence: (i) train with RBA alone, disabling all other adaptive techniques; (ii) disable RBA while keeping all other techniques enabled; (iii) disable RBA and RAD while keeping causal training and LRA; (iv) disable RBA and causal training while keeping RAD and LRA; and finally (v) disable RBA and LRA while keeping RAD and causal training. Each configuration is again trained with three different random seeds. The following subsections present the results for each PDE benchmark.

\subsection{Allen--Cahn Equation} \label{sec5.1}

\begin{figure}[b!]
\centering
\includegraphics[width=\linewidth]{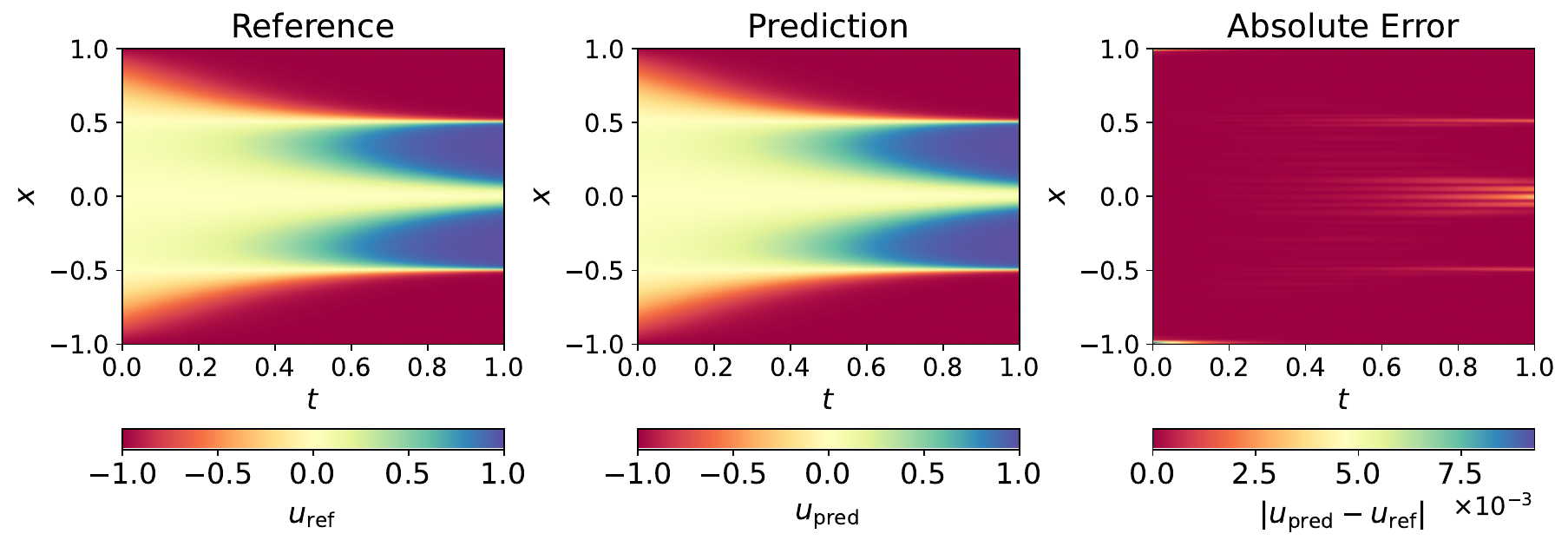}
\caption{Reference solution (left), RGA KAN prediction (middle) and absolute error (right) for the Allen--Cahn equation, shown for the random seed corresponding to the best-performing model instance.}
\label{fig9}
\end{figure}

The first benchmark considered is the Allen--Cahn equation (see \ref{app:pdes1} for details), which has already served as the testbed in previous sections. We begin by examining the effect of different $\left(\alpha, \beta\right)$ initializations on RGA KAN performance. As shown in Table \ref{tab1}, initializing with $\alpha = 1$ and $\beta = 0$ yields the best results, closely followed by $\alpha = 1, \beta = 1$. From the three trained instances under this optimal setting, we retain the one achieving the lowest relative $L^2$ error ($3.96 \times 10^{-4}$) for visualization. Figure \ref{fig9} compares the model prediction against the reference solution and shows the absolute error field, which remains at most $\mathcal{O}\left(10^{-3}\right)$.

\begin{table}[b!]
	\centering
	\renewcommand{\arraystretch}{1.2}
	\setlength{\tabcolsep}{8pt}
	\small
	\begin{tabular}{ccc}
		\hline
		\textbf{Configuration} & \textbf{Relative $L^2$ Error} & \textbf{Final Loss} \\
		\hline
		$\alpha = 0$, $\beta = 0$ & $(1.62 \pm 0.20) \times 10^{-3}$ & $(2.99 \pm 0.37) \times 10^{-6}$ \\
		$\mathbf{\boldsymbol{\alpha} = 1}$, $\mathbf{\boldsymbol{\beta} = 0}$ & $\mathbf{(4.46 \pm 0.25) \times 10^{-4}}$ & $\mathbf{(5.27 \pm 0.51) \times 10^{-7}}$ \\
		$\alpha = 0$, $\beta = 1$ & $(1.39 \pm 0.43) \times 10^{-3}$ & $(1.89 \pm 0.54) \times 10^{-6}$ \\
		$\alpha = 1$, $\beta = 1$ & $(4.99 \pm 0.84) \times 10^{-4}$ & $(2.78 \pm 0.49) \times 10^{-7}$ \\
		\hline
	\end{tabular}
	\caption{Results for different RGA KAN $(\alpha, \beta)$ initializations on the Allen--Cahn equation. Reported values are mean $\pm$ SEM over three seeds. The best performing configuration in terms of relative $L^2$ error is indicated in bold.}
	\label{tab1}
\end{table}

We next compare RGA KANs with baseline cPIKAN and PirateNet architectures under similar total parameter numbers. Table \ref{tab2} summarizes the results. While cPIKAN and PirateNet are more time-efficient per iteration, RGA KAN achieves a substantially lower relative $L^2$ error, outperforming PirateNet by roughly an order of magnitude. As expected, the cPIKAN models diverge, reflecting the absence of a transition to the diffusion equilibrium phase observed in Section \ref{sec4.3}. PirateNets exhibit stable training but are consistently outperformed by RGA KANs in terms of accuracy. Notably, all RGA KAN configurations in Table \ref{tab1} outperform PirateNet, indicating robustness with respect to the initial choice of $\left(\alpha, \beta\right)$. Regarding the required training time per iteration, the RGA KAN architecture is slower than the other two, which is consistent across all benchmarks. This is expected, as cPIKANs do not include additional gating operations, while PirateNets rely on MLPs rather than KANs, resulting in faster iterations.

\begin{table}[t!]
	\centering
	\renewcommand{\arraystretch}{1.2}
	\setlength{\tabcolsep}{8pt}
	\small
	\begin{tabular}{cccc}
		\hline
		\textbf{Architecture} & \textbf{Parameters} & \textbf{Relative $L^2$ Error} & \textbf{Time / Iter.} \\
		\hline
		cPIKAN & 18,397 & $(5.21 \pm 0.01) \times 10^{-1}$ & 3.44 ms \\
		PirateNet & 19,246 & $(2.46 \pm 1.45) \times 10^{-3}$ & 3.61 ms \\
		\textbf{RGA KAN} & \textbf{18,502} & $\mathbf{(4.46 \pm 0.25) \times 10^{-4}}$ & \textbf{5.51 ms} \\
		\hline
	\end{tabular}
	\caption{Performance comparison on the Allen--Cahn equation across different architectures. Reported values are mean $\pm$ SEM over three random seeds. The RGA KAN row uses the best $(\alpha,\beta)$ initialization from Table \ref{tab1}. The best performing architecture in terms of relative $L^2$ error is indicated in bold.}
	\label{tab2}
\end{table}

\begin{table}[b!]
	\centering
	\renewcommand{\arraystretch}{1.2}
	\setlength{\tabcolsep}{10pt}
	\footnotesize
	\begin{tabular}{cccccc}
		\hline
		\textbf{RBA} & \textbf{RAD} & \textbf{Causal} & \textbf{LRA} & \textbf{Relative $L^2$ Error} & \textbf{Time / Iter.} \\
		\hline
		\cmark & \cmark & \cmark & \cmark & $(4.46 \pm 0.25) \times 10^{-4}$ & 5.51 ms \\
		\cmark & \xmark & \xmark & \xmark & $(2.88 \pm 2.85) \times 10^{-1}$ & 5.37 ms \\
		\xmark & \cmark & \cmark & \cmark & $(8.00 \pm 2.92) \times 10^{-4}$ & 5.33 ms \\
		\xmark & \xmark & \cmark & \cmark & $(1.43 \pm 0.40) \times 10^{-3}$ & 5.30 ms \\
		\xmark & \cmark & \xmark & \cmark & $(3.66 \pm 1.03) \times 10^{-3}$ & 5.27 ms \\
		\xmark & \cmark & \cmark & \xmark & $(9.26 \pm 1.84) \times 10^{-3}$ & 5.30 ms \\
		\hline
	\end{tabular}
	\caption{Ablation study on adaptive training components for the Allen--Cahn equation using the best $(\alpha,\beta)$ initialization from Table \ref{tab1}. Each row corresponds to a different combination of enabled (\cmark) or disabled (\xmark) components. Reported values for relative $L^2$ error are mean $\pm$ SEM over three seeds.}
	\label{tab3}
\end{table}

Finally, we quantify the contribution of the adaptive training components through ablation studies (Table \ref{tab3}). We include the training time per iteration in this table to demonstrate that the computational overhead of enabling or disabling specific adaptive components is marginal; as shown, the time per iteration remains approximately constant across all configurations. Given this consistent behavior, we omit this column in the ablation tables for the subsequent benchmarks to avoid redundancy. In terms of accuracy, using only RBA leads to the largest errors and high variability, suggesting strong sensitivity to weight initialization. Disabling RBA while retaining the other components degrades performance by less than one order of magnitude, indicating that the remaining adaptive techniques are sufficient to preserve stability. Among them, LRA has the largest individual impact, followed by causal training, while RAD plays a less dominant role for this PDE.

\subsection{Burgers' Equation} \label{sec5.2}

\begin{table}[b!]
	\centering
	\renewcommand{\arraystretch}{1.2}
	\setlength{\tabcolsep}{8pt}
	\small
	\begin{tabular}{ccc}
		\hline
		\textbf{Configuration} & \textbf{Relative $L^2$ Error} & \textbf{Final Loss} \\
		\hline
		$\alpha = 0$, $\beta = 0$ & $(4.57 \pm 1.25) \times 10^{-4}$ & $(1.13 \pm 0.42) \times 10^{-4}$ \\
		$\alpha = 1$, $\beta = 0$ & $(7.34 \pm 2.07) \times 10^{-4}$ & $(1.69 \pm 0.68) \times 10^{-4}$ \\
		$\mathbf{\boldsymbol{\alpha} = 0}$, $\mathbf{\boldsymbol{\beta} = 1}$ & $\mathbf{(3.06 \pm 0.31) \times 10^{-4}}$ & $\mathbf{(4.14 \pm 0.48) \times 10^{-5}}$ \\
		$\alpha = 1$, $\beta = 1$ & $(3.07 \pm 0.84) \times 10^{-4}$ & $(2.87 \pm 1.12) \times 10^{-5}$ \\
		\hline
	\end{tabular}
	\caption{Results for different RGA KAN $(\alpha, \beta)$ initializations on Burgers' equation. Reported values are mean $\pm$ SEM over three seeds. The best performing configuration in terms of relative $L^2$ error is indicated in bold.}
	\label{tab4}
\end{table}

\begin{figure}[b!]
	\centering
	\includegraphics[width=\linewidth]{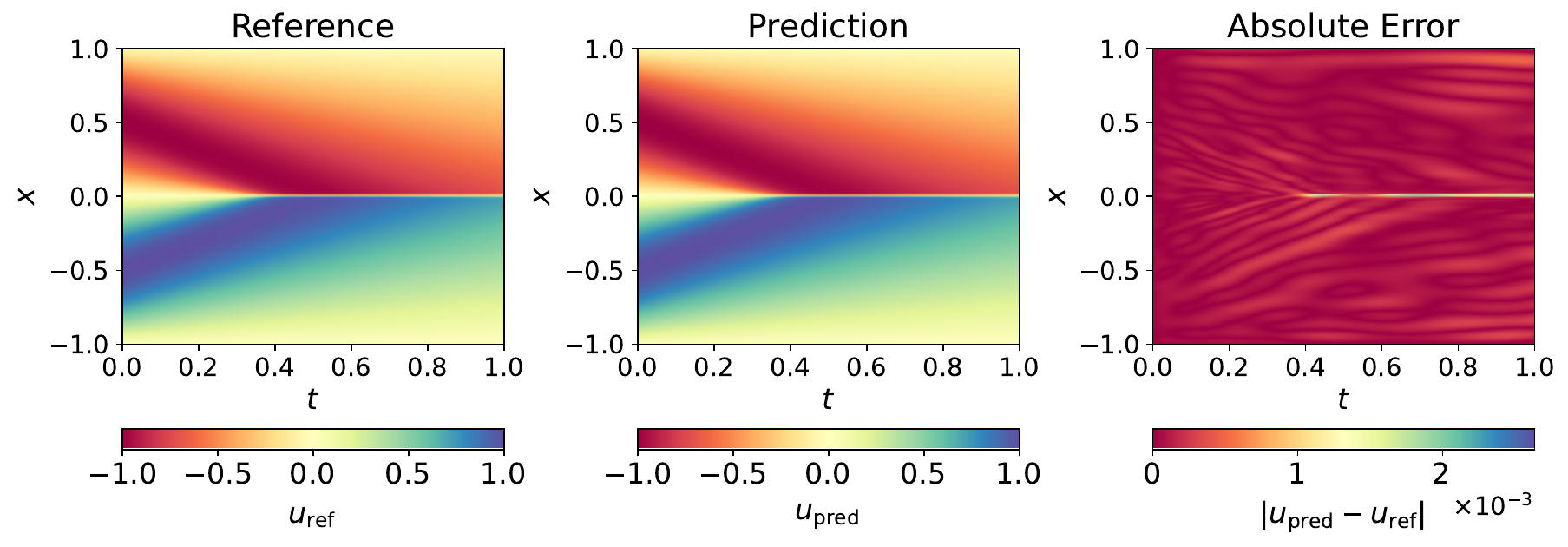}
	\caption{Reference solution (left), RGA KAN prediction (middle) and absolute error (right) for Burgers' equation, shown for the random seed corresponding to the best-performing model instance.}
	\label{fig10}
\end{figure}

We next consider Burgers' equation (see \ref{app:pdes2} for details) and follow the same experimental procedure. As summarized in Table \ref{tab4}, the best-performing configuration corresponds to $\alpha = 0$ and $\beta = 1$, closely followed by $\alpha = 1$, $\beta = 1$. Although the difference in mean relative $L^2$ error between these two configurations is small, the former exhibits a substantially lower SEM, indicating more robustness. We therefore retain $\alpha = 0$, $\beta = 1$ for the subsequent experiments. Figure \ref{fig10} shows the predicted solution for the seed with the lowest relative $L^2$ error ($2.46 \times 10^{-4}$), together with the reference solution and the absolute error field.

\begin{table}[b!]
	\centering
	\renewcommand{\arraystretch}{1.2}
	\setlength{\tabcolsep}{8pt}
	\small
	\begin{tabular}{cccc}
		\hline
		\textbf{Architecture} & \textbf{Parameters} & \textbf{Relative $L^2$ Error} & \textbf{Time / Iter.} \\
		\hline
		cPIKAN & 18,307 & $(5.13 \pm 1.18) \times 10^{-4}$ & 3.52 ms \\
		PirateNet & 19,228 & $(5.37 \pm 1.32) \times 10^{-4}$ & 3.88 ms \\
		\textbf{RGA KAN} & \textbf{18,422} & $\mathbf{(3.06 \pm 0.31) \times 10^{-4}}$ & \textbf{5.72 ms} \\
		\hline
	\end{tabular}
	\caption{Performance comparison on Burgers' equation across different architectures. Reported values are mean $\pm$ SEM over three random seeds. The RGA KAN row uses the best $(\alpha,\beta)$ initialization from Table \ref{tab4}. The best performing architecture in terms of relative $L^2$ error is indicated in bold.}
	\label{tab5}
\end{table}

\begin{table}[b!]
	\centering
	\renewcommand{\arraystretch}{1.2}
	\setlength{\tabcolsep}{10pt}
	\small
	\begin{tabular}{ccccc}
		\hline
		\textbf{RBA} & \textbf{RAD} & \textbf{Causal} & \textbf{LRA} & \textbf{Relative $L^2$ Error} \\
		\hline
		\cmark & \cmark & \cmark & \cmark & $(3.06 \pm 0.31) \times 10^{-4}$ \\
		\cmark & \xmark & \xmark & \xmark & $(1.06 \pm 0.39) \times 10^{-2}$ \\
		\xmark & \cmark & \cmark & \cmark & $(2.50 \pm 0.62) \times 10^{-4}$ \\
		\xmark & \xmark & \cmark & \cmark & $(4.08 \pm 1.95) \times 10^{-3}$ \\
		\xmark & \cmark & \xmark & \cmark & $(6.47 \pm 1.16) \times 10^{-4}$ \\
		\xmark & \cmark & \cmark & \xmark & $(3.34 \pm 0.64) \times 10^{-4}$ \\
		\hline
	\end{tabular}
	\caption{Ablation study on adaptive training components for Burgers' equation using the best $(\alpha,\beta)$ initialization from Table \ref{tab4}. Each row corresponds to a different combination of enabled (\cmark) or disabled (\xmark) components. Reported values for relative $L^2$ error are mean $\pm$ SEM over three seeds.}
	\label{tab6}
\end{table}

We then compare RGA KAN with baseline cPIKAN and PirateNet architectures at matched parameter counts. Table \ref{tab5} reports the corresponding results. Again, the training times per iteration are lower for the baseline architectures, and RGA KAN achieves the lowest relative $L^2$ error, approximately halving the error of cPIKAN and PirateNet. Interestingly, unlike the Allen--Cahn case, the cPIKAN model initialized using our proposed Glorot-like scheme does not diverge here and even achieves slightly better performance than PirateNet. However, given the overlapping standard errors, the two are essentially comparable. RGA KAN exhibits both the lowest mean error and the smallest variability across seeds.

Finally, we examine the contribution of each adaptive training technique (Table \ref{tab6}). Enabling only RBA leads to a sharp increase in relative error by nearly two orders of magnitude, though training remains stable. Strikingly, when RBA is disabled but the other adaptive techniques are retained, the error actually drops slightly below the fully adaptive configuration, indicating that RBA may act as a mild hindrance in this specific setting. Among the remaining techniques, RAD resampling has the largest individual impact, followed by causal training and LRA. This pattern contrasts with the Allen--Cahn case, highlighting that the relative importance of adaptive components can depend strongly on the PDE at hand.

\subsection{Korteweg--De Vries Equation} \label{sec5.3}

\begin{table}[b!]
	\centering
	\renewcommand{\arraystretch}{1.2}
	\setlength{\tabcolsep}{8pt}
	\small
	\begin{tabular}{ccc}
		\hline
		\textbf{Configuration} & \textbf{Relative $L^2$ Error} & \textbf{Final Loss} \\
		\hline
		$\alpha = 0$, $\beta = 0$ & $(4.73 \pm 0.61) \times 10^{-3}$ & $(1.58 \pm 0.44) \times 10^{-3}$ \\
		$\alpha = 1$, $\beta = 0$ & $(4.45 \pm 0.45) \times 10^{-3}$ & $(1.40 \pm 0.26) \times 10^{-3}$ \\
		$\mathbf{\boldsymbol{\alpha} = 0}$, $\mathbf{\boldsymbol{\beta} = 1}$ & $\mathbf{(3.87 \pm 0.52) \times 10^{-3}}$ & $\mathbf{(8.31 \pm 0.76) \times 10^{-4}}$ \\
		$\alpha = 1$, $\beta = 1$ & $(7.84 \pm 0.68) \times 10^{-3}$ & $(1.42 \pm 0.33) \times 10^{-3}$ \\
		\hline
	\end{tabular}
	\caption{Results for different RGA KAN $(\alpha, \beta)$ initializations on the Korteweg--De Vries equation. Reported values are mean $\pm$ SEM over three seeds. The best performing configuration in terms of relative $L^2$ error is indicated in bold.}
	\label{tab7}
\end{table}

\begin{figure}[b!]
	\centering
	\includegraphics[width=\linewidth]{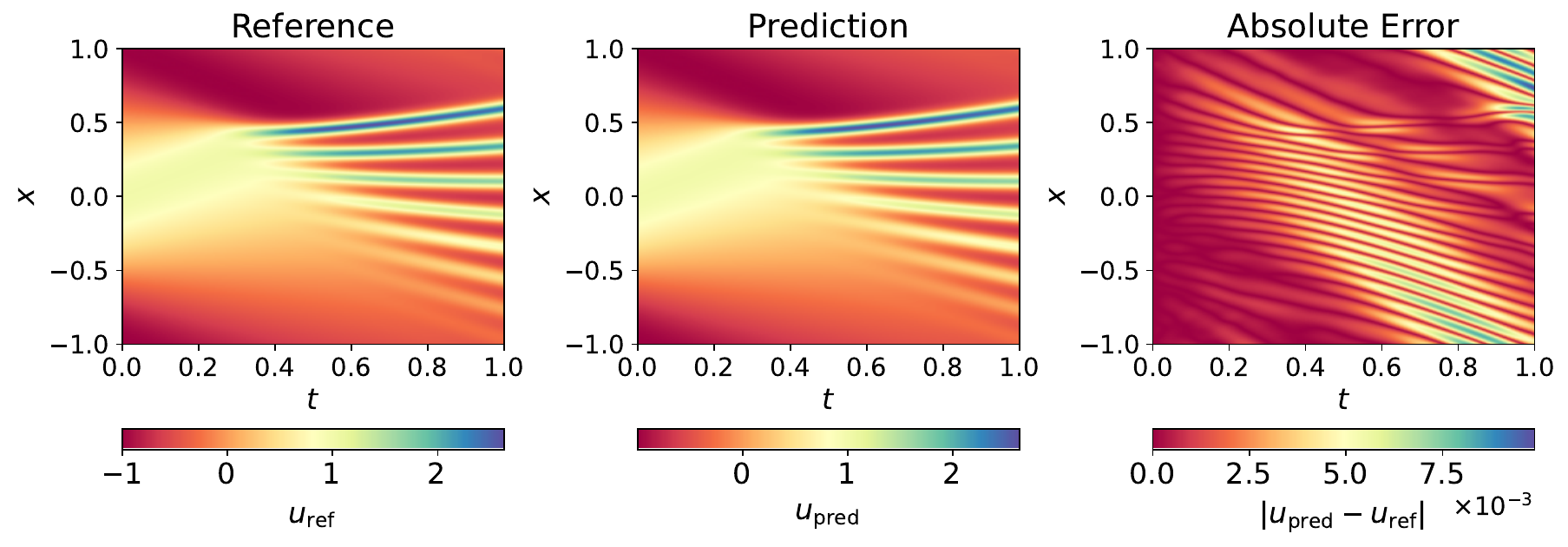}
	\caption{Reference solution (left), RGA KAN prediction (middle) and absolute error (right) for the Korteweg--De Vries equation, shown for the random seed corresponding to the best-performing model instance.}
	\label{fig11}
\end{figure}

We then turn to the Korteweg--De Vries equation (see \ref{app:pdes3} for details). As summarized in Table \ref{tab7}, the best-performing configuration once again corresponds to $\alpha = 0$, $\beta = 1$. Using the best-performing seed, with a final relative $L^2$ error of $3.21 \times 10^{-3}$, we plot the predicted solution alongside the reference and the absolute error field (Figure \ref{fig11}).

Table \ref{tab8} presents the comparison between architectures. Interestingly, PirateNet diverges in this setting, with a high relative $L^2$ error of approximately $7.7 \times 10^{-1}$ and very low SEM, indicating consistent failure across seeds. This stands in contrast to the results reported in \cite{PirateNets}, where PirateNet remained stable at similar depths but larger widths, hinting at a sensitivity to model capacity or hyperparameter choices. The cPIKAN model, while also performing poorly, achieves a lower error than PirateNet, suggesting that the proposed Glorot-like initialization once again helps stabilize training. RGA KANs, by comparison, achieve an error nearly two orders of magnitude lower than both baselines.

\begin{table}[b!]
	\centering
	\renewcommand{\arraystretch}{1.2}
	\setlength{\tabcolsep}{8pt}
	\small
	\begin{tabular}{cccc}
		\hline
		\textbf{Architecture} & \textbf{Parameters} & \textbf{Relative $L^2$ Error} & \textbf{Time / Iter.} \\
		\hline
		cPIKAN & 18,397 & $(1.16 \pm 0.03) \times 10^{-1}$ & 4.52 ms \\
		PirateNet & 19,246 & $(7.73 \pm 0.10) \times 10^{-1}$ & 4.88 ms \\
		\textbf{RGA KAN} & \textbf{18,502} & $\mathbf{(3.87 \pm 0.52) \times 10^{-3}}$ & \textbf{7.44 ms} \\
		\hline
	\end{tabular}
	\caption{Performance comparison on the Korteweg--De Vries equation across different architectures. Reported values are mean $\pm$ SEM over three random seeds. The RGA KAN row uses the best $(\alpha,\beta)$ initialization from Table \ref{tab7}. The best performing architecture in terms of relative $L^2$ error is indicated in bold.}
	\label{tab8}
\end{table}

\begin{table}[b!]
	\centering
	\renewcommand{\arraystretch}{1.2}
	\setlength{\tabcolsep}{10pt}
	\small
	\begin{tabular}{ccccc}
		\hline
		\textbf{RBA} & \textbf{RAD} & \textbf{Causal} & \textbf{LRA} & \textbf{Relative $L^2$ Error} \\
		\hline
		\cmark & \cmark & \cmark & \cmark & $(3.87 \pm 0.52) \times 10^{-3}$ \\
		\cmark & \xmark & \xmark & \xmark & $(5.76 \pm 1.31) \times 10^{-1}$ \\
		\xmark & \cmark & \cmark & \cmark & $(5.99 \pm 0.54) \times 10^{-3}$ \\
		\xmark & \xmark & \cmark & \cmark & $(7.80 \pm 0.70) \times 10^{-1}$ \\
		\xmark & \cmark & \xmark & \cmark & $(4.20 \pm 1.23) \times 10^{-2}$ \\
		\xmark & \cmark & \cmark & \xmark & $(2.41 \pm 0.24) \times 10^{-2}$ \\
		\hline
	\end{tabular}
	\caption{Ablation study on adaptive training components for the Korteweg--De Vries equation using the best $(\alpha,\beta)$ initialization from Table \ref{tab7}. Each row corresponds to a different combination of enabled (\cmark) or disabled (\xmark) components. Reported values for relative $L^2$ error are mean $\pm$ SEM over three seeds.}
	\label{tab9}
\end{table}

The ablation results are summarized in Table \ref{tab9}. Retaining only RBA leads to significant performance degradation and divergence, as reflected by an increase of nearly two orders of magnitude in error. Conversely, when RBA is disabled and the other adaptive techniques remain active, the model preserves stable training with only a moderate error increase compared to the fully adaptive configuration. Among the remaining methods, RAD again emerges as the most critical, as removing it leads to divergence. The removal of causal training or LRA results in final errors on the order of $10^{-2}$, suggesting both play meaningful, but secondary, roles.

\subsection{Sine Gordon Equation} \label{sec5.4}

\begin{table}[b!]
	\centering
	\renewcommand{\arraystretch}{1.2}
	\setlength{\tabcolsep}{8pt}
	\small
	\begin{tabular}{ccc}
		\hline
		\textbf{Configuration} & \textbf{Relative $L^2$ Error} & \textbf{Final Loss} \\
		\hline
		$\mathbf{\boldsymbol{\alpha} = 0}$, $\mathbf{\boldsymbol{\beta} = 0}$ & $\mathbf{(3.26 \pm 0.21) \times 10^{-2}}$ & $\mathbf{(1.24 \pm 0.17) \times 10^{-6}}$ \\
		$\alpha = 1$, $\beta = 0$ & $(4.15 \pm 0.51) \times 10^{-2}$ & $(4.15 \pm 1.48) \times 10^{-7}$ \\
		$\alpha = 0$, $\beta = 1$ & $(7.64 \pm 3.61) \times 10^{-2}$ & $(4.28 \pm 0.87) \times 10^{-7}$ \\
		$\alpha = 1$, $\beta = 1$ & $(5.61 \pm 0.84) \times 10^{-2}$ & $(3.28 \pm 3.02) \times 10^{-6}$ \\
		\hline
	\end{tabular}
	\caption{Results for different RGA KAN $(\alpha, \beta)$ initializations on the Sine Gordon equation. Reported values are mean $\pm$ SEM over three seeds. The best performing configuration in terms of relative $L^2$ error is indicated in bold.}
	\label{tab10}
\end{table}

\begin{figure}[b!]
	\centering
	\includegraphics[width=\linewidth]{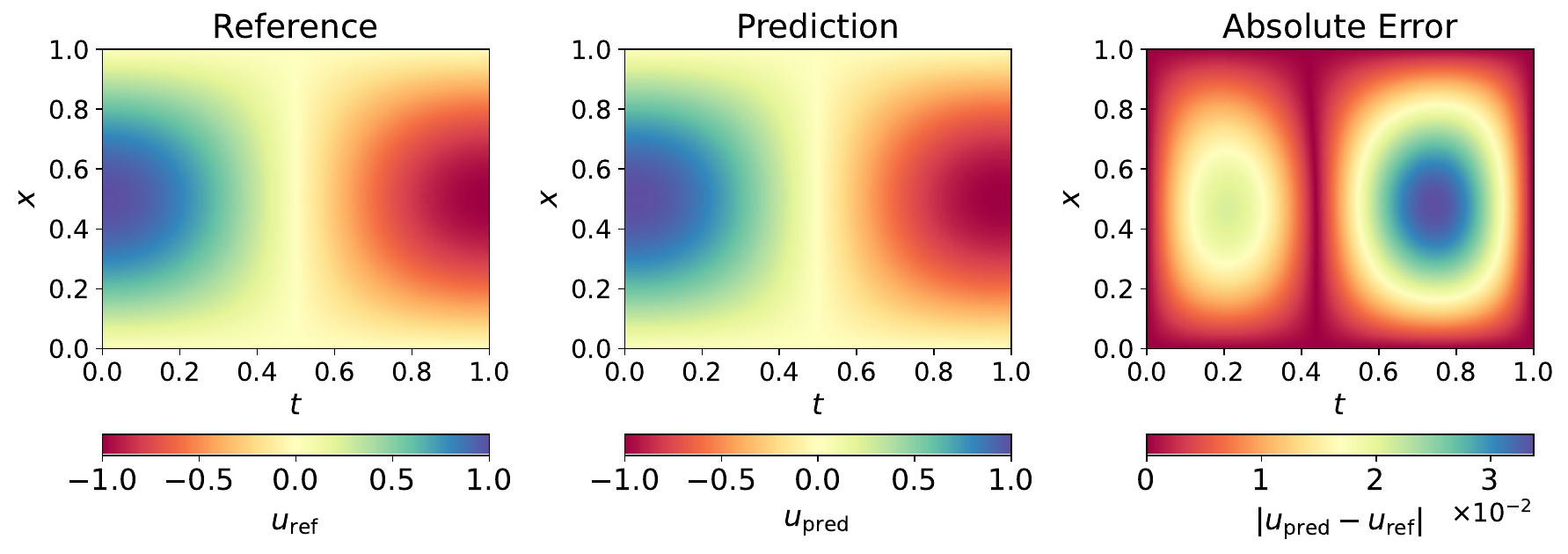}
	\caption{Reference solution (left), RGA KAN prediction (middle) and absolute error (right) for the Sine Gordon equation, shown for the random seed corresponding to the best-performing model instance.}
	\label{fig12}
\end{figure}

We next consider the Sine Gordon equation (see \ref{app:pdes4} for details). As shown in Table \ref{tab10}, the configuration with $\alpha = 0$ and $\beta = 0$ achieves the lowest mean relative $L^2$ error, indicating that initializing the network with an effective depth of a single layer is beneficial in this setting. Among the trained instances, the lowest error achieved for this configuration is $2.84 \times 10^{-2}$. Figure \ref{fig12} shows the corresponding prediction, reference solution and absolute error. Notably, the error grows toward the final stages of the temporal domain, suggesting that a more strict tolerance for causal training or additional training iterations could further improve performance.

The comparative results across architectures are presented in Table \ref{tab11}. Here, the cPIKAN model diverges with a mean relative $L^2$ error around $50\%$. PirateNet performs substantially better but still lags behind RGA KANs, which achieve a lower error compared to PirateNets across all $(\alpha,\beta)$ configurations tested. Moreover, the variability across seeds is noticeably lower for RGA KANs, indicating more consistent performance.

\begin{table}[b!]
	\centering
	\renewcommand{\arraystretch}{1.2}
	\setlength{\tabcolsep}{8pt}
	\small
	\begin{tabular}{cccc}
		\hline
		\textbf{Architecture} & \textbf{Parameters} & \textbf{Relative $L^2$ Error} & \textbf{Time / Iter.} \\
		\hline
		cPIKAN & 18,307 & $(5.01 \pm 0.62) \times 10^{-1}$ & 3.93 ms \\
		PirateNet & 19,228 & $(8.02 \pm 1.70) \times 10^{-2}$ & 3.68 ms \\
		\textbf{RGA KAN} & \textbf{18,422} & $\mathbf{(3.26 \pm 0.21) \times 10^{-2}}$ & \textbf{6.30 ms} \\
		\hline
	\end{tabular}
	\caption{Performance comparison on the Sine Gordon equation across different architectures. Reported values are mean $\pm$ SEM over three random seeds. The RGA KAN row uses the best $(\alpha,\beta)$ initialization from Table \ref{tab10}. The best performing architecture in terms of relative $L^2$ error is indicated in bold.}
	\label{tab11}
\end{table}

\begin{table}[b!]
	\centering
	\renewcommand{\arraystretch}{1.2}
	\setlength{\tabcolsep}{10pt}
	\small
	\begin{tabular}{ccccc}
		\hline
		\textbf{RBA} & \textbf{RAD} & \textbf{Causal} & \textbf{LRA} & \textbf{Relative $L^2$ Error} \\
		\hline
		\cmark & \cmark & \cmark & \cmark & $(3.26 \pm 0.21) \times 10^{-2}$ \\
		\cmark & \xmark & \xmark & \xmark & $(8.85 \pm 0.73) \times 10^{-2}$ \\
		\xmark & \cmark & \cmark & \cmark & $(4.74 \pm 0.96) \times 10^{-2}$ \\
		\xmark & \xmark & \cmark & \cmark & $(4.74 \pm 1.16) \times 10^{-2}$ \\
		\xmark & \cmark & \xmark & \cmark & $(3.46 \pm 0.43) \times 10^{-2}$ \\
		\xmark & \cmark & \cmark & \xmark & $(1.80 \pm 0.20) \times 10^{-1}$ \\
		\hline
	\end{tabular}
	\caption{Ablation study on adaptive training components for the Sine Gordon equation using the best $(\alpha,\beta)$ initialization from Table \ref{tab10}. Each row corresponds to a different combination of enabled (\cmark) or disabled (\xmark) components. Reported values for relative $L^2$ error are mean $\pm$ SEM over three seeds.}
	\label{tab12}
\end{table}

Finally, the ablation results are summarized in Table \ref{tab12}. In this case, all ablations lead to relative $L^2$ errors of the same order of magnitude as the fully adaptive configuration, with the notable exception of removing LRA. In that scenario, the error increases to an average of $1.8 \times 10^{-1}$, highlighting the key role of LRA for this benchmark.

\subsection{Advection Equation} \label{sec5.5}

\begin{table}[b!]
	\centering
	\renewcommand{\arraystretch}{1.2}
	\setlength{\tabcolsep}{8pt}
	\small
	\begin{tabular}{ccc}
		\hline
		\textbf{Configuration} & \textbf{Relative $L^2$ Error} & \textbf{Final Loss} \\
		\hline
		$\alpha = 0$, $\beta = 0$ & $(6.29 \pm 1.02) \times 10^{-4}$ & $(2.46 \pm 0.54) \times 10^{-4}$ \\
		$\alpha = 1$, $\beta = 0$ & $(4.78 \pm 1.18) \times 10^{-4}$ & $(2.40 \pm 0.96) \times 10^{-4}$ \\
		$\alpha = 0$, $\beta = 1$ & $(3.08 \pm 1.98) \times 10^{-3}$ & $(1.74 \pm 1.39) \times 10^{-3}$ \\
		$\mathbf{\boldsymbol{\alpha} = 1}$, $\mathbf{\boldsymbol{\beta} = 1}$ & $\mathbf{(2.41 \pm 0.39) \times 10^{-4}}$ & $\mathbf{(5.89 \pm 3.04) \times 10^{-5}}$ \\
		\hline
	\end{tabular}
	\caption{Results for different RGA KAN $(\alpha, \beta)$ initializations on the advection equation. Reported values are mean $\pm$ SEM over three seeds. The best performing configuration in terms of relative $L^2$ error is indicated in bold.}
	\label{tab13}
\end{table}

\begin{figure}[b!]
	\centering
	\includegraphics[width=\linewidth]{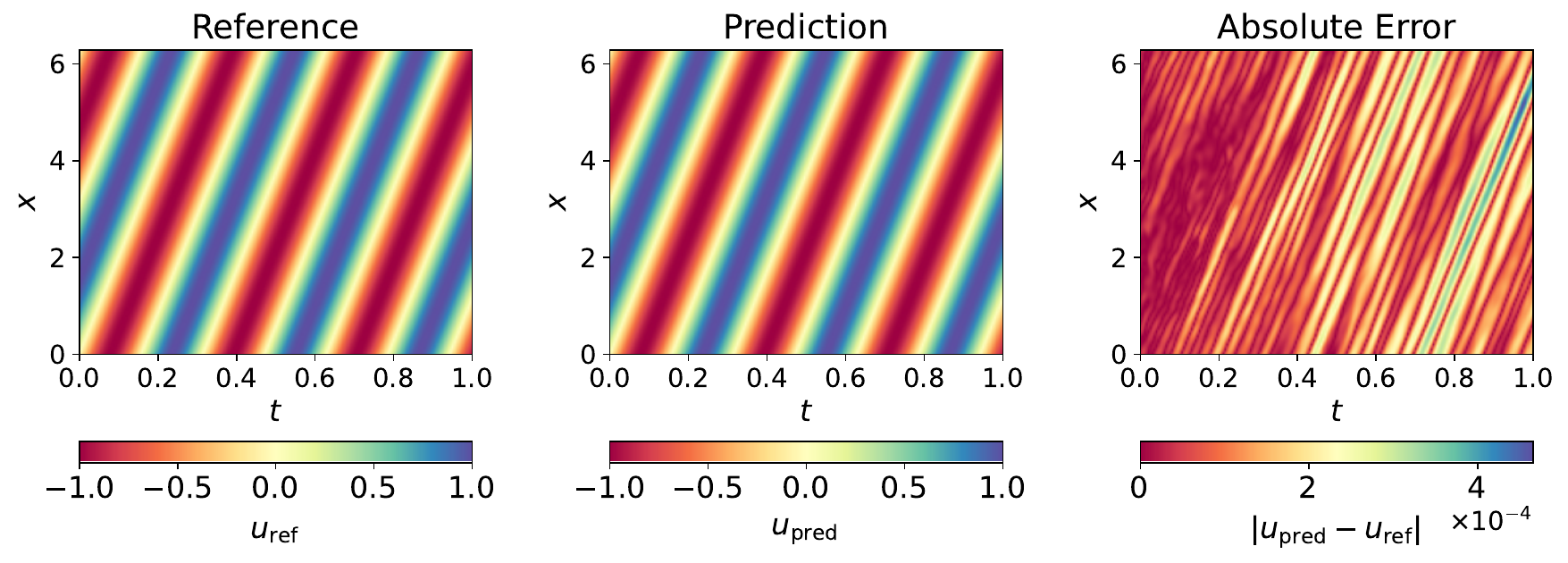}
	\caption{Reference solution (left), RGA KAN prediction (middle) and absolute error (right) for the advection equation, shown for the random seed corresponding to the best-performing model instance.}
	\label{fig13}
\end{figure}

The advection equation (see \ref{app:pdes5} for details), especially at high transport velocities (the constant multiplying the spatial derivative of the solution field in Eq. \eqref{eqD14}), is a challenging benchmark that typically requires specialized training strategies (e.g., learnable spatial periodical embeddings as in \cite{PirateNets}) to obtain accurate solutions. In this work, however, we intentionally refrain from introducing any problem-specific modifications to maintain a unified training pipeline across all PDEs. To this end, we set the advection velocity to $c = 20$, which remains a nontrivial setting. Table \ref{tab13} shows that initializing RGA KANs with $\alpha = 1$ and $\beta = 1$ yields the best performance, with both the lowest mean error and the smallest variability across seeds. Among these runs, the best-performing model achieves a final relative $L^2$ error of $1.81 \times 10^{-4}$, and its prediction is depicted in Figure \ref{fig13} alongside the reference solution and absolute error.

The comparison with the other architectures (Table \ref{tab14}) reveals a stark contrast. Both cPIKAN and PirateNet exhibit poor performance, with large errors and clear indications of instability. While two out of three cPIKAN runs produce moderate errors, the third diverges, resulting in a large mean error and high standard deviation. PirateNets fail consistently across all seeds. In contrast, RGA KANs outperform both baselines by several orders of magnitude, maintaining low errors and small variance, once again showcasing their stability and reliability.

\begin{table}[b!]
	\centering
	\renewcommand{\arraystretch}{1.2}
	\setlength{\tabcolsep}{8pt}
	\small
	\begin{tabular}{cccc}
		\hline
		\textbf{Architecture} & \textbf{Parameters} & \textbf{Relative $L^2$ Error} & \textbf{Time / Iter.} \\
		\hline
		cPIKAN & 18,397 & $(4.11 \pm 4.01) \times 10^{-1}$ & 2.19 ms \\
		PirateNet & 19,246 & $(1.13 \pm 0.13) \times 10^{0}$ & 2.67 ms \\
		\textbf{RGA KAN} & \textbf{18,502} & $\mathbf{(2.41 \pm 0.39) \times 10^{-4}}$ & \textbf{3.98 ms} \\
		\hline
	\end{tabular}
	\caption{Performance comparison on the advection equation across different architectures. Reported values are mean $\pm$ SEM over three random seeds. The RGA KAN row uses the best $(\alpha,\beta)$ initialization from Table \ref{tab13}. The best performing architecture in terms of relative $L^2$ error is indicated in bold.}
	\label{tab14}
\end{table}

\begin{table}[b!]
	\centering
	\renewcommand{\arraystretch}{1.2}
	\setlength{\tabcolsep}{10pt}
	\small
	\begin{tabular}{ccccc}
		\hline
		\textbf{RBA} & \textbf{RAD} & \textbf{Causal} & \textbf{LRA} & \textbf{Relative $L^2$ Error} \\
		\hline
		\cmark & \cmark & \cmark & \cmark & $(2.41 \pm 0.39) \times 10^{-4}$ \\
		\cmark & \xmark & \xmark & \xmark & $(1.01 \pm 0.02) \times 10^{0}$ \\
		\xmark & \cmark & \cmark & \cmark & $(9.22 \pm 4.90) \times 10^{-4}$ \\
		\xmark & \xmark & \cmark & \cmark & $(6.96 \pm 5.31) \times 10^{-3}$ \\
		\xmark & \cmark & \xmark & \cmark & - \\
		\xmark & \cmark & \cmark & \xmark & $(7.29 \pm 3.72) \times 10^{-1}$ \\
		\hline
	\end{tabular}
	\caption{Ablation study on adaptive training components for the advection equation using the best $(\alpha,\beta)$ initialization from Table \ref{tab13}. Each row corresponds to a different combination of enabled (\cmark) or disabled (\xmark) components. Reported values for relative $L^2$ error are mean $\pm$ SEM over three seeds.}
	\label{tab15}
\end{table}

The ablation results in Table \ref{tab15} highlight the importance of adaptive training for this PDE. Training with only RBA leads to divergence, as does removing both RBA and causal training (indicated by the dash in the table), or RBA and LRA. Notably, RAD is the only adaptive strategy whose removal alongside RBA does not cause divergence, although performance is still noticeably degraded. This stands in contrast to previous benchmarks, where RAD often had the most pronounced effect, emphasizing that the relative contribution of each adaptive technique is highly problem dependent and should be evaluated individually for each PDE, as done in this study.

\subsection{Helmholtz Equation} \label{sec5.6}

\begin{table}[b!]
	\centering
	\renewcommand{\arraystretch}{1.2}
	\setlength{\tabcolsep}{8pt}
	\small
	\begin{tabular}{ccc}
		\hline
		\textbf{Configuration} & \textbf{Relative $L^2$ Error} & \textbf{Final Loss} \\
		\hline
		$\alpha = 0$, $\beta = 0$ & $(9.91 \pm 2.28) \times 10^{-5}$ & $(2.63 \pm 0.25) \times 10^{-3}$ \\
		$\alpha = 1$, $\beta = 0$ & $(1.08 \pm 0.36) \times 10^{-4}$ & $(1.47 \pm 0.31) \times 10^{-3}$ \\
		$\alpha = 0$, $\beta = 1$ & $(8.60 \pm 1.78) \times 10^{-5}$ & $(2.20 \pm 0.24) \times 10^{-3}$ \\
		$\mathbf{\boldsymbol{\alpha} = 1}$, $\mathbf{\boldsymbol{\beta} = 1}$ & $\mathbf{(2.44 \pm 0.34) \times 10^{-5}}$ & $\mathbf{(2.84 \pm 1.24) \times 10^{-4}}$ \\
		\hline
	\end{tabular}
	\caption{Results for different RGA KAN $(\alpha, \beta)$ initializations on the Helmholtz equation. Reported values are mean $\pm$ SEM over three seeds. The best performing configuration in terms of relative $L^2$ error is indicated in bold.}
	\label{tab16}
\end{table}

\begin{figure}[b!]
	\centering
	\includegraphics[width=\linewidth]{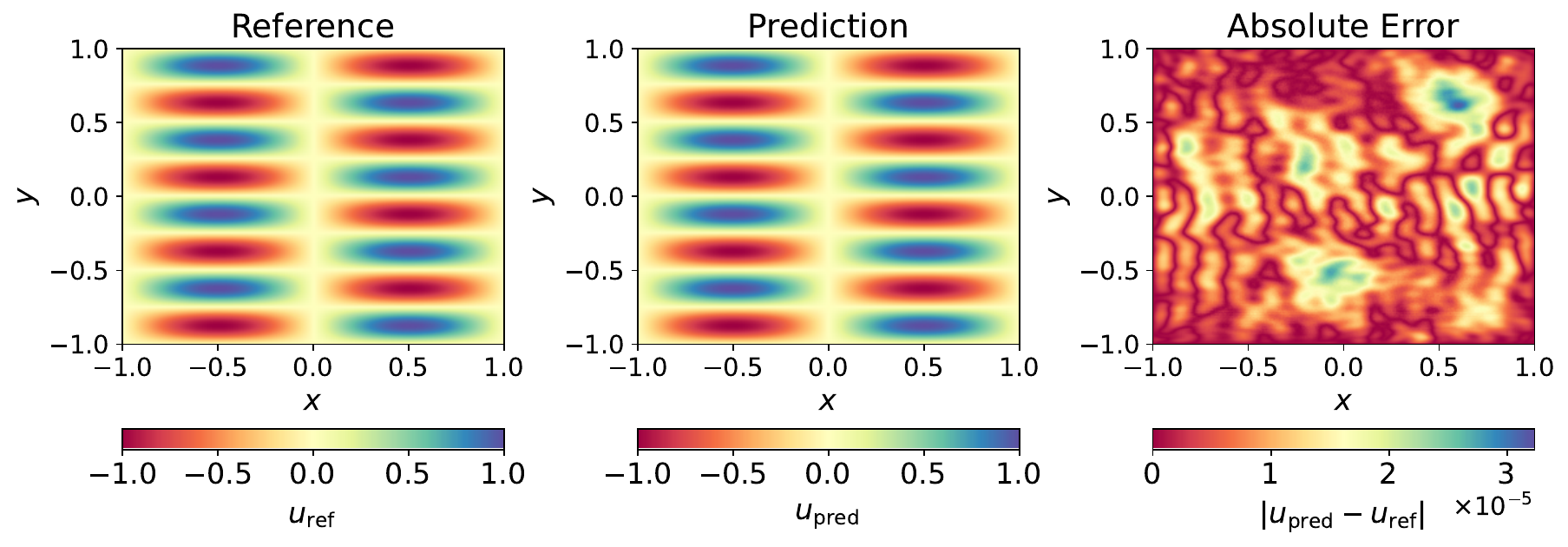}
	\caption{Reference solution (left), RGA KAN prediction (middle) and absolute error (right) for the Helmholtz equation, shown for the random seed corresponding to the best-performing model instance.}
	\label{fig14}
\end{figure}

We next consider the Helmholtz equation (see \ref{app:pdes6} for details), defined solely on a spatial domain and thus involving no initial condition. Consequently, this benchmark excludes both LRA (as the loss contains only one PDE residual term) and causal training (due to the absence of temporal dependencies). Only RBA and RAD are employed as adaptive training strategies. Table \ref{tab16} shows that initializing RGA KANs with $\alpha = 1$ and $\beta = 1$ yields the best performance, achieving both the lowest mean relative $L^2$ error and the smallest variance across seeds. Notably, this configuration also attains the smallest final loss of order $\mathcal{O}(10^{-4})$, which is particularly meaningful for this benchmark: the Helmholtz equation is well known to exhibit a high and challenging loss landscape at initialization \cite{FAIR}. The best-performing model for this configuration achieves a relative $L^2$ error of $1.76 \times 10^{-5}$, and its prediction is shown in Figure \ref{fig14}.

\begin{table}[b!]
	\centering
	\renewcommand{\arraystretch}{1.2}
	\setlength{\tabcolsep}{8pt}
	\small
	\begin{tabular}{cccc}
		\hline
		\textbf{Architecture} & \textbf{Parameters} & \textbf{Relative $L^2$ Error} & \textbf{Time / Iter.} \\
		\hline
		cPIKAN & 18,307 & $(1.03 \pm 0.21) \times 10^{-3}$ & 3.25 ms \\
		PirateNet & 19,230 & $(1.89 \pm 0.25) \times 10^{-4}$ & 3.05 ms \\
		\textbf{RGA KAN} & \textbf{18,423} & $\mathbf{(2.44 \pm 0.34) \times 10^{-5}}$ & \textbf{4.30 ms} \\
		\hline
	\end{tabular}
	\caption{Performance comparison on the Helmholtz equation across different architectures. Reported values are mean $\pm$ SEM over three random seeds. The RGA KAN row uses the best $(\alpha,\beta)$ initialization from Table \ref{tab16}. The best performing architecture in terms of relative $L^2$ error is indicated in bold.}
	\label{tab17}
\end{table}

\begin{table}[b!]
	\centering
	\renewcommand{\arraystretch}{1.2}
	\setlength{\tabcolsep}{10pt}
	\small
	\begin{tabular}{ccc}
		\hline
		\textbf{RBA} & \textbf{RAD} & \textbf{Relative $L^2$ Error} \\
		\hline
		\cmark & \cmark & $(2.44 \pm 0.34) \times 10^{-5}$ \\
		\cmark & \xmark & $(1.61 \pm 1.05) \times 10^{-3}$ \\
		\xmark & \cmark & $(6.81 \pm 1.82) \times 10^{-5}$ \\
		\xmark & \xmark & $(1.95 \pm 0.54) \times 10^{-4}$ \\
		\hline
	\end{tabular}
	\caption{Ablation study on adaptive training components for the Helmholtz equation using the best $(\alpha,\beta)$ initialization from Table \ref{tab16}. Each row corresponds to a different combination of enabled (\cmark) or disabled (\xmark) components. Reported values for relative $L^2$ error are mean $\pm$ SEM over three seeds.}
	\label{tab18}
\end{table}

In the cross-architecture comparison (Table \ref{tab17}), all three architectures converge, but with clearly different accuracies. cPIKANs achieve a relative $L^2$ error of order $\mathcal{O}(10^{-3})$, PirateNets improve by one order of magnitude, and RGA KANs outperform both by another order of magnitude, reaching the $\mathcal{O}\left(10^{-5}\right)$ range. The variance across seeds remains low for all three architectures.

The ablation study (Table \ref{tab18}) provides interesting insights for the training dynamics of the models trained for this PDE. Unlike in the case of the Korteweg--De Vries or the advection equation, the RGA KAN architecture converges even without any adaptive training. In fact, the performance in this setting is comparable to that of PirateNets with both adaptive strategies active. Another notable observation is that disabling only RAD leads to higher errors than disabling both RBA and RAD. This suggests that the interaction between adaptive components can be nontrivial: while RBA combined with RAD yields the best performance, using RBA alone may actually be suboptimal for this PDE under the chosen hyperparameter configuration.

\subsection{Poisson Equation} \label{sec5.7}

\begin{table}[b!]
	\centering
	\renewcommand{\arraystretch}{1.5}
	\setlength{\tabcolsep}{6pt}
	\scriptsize
	\begin{tabular}{c|c|ccc}
		\hline
		$(\boldsymbol{\alpha}, \boldsymbol{\beta})$ & \textbf{Metric} & $\boldsymbol{\omega=1}$ & $\boldsymbol{\omega=2}$ & $\boldsymbol{\omega=4}$ \\
		\hline
		\multirow{2}{*}{$(0,0)$}
		& Rel. $L^2$ Error & $(8.95 \pm 1.97)\!\times\!10^{-6}$ & $(2.92 \pm 2.20)\!\times\!10^{-4}$ & $(2.15 \pm 1.09)\!\times\!10^{-2}$ \\
		& Final Loss & $(1.98 \pm 0.08)\!\times\!10^{-6}$ & $(3.35 \pm 0.69)\!\times\!10^{-4}$ & $(1.30 \pm 0.07)\!\times\!10^{-1}$ \\
		\hline
		\multirow{2}{*}{$(1,0)$}
		& Rel. $L^2$ Error & $(2.03 \pm 0.70)\!\times\!10^{-6}$ & $\mathbf{(4.53 \pm 0.90)\!\times\!10^{-5}}$ & $(1.93 \pm 0.83)\!\times\!10^{-2}$ \\
		& Final Loss & $(7.08 \pm 0.31)\!\times\!10^{-7}$ & $\mathbf{(9.30 \pm 1.29)\!\times\!10^{-5}}$ & $(1.44 \pm 0.59)\!\times\!10^{-1}$ \\
		\hline
		\multirow{2}{*}{$(0,1)$}
		& Rel. $L^2$ Error & $(3.33 \pm 1.58)\!\times\!10^{-6}$ & $(6.42 \pm 0.34)\!\times\!10^{-5}$ & $\mathbf{(9.67 \pm 5.80)\!\times\!10^{-3}}$ \\
		& Final Loss & $(7.35 \pm 0.34)\!\times\!10^{-7}$ & $(2.69 \pm 0.58)\!\times\!10^{-4}$ & $\mathbf{(9.40 \pm 3.03)\!\times\!10^{-2}}$ \\
		\hline
		\multirow{2}{*}{$(1,1)$}
		& Rel. $L^2$ Error & $\mathbf{(1.10 \pm 0.26)\!\times\!10^{-6}}$ & $(9.24 \pm 2.05)\!\times\!10^{-5}$ & $(2.12 \pm 0.94)\!\times\!10^{-2}$ \\
		& Final Loss & $\mathbf{(5.66 \pm 0.18)\!\times\!10^{-7}}$ & $(2.00 \pm 0.05)\!\times\!10^{-5}$ & $(2.48 \pm 0.80)\!\times\!10^{-2}$ \\
		\hline
	\end{tabular}
	\caption{Results for different RGA KAN $(\alpha,\beta)$ initializations on the Poisson equation for $\omega\in\{1,2,4\}$. Reported values are mean $\pm$ SEM over three seeds. The best performing configuration in terms of relative $L^2$ error per column is indicated in bold.}
	\label{tab19}
\end{table}

\begin{figure}[t!]
	\centering
	\includegraphics[width=\linewidth]{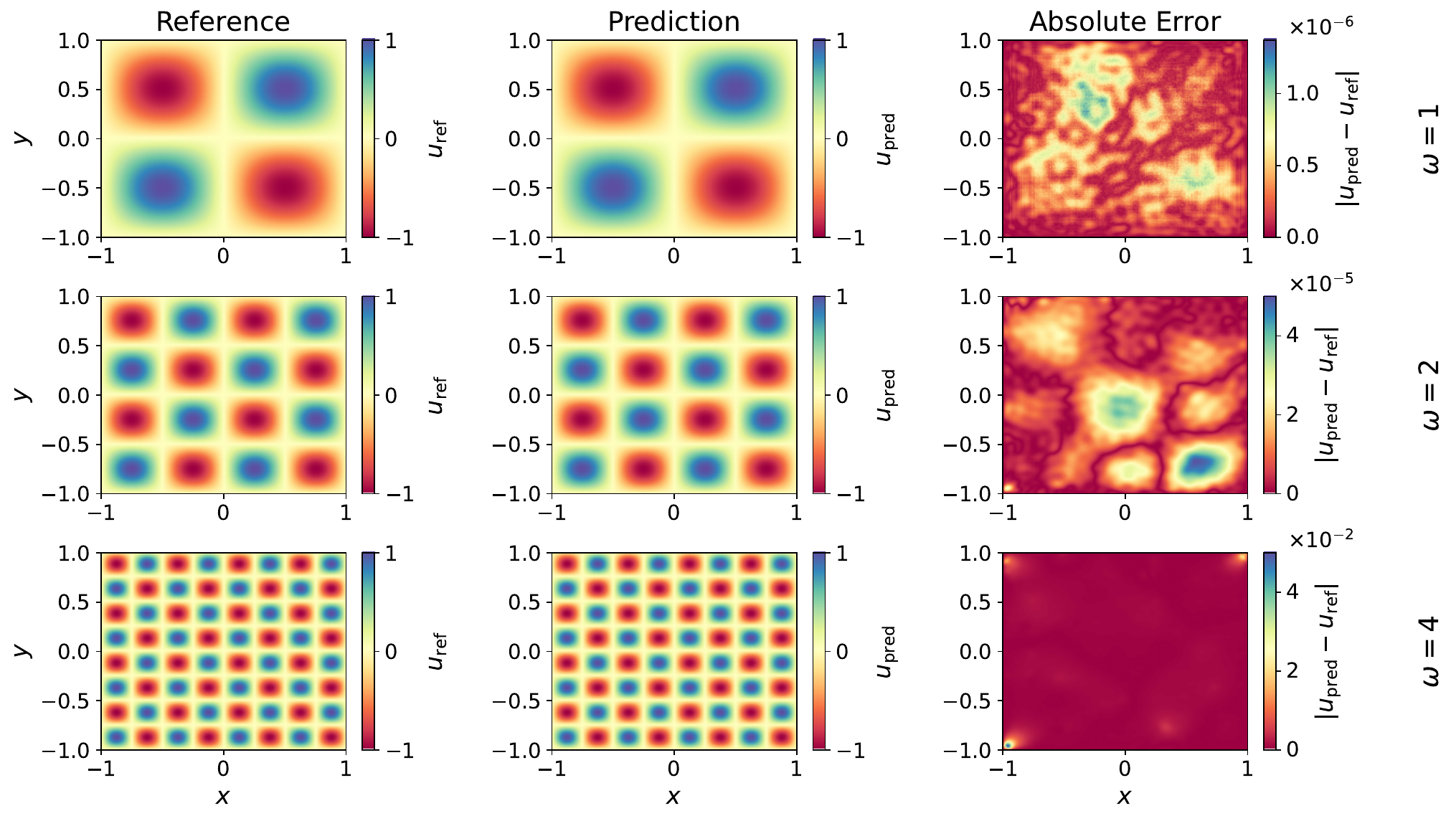}
	\caption{Reference solution (left column), RGA KAN prediction (middle column) and absolute error (right column) for the Poisson equation, shown for the random seed corresponding to the best-performing model instance per value of $\omega \in \{1,2,4\}$.}
	\label{fig15}
\end{figure}

The next benchmark is the Poisson equation (see \ref{app:pdes7} for details), which -- similarly to the Helmholtz equation -- is defined on a purely spatial domain and therefore involves no initial condition. As a result, only RBA and RAD are employed as adaptive training strategies. The source term is chosen such that the analytical solution is $u(x,y) = \sin(\pi \omega x)\sin(\pi \omega y)$, allowing us to systematically investigate how performance degrades as the frequency parameter $\omega$ increases. We consider $\omega \in \{1,2,4\}$ and Table \ref{tab19} presents the results of the $(\alpha,\beta)$ initialization study for each $\omega$. Evidently, different initialization configurations are optimal for different frequencies, and performance deteriorates with increasing $\omega$. Moreover, the SEM for the optimal configuration grows alongside $\omega$: for $\omega=1$, it is roughly an order of magnitude lower than the mean error, whereas for $\omega=4$ it exceeds half of it, indicating increased sensitivity to initialization and optimization. For each $\omega$ value, we select the best-performing seed of the optimal configuration, achieving relative $L^2$ errors of $7.33 \times 10^{-7}$, $2.74 \times 10^{-5}$, and $3.34 \times 10^{-3}$ for $\omega = 1, 2$, and $4$, respectively. Figure \ref{fig15} shows the reference solution, prediction and absolute error for these runs.

\begin{table}[b!]
\centering
\renewcommand{\arraystretch}{1.2}
\setlength{\tabcolsep}{8pt}
\small
\begin{tabular}{ccc}
	\hline
	\textbf{Architecture} & \textbf{Parameters} & \textbf{Time / Iter.} \\
	\hline
	cPIKAN    &	18,307	&  2.93 ms \\
	PirateNet &	19,230	&  2.94 ms \\
	RGA KAN   &	18,423	&  3.81 ms \\
	\hline
\end{tabular}
\caption{Number of parameters and average time per training iteration (milliseconds) for each architecture on the Poisson equation.}
\label{tab20}
\end{table}

We next compare architectures using the best $(\alpha,\beta)$ initialization for each $\omega$. The parameter counts and iteration times are shown in Table \ref{tab20}, while the accuracy results in terms of relative $L^2$ error are summarized in Table \ref{tab21}. For $\omega = 1$ and $\omega = 2$, all architectures converge, with RGA KANs achieving the lowest error by a wide margin. However, for $\omega = 4$ the situation changes: PirateNets diverge, while, interestingly, for the first time across all benchmarks, cPIKANs outperform RGA KANs under the default training settings, although both achieve errors of the same order of magnitude.

\begin{table}[t!]
	\centering
	\renewcommand{\arraystretch}{1.5}
	\setlength{\tabcolsep}{6pt}
	\footnotesize
	\begin{tabular}{cccc}
		\hline
		\textbf{Architecture} & $\boldsymbol{\omega=1}$ & $\boldsymbol{\omega=2}$ & $\boldsymbol{\omega=4}$ \\
		\hline
		cPIKAN     & $(1.56 \pm 0.48)\times 10^{-5}$ & $(2.15 \pm 0.99)\times 10^{-3}$ & $\mathbf{(5.08 \pm 1.41)\times 10^{-3}}$ \\
		PirateNet  & $(7.66 \pm 2.72)\times 10^{-6}$ & $(1.58 \pm 0.47)\times 10^{-4}$ & $(2.57 \pm 1.86)\times 10^{0}$ \\
		RGA KAN    & $\mathbf{(1.10 \pm 0.26)\times 10^{-6}}$ & $\mathbf{(4.53 \pm 0.90)\!\times\!10^{-5}}$ & $(9.67 \pm 5.80)\times 10^{-3}$ \\
		\hline
	\end{tabular}
	\caption{Performance comparison on the Poisson equation across different architectures. Reported values are mean $\pm$ SEM over three random seeds for $\omega\in\{1,2,4\}$. The RGA KAN row uses the best $(\alpha,\beta)$ initialization from Table \ref{tab19}. The best performing architecture in terms of relative $L^2$ error per column is indicated in bold.}
	\label{tab21}
\end{table}

The ablation study (Table \ref{tab22}) provides additional insight. For $\omega = 1$ and $\omega = 2$, RGA KANs converge even without adaptive training, albeit with somewhat higher errors. For $\omega = 4$, however, performance collapses without adaptive methods. Remarkably, when RBA is disabled but RAD is retained, the error decreases by nearly an order of magnitude compared to the non-adaptive case, with very low variance across runs. In fact, this configuration outperforms the average error achieved by cPIKANs at $\omega = 4$, effectively restoring the advantage of RGA KANs for this challenging regime. This observation is consistent with the findings for the Helmholtz equation, which is structurally similar to the Poisson equation, where RAD also emerged as the dominant adaptive component.

\begin{table}[h!]
	\centering
	\renewcommand{\arraystretch}{1.5}
	\setlength{\tabcolsep}{8pt}
	\footnotesize
	\begin{tabular}{ccccc}
		\hline
		\textbf{RBA} & \textbf{RAD} & \(\boldsymbol{\omega=1}\) & \(\boldsymbol{\omega=2}\) & \(\boldsymbol{\omega=4}\) \\
		\hline
		\cmark & \cmark & $(1.10 \pm 0.26)\!\times\!10^{-6}$ & $(4.53 \pm 0.90)\!\times\!10^{-5}$ & $(9.67 \pm 5.80)\!\times\!10^{-3}$ \\
		\cmark & \xmark & $(1.95 \pm 0.12)\!\times\!10^{-6}$ & $(7.97 \pm 1.29)\!\times\!10^{-5}$ & $(8.01 \pm 3.72)\!\times\!10^{-2}$ \\
		\xmark & \cmark & $(1.04 \pm 0.73)\!\times\!10^{-5}$ & $(5.01 \pm 0.27)\!\times\!10^{-5}$ & $(1.19 \pm 0.25)\!\times\!10^{-3}$ \\
		\xmark & \xmark & $(2.54 \pm 1.31)\!\times\!10^{-5}$ & $(5.09 \pm 0.62)\!\times\!10^{-5}$ & $(1.23 \pm 0.79)\!\times\!10^{-1}$ \\
		\hline
	\end{tabular}
	\caption{Ablation study on adaptive training components for the Poisson equation using the best $(\alpha,\beta)$ initialization from Table \ref{tab19}. Each row corresponds to a different combination of enabled (\cmark) or disabled (\xmark) components. Reported values for relative $L^2$ error are mean $\pm$ SEM over three seeds for $\omega\in\{1,2,4\}$.}
	\label{tab22}
\end{table}

\subsection{Heat Equation} \label{sec5.8}

\begin{figure}[t!]
	\centering
	\includegraphics[width=\linewidth]{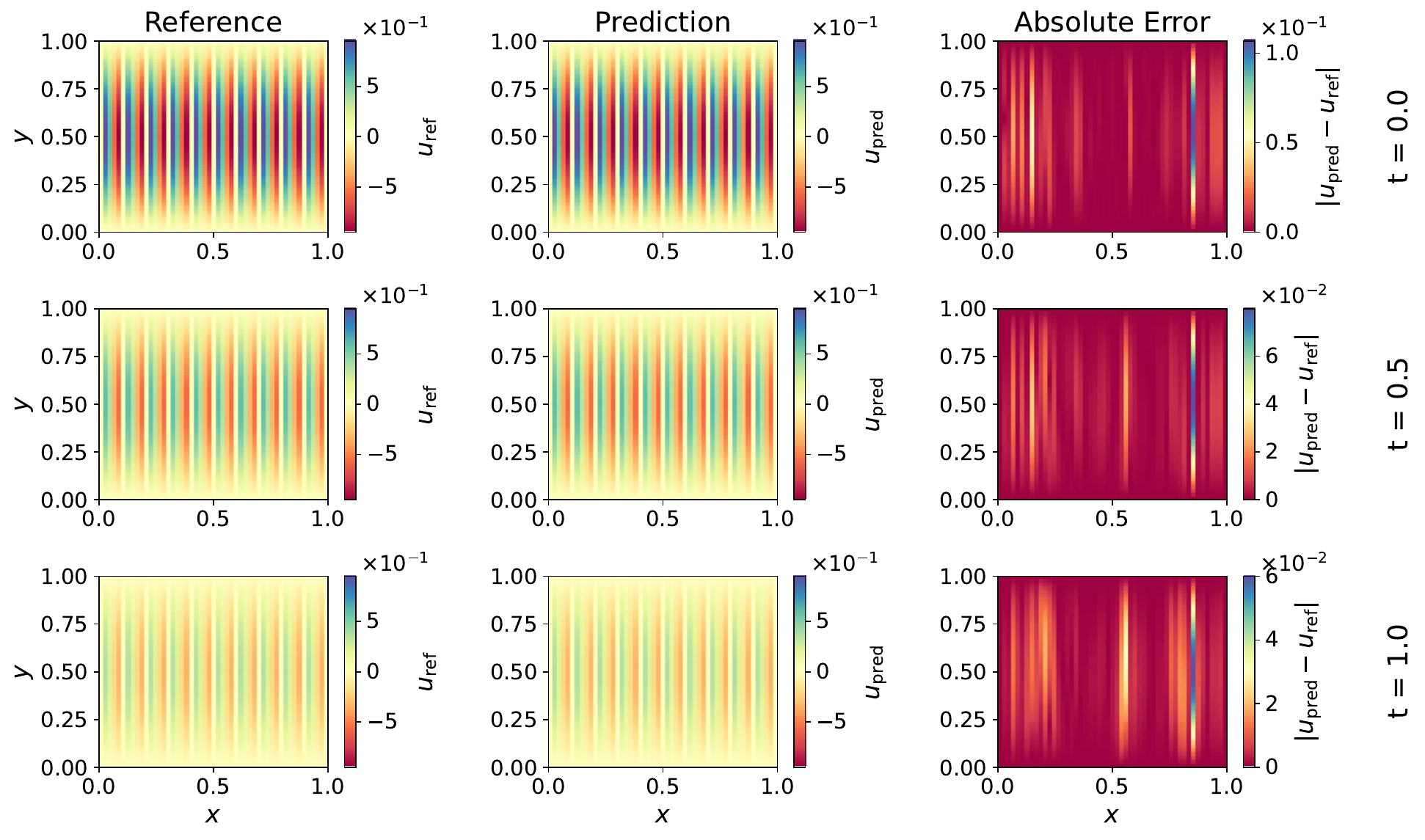}
	\caption{Reference solution (left column), RGA KAN prediction (middle column) and absolute error (right column) for the heat equation, shown for the random seed corresponding to the best-performing model instance at three different snapshots: $t = 0$ (top row), $t = 0.5$ (middle row), $t = 1.0$ (bottom row).}
	\label{fig16}
\end{figure}

For the subsequent benchmark we choose an equation with higher dimensionality, namely the (2+1)-dimensional heat equation (see \ref{app:pdes8} for details). As indicated in Table \ref{tab23}, the initialization with $\alpha = 0$ and $\beta = 1$ yields the best performance, achieving the lowest mean relative $L^2$ error. The best individual run under this configuration achieved a relative $L^2$ error of $3.11 \times 10^{-2}$. Figure \ref{fig16} visualizes the model's predictions at three temporal snapshots: $t=0$, $t=0.5$ and $t=1.0$. The absolute error field at $t=0$ reveals a localized difficulty in fitting the initial condition around $x \approx 0.85$, $y \in (0.25, 0.75)$. This propagates into subsequent time instances, although it remains bounded. Despite this local artifact, the global dynamics are captured accurately, as the results demonstrate a very good overall agreement with the reference solution.

\begin{table}[t!]
	\centering
	\renewcommand{\arraystretch}{1.2}
	\setlength{\tabcolsep}{8pt}
	\small
	\begin{tabular}{ccc}
		\hline
		\textbf{Configuration} & \textbf{Relative $L^2$ Error} & \textbf{Final Loss} \\
		\hline
		$\alpha = 0$, $\beta = 0$ & $(1.14 \pm 0.26) \times 10^{-1}$ & $(5.46 \pm 2.33) \times 10^{-5}$ \\
		$\alpha = 1$, $\beta = 0$ & $(4.91 \pm 0.77) \times 10^{-2}$ & $(2.71 \pm 0.31) \times 10^{-5}$ \\
		$\mathbf{\boldsymbol{\alpha} = 0}$, $\mathbf{\boldsymbol{\beta} = 1}$ & $\mathbf{(3.73 \pm 0.36) \times 10^{-2}}$ & $\mathbf{(2.57 \pm 0.32) \times 10^{-5}}$ \\
		$\alpha = 1$, $\beta = 1$ & $(6.44 \pm 1.39) \times 10^{-2}$ & $(3.78 \pm 2.93) \times 10^{-5}$ \\
		\hline
	\end{tabular}
	\caption{Results for different RGA KAN $(\alpha, \beta)$ initializations on the heat equation. Reported values are mean $\pm$ SEM over three seeds. The best performing configuration in terms of relative $L^2$ error is indicated in bold.}
	\label{tab23}
\end{table}

The comparative results across architectures are presented in Table \ref{tab24}. Consistent with previous benchmarks, the cPIKAN model struggles to learn the solution, exhibiting a high mean relative error and significant variance. PirateNet offers a marked improvement over cPIKAN but fails to match the precision of the proposed method. RGA KANs achieve the lowest error and demonstrate superior consistency, with a SEM that is substantially lower than that of the baseline architectures.

\begin{table}[t!]
	\centering
	\renewcommand{\arraystretch}{1.2}
	\setlength{\tabcolsep}{8pt}
	\small
	\begin{tabular}{cccc}
		\hline
		\textbf{Architecture} & \textbf{Parameters} & \textbf{Relative $L^2$ Error} & \textbf{Time / Iter.} \\
		\hline
		cPIKAN & 18,397 & $(5.12 \pm 2.52) \times 10^{-1}$ & 3.60 ms \\
		PirateNet & 19,246 & $(1.78 \pm 0.82) \times 10^{-1}$ & 3.78 ms \\
		\textbf{RGA KAN} & \textbf{18,502} & $\mathbf{(3.73 \pm 0.36) \times 10^{-2}}$ & \textbf{5.52 ms} \\
		\hline
	\end{tabular}
	\caption{Performance comparison on the heat equation across different architectures. Reported values are mean $\pm$ SEM over three random seeds. The RGA KAN row uses the best $(\alpha,\beta)$ initialization from Table \ref{tab23}. The best performing architecture in terms of relative $L^2$ error is indicated in bold.}
	\label{tab24}
\end{table}

\begin{table}[b!]
	\centering
	\renewcommand{\arraystretch}{1.2}
	\setlength{\tabcolsep}{10pt}
	\small
	\begin{tabular}{ccccc}
		\hline
		\textbf{RBA} & \textbf{RAD} & \textbf{Causal} & \textbf{LRA} & \textbf{Relative $L^2$ Error} \\
		\hline
		\cmark & \cmark & \cmark & \cmark & $(3.73 \pm 0.36) \times 10^{-2}$ \\
		\cmark & \xmark & \xmark & \xmark & $(3.23 \pm 0.94) \times 10^{-2}$ \\
		\xmark & \cmark & \cmark & \cmark & $(7.92 \pm 1.01) \times 10^{-2}$ \\
		\xmark & \xmark & \cmark & \cmark & $(8.68 \pm 1.65) \times 10^{-2}$ \\
		\xmark & \cmark & \xmark & \cmark & $(1.13 \pm 0.11) \times 10^{-1}$ \\
		\xmark & \cmark & \cmark & \xmark & $(1.05 \pm 0.18) \times 10^{-1}$ \\
		\hline
	\end{tabular}
	\caption{Ablation study on adaptive training components for the heat equation using the best $(\alpha,\beta)$ initialization from Table \ref{tab23}. Each row corresponds to a different combination of enabled (\cmark) or disabled (\xmark) components. Reported values for relative $L^2$ error are mean $\pm$ SEM over three seeds.}
	\label{tab25}
\end{table}

Finally, Table \ref{tab25} summarizes the ablation study for this benchmark. A notable observation here is the critical role of RBA: the configuration using only RBA performs competitively with the fully adaptive setup, yielding a slightly lower mean error but with higher variance. Conversely, disabling RBA roughly doubles the error. Further removing either the causal training or LRA mechanisms leads to a significant degradation in performance, increasing the error by an order of magnitude.

\subsection{Navier-Stokes Equation} \label{sec5.9}

The final benchmark is the Navier-Stokes equation (see \ref{app:pdes9} for details), modeling viscous fluid dynamics on a torus. Following established practices in the literature \cite{jaxpi}, we introduce a weighting factor of $100$ for the continuity equation residual and initialize the initial condition weight $\lambda_\text{ic}$ to $10^5$ to prioritize the fitting of the initial snapshot. We adopt these settings as the default for our experiments to ensure a fair comparison with baselines, though we explicitly investigate their necessity in our ablation study.

\begin{table}[b!]
	\centering
	\renewcommand{\arraystretch}{1.2}
	\setlength{\tabcolsep}{8pt}
	\small
	\begin{tabular}{cclc}
		\hline
		\textbf{Configuration} & \multicolumn{2}{c}{\textbf{Relative $L^2$ Error}} & \textbf{Final Loss} \\
		\hline
		\multirow{3}{*}{$\alpha = 0$, $\beta = 0$} & $u$: & $(1.08 \pm 0.40) \times 10^{-2}$ & \multirow{3}{*}{$(5.14 \pm 0.20) \times 10^{-2}$} \\
		& $v$: & $(7.01 \pm 2.26) \times 10^{-3}$ & \\
		& $w$: & $(1.54 \pm 0.15) \times 10^{-3}$ & \\
		\cline{1-4}
		\multirow{3}{*}{$\alpha = 1$, $\beta = 0$} & $u$: & $\mathbf{(8.90 \pm 3.44) \times 10^{-3}}$ & \multirow{3}{*}{$(6.18 \pm 0.16) \times 10^{-2}$} \\
		& $v$: & $\mathbf{(4.78 \pm 1.16) \times 10^{-3}}$ & \\
		& $w$: & $(1.86 \pm 0.25) \times 10^{-3}$ & \\
		\cline{1-4}
		\multirow{3}{*}{$\alpha = 0$, $\beta = 1$} & $u$: & $(1.33 \pm 0.50) \times 10^{-2}$ & \multirow{3}{*}{$(2.81 \pm 0.08) \times 10^{-2}$} \\
		& $v$: & $(8.37 \pm 3.04) \times 10^{-3}$ & \\
		& $w$: & $\mathbf{(1.60 \pm 0.23) \times 10^{-3}}$ & \\
		\cline{1-4}
		\multirow{3}{*}{$\alpha = 1$, $\beta = 1$} & $u$: & $(1.51 \pm 0.51) \times 10^{-2}$ & \multirow{3}{*}{$(2.81 \pm 0.08) \times 10^{-2}$} \\
		& $v$: & $(8.11 \pm 2.49) \times 10^{-3}$ & \\
		& $w$: & $(7.73 \pm 2.52) \times 10^{-3}$ & \\
		\hline
	\end{tabular}
	\caption{Results for different RGA KAN $(\alpha, \beta)$ initializations on the Navier-Stokes equation. Reported values are mean $\pm$ SEM over three seeds. Errors are reported separately for velocity components ($u, v$) and vorticity ($w$). The best performance per field is indicated in bold.}
	\label{tab26}
\end{table}

\begin{figure}[t!]
	\centering
	\includegraphics[width=\linewidth]{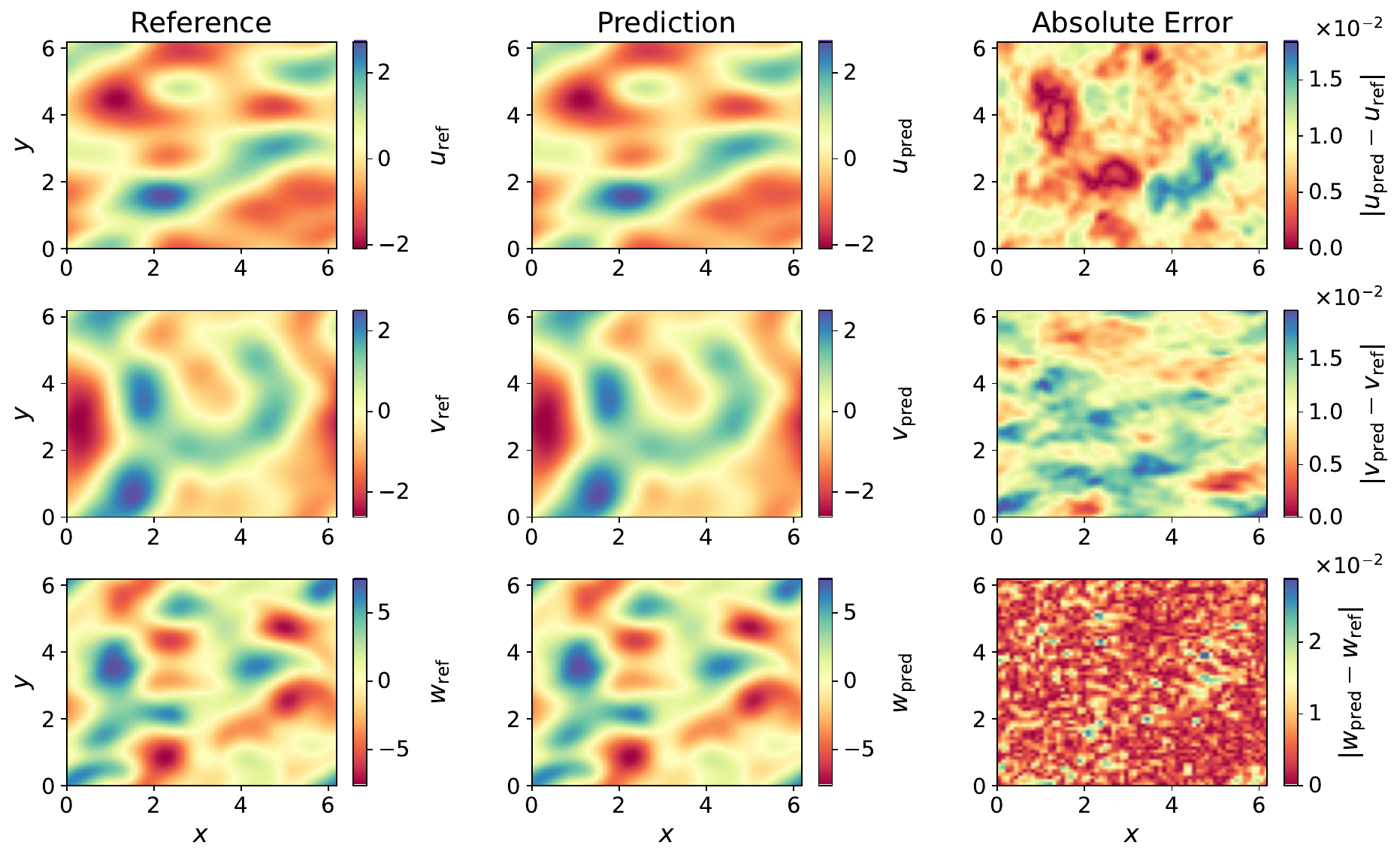}
	\caption{Reference solution (left column), RGA KAN prediction (middle column) and absolute error (right column) for the solutions to the Navier-Stokes equation, shown for the random seed corresponding to the best-performing model instance at the final temporal snapshot $t=0.1$. The top, middle and bottom row corresponds to $u$, $v$ and $w$, respectively.}
	\label{fig17}
\end{figure}

We first examine the impact of initialization on model performance. While the vorticity field ($w$) is the standard metric reported for this benchmark, we also provide errors for the velocity components ($u, v$) in Table \ref{tab26} for completeness. The results show that while the configuration $\alpha = 0, \beta = 1$ yields the lowest error for vorticity, the configuration $\alpha = 1, \beta = 0$ achieves significantly better accuracy for the velocity fields $u$ and $v$ with only a marginal trade-off in vorticity performance. Consequently, we select $\alpha = 1, \beta = 0$ as the optimal configuration for the subsequent comparisons. Figure \ref{fig17} visualizes the predictions for this model at $t=0.1$, for a run achieving relative $L^2$ errors of $6.02 \times 10^{-3}$ for $u$, $4.93 \times 10^{-3}$ for $v$, and $1.63 \times 10^{-3}$ for $w$.

Table \ref{tab27} compares the RGA KAN against the cPIKAN and PirateNet baselines. A notable observation here is the shift in computational cost compared to previous benchmarks. While cPIKAN remains the fastest, the training times for PirateNet and RGA KAN are now comparable. This is due to the heavy computational load of the Navier-Stokes loss function, which involves multiple high-order derivatives. In this regime, the gating mechanisms shared by both PirateNet and RGA KAN become the primary bottleneck, rendering the difference between computing KAN basis functions and MLP operations less significant. Despite the similar cost, the performance gap is substantial: RGA KAN outperforms PirateNet by an order of magnitude on the velocity fields and achieves a vorticity error nearly three times lower.

\begin{table}[t!]
	\centering
	\renewcommand{\arraystretch}{1.2}
	\setlength{\tabcolsep}{8pt}
	\small
	\begin{tabular}{ccclc}
		\hline
		\textbf{Architecture} & \textbf{Params} & \multicolumn{2}{c}{\textbf{Relative $L^2$ Error}} & \textbf{Time / Iter.} \\
		\hline
		\multirow{3}{*}{cPIKAN} & \multirow{3}{*}{18,578} & $u$: & $(3.05 \pm 0.40) \times 10^{-2}$ & \multirow{3}{*}{15.88 ms} \\
		& & $v$: & $(2.87 \pm 0.32) \times 10^{-2}$ & \\
		& & $w$: & $(9.25 \pm 0.66) \times 10^{-3}$ & \\
		\cline{1-5}
		\multirow{3}{*}{PirateNet} & \multirow{3}{*}{19,300} & $u$: & $(5.00 \pm 1.22) \times 10^{-2}$ & \multirow{3}{*}{24.85 ms} \\
		& & $v$: & $(2.31 \pm 0.12) \times 10^{-2}$ & \\
		& & $w$: & $(4.19 \pm 0.73) \times 10^{-3}$ & \\
		\cline{1-5}
		\multirow{3}{*}{\textbf{RGA KAN}} & \multirow{3}{*}{\textbf{18,662}} & $u$: & $\mathbf{(8.90 \pm 3.44) \times 10^{-3}}$ & \multirow{3}{*}{\textbf{26.40 ms}} \\
		& & $v$: & $\mathbf{(4.78 \pm 1.16) \times 10^{-3}}$ & \\
		& & $w$: & $\mathbf{(1.86 \pm 0.25) \times 10^{-3}}$ & \\
		\hline
	\end{tabular}
	\caption{Performance comparison on the Navier-Stokes equation across different architectures. Reported values are mean $\pm$ SEM over three random seeds. The RGA KAN row uses $(\alpha,\beta) = (1.0,0.0)$. The best performing architecture in terms of relative $L^2$ error is indicated in bold.}
	\label{tab27}
\end{table}

\begin{table}[b!]
	\centering
	\renewcommand{\arraystretch}{1.2}
	\setlength{\tabcolsep}{10pt}
	\small
	\begin{tabular}{cccccc}
		\hline
		\textbf{Weights} & \textbf{RBA} & \textbf{RAD} & \textbf{Causal} & \textbf{LRA} & \textbf{Relative $L^2$ Error} \\
		\hline
		\cmark & \cmark & \cmark & \cmark & \cmark & $(1.86 \pm 0.25) \times 10^{-3}$ \\
		\xmark & \cmark & \cmark & \cmark & \cmark & $(2.96 \pm 0.18) \times 10^{-3}$ \\
		\cmark & \cmark & \xmark & \xmark & \xmark & $(1.48 \pm 0.19) \times 10^{-2}$ \\
		\cmark & \xmark & \cmark & \cmark & \cmark & $(1.58 \pm 0.08) \times 10^{-3}$ \\
		\cmark & \xmark & \xmark & \cmark & \cmark & $(7.69 \pm 0.75) \times 10^{-3}$ \\
		\cmark & \xmark & \cmark & \xmark & \cmark & $(1.64 \pm 0.11) \times 10^{-3}$ \\
		\cmark & \xmark & \cmark & \cmark & \xmark & $(6.57 \pm 0.27) \times 10^{-2}$ \\
		\hline
	\end{tabular}
	\caption{Ablation study on adaptive training components for the Navier-Stokes equation using $(\alpha,\beta) = (1.0,0.0)$. Each row corresponds to a different combination of enabled (\cmark) or disabled (\xmark) components. Reported values for relative $L^2$ error of the vorticity field ($w$) are mean $\pm$ SEM over three seeds.}
	\label{tab28}
\end{table}

Finally, we extend the ablation study in Table \ref{tab28} to investigate the role of the manual loss weights. We find that removing these heuristic weights (``Weights'' column) increases the error, confirming their utility; however, the model does not diverge. Crucially, even without manual weighting, the RGA KAN yields a lower error than the weighted cPIKAN and PirateNet models. Regarding the adaptive components, we observe a distinct behavior compared to, e.g., the heat equation: here, RBA appears to hinder fine-scale convergence. Removing RBA leads to the lowest overall error, whereas relying on RBA alone (row 3) degrades performance by an order of magnitude.

\section{Conclusion and Outlook} \label{sec6}

In this work, we studied the deep training of Chebyshev-based KANs with the goal of improving their stability and accuracy in PDE benchmarks under a uniform training setup. We began by examining their initialization properties and proposed a Glorot-like initialization scheme that is basis-agnostic (as demonstrated by its application to sine-based KANs within the RGA KAN architecture). Preliminary results on function fitting and PDE tasks showed that this initialization alone significantly improved training outcomes compared to the default initialization, in some cases by several orders of magnitude. For certain benchmarks such as Burgers' equation, the proposed initialization was sufficient to train deeper models successfully, whereas for others, namely Allen--Cahn, it fell short. This observation made it clear that initialization alone was not enough to fully address the depth-scaling issue.

Motivated by this, and inspired by PirateNets, we analyzed the properties of Chebyshev KANs at initialization and observed strong parallels to standard MLPs. This insight led to the design of the RGA KAN architecture. This architecture proved effective in overcoming divergence for deeper networks, with increased depth and parameter count leading to improved accuracy rather than degradation. Through the IB perspective, we linked this desirable behavior to the model's ability to traverse all three characteristic phases of training -- fitting, diffusion, and diffusion equilibrium -- unlike baseline cPIKANs that tend to stall prematurely. Equipped with the proposed initialization and architecture, we then established a fixed training pipeline with adaptive components and compared RGA KANs against parameter-matched cPIKANs (also using the new initialization) and PirateNets, which are widely considered state of the art for many PDE benchmarks. Importantly, our aim was not to finely tune hyperparameters on a per-task basis but to test the performance of the proposed design in a uniform setting. Across all PDE benchmarks, RGA KANs outperformed both baselines, often by large margins, and remained stable in cases where the others diverged. The ablation studies further clarified the relative contribution of each adaptive method, but also revealed that in several cases the combination of the proposed initialization and architecture alone was sufficient to achieve good accuracy without divergence.

Regarding the architecture's specific configuration, we extensively tested four distinct initialization pairs for the gating parameters $\alpha$ and $\beta$, namely combinations of 0 and 1. While this sweep highlights how the network responds to different initial effective depths, it does not imply that a parametric search is strictly necessary for every new problem. In fact, our experiments suggest that initializing with mixed states -- either $(\alpha, \beta)=(1,0)$ or $(0,1)$ -- consistently yields superior performance compared to baseline architectures. The choice of 0 and 1 as initialization values carries physical significance: $(0,0)$ forces the network to adaptively grow its effective depth from a shallow state, while $(1,1)$ initializes the network at full depth, relying on the robustness of the proposed initialization scheme. It is important to clarify that while we initialize these parameters to discrete extremes, they are defined as continuous, trainable variables. During optimization, they naturally evolve into floating-point values.

While performance and stability are paramount, computational complexity remains a critical factor. Under fair parameter-matched settings, the time complexity of the RGA KAN is generally higher than that of PirateNets or cPIKANs. As observed in simpler benchmarks, this overhead stems from the simultaneous computation of multiple basis functions combined with the additional operations within the gating mechanisms. However, this dynamic changes as the problem complexity increases. As demonstrated in the Navier-Stokes benchmark, where the loss function involves multiple high-order spatial derivatives, the primary computational bottleneck shifts toward the backpropagation through the dense gating mechanisms. This explains why the training times of RGA KANs and PirateNets (both of which utilize gating) converge to similar values in this regime.

Naturally, this study also has limitations. To conserve computational resources, we deliberately avoided per-task hyperparameter tuning and considered networks of medium width at 16 neurons for the final PDE benchmarks. Our focus was not to set state-of-the-art results through exhaustive tuning, but to show that the proposed initialization and architecture provide a strong and robust foundation that already outperforms both baseline cPIKANs and PirateNets under default settings. A more extensive hyperparameter search would likely further improve performance. Another limitation is that all experiments relied on first-order optimization, specifically Adam. While this is standard in the PIML literature, recent work has shown that higher-order optimizers can yield remarkable improvements \cite{SSBroyden, SOAP}, and it would be valuable to explore the proposed architecture under such optimization regimes.

The proposed framework opens several avenues for future work. The initialization's basis-agnostic nature suggests potential applications to KAN variants using other bases, while the RGA KAN's transformer-like structure hints at relevance in broader domains. A particularly promising direction is the extension of this architecture to operator learning \cite{DeepONet}. While this paradigm focuses on learning mappings between function spaces rather than solving single instances, it shares the fundamental challenge of approximating PDE solutions. Given that KAN-based architectures have already shown promise in this field \cite{DeepOKAN, KANO, KKANs}, utilizing the proposed architecture in that context has potential. Finally, it would be interesting to test the architecture using KANs that are based on alternative representations of the Kolmogorov--Arnold theorem, such as ActNets \cite{ActNet} or KKANs \cite{KKANs}, to assess whether the benefits of residual gating and adaptive training extend beyond the formulation used in this work.

\newpage

\appendix

\section{Adaptive Training Methods}  \label{app:adaptive}

Regardless of the backbone architecture, the loss function minimized via gradient descent in PIML is generally more complex than in conventional neural networks, as it involves not only the neural network's output but also its gradients of various orders with respect to different input variables. This leads to a highly intricate loss landscape, where reaching minima requires not only expressive architectures but also effective adaptive training techniques. In this work, we port four such techniques which have been extensively utilized in PINNs: one focuses on the selection of collocation points for loss evaluation, while the other three introduce modifications to the composite loss function itself.

\subsection{Collocation Points Resampling} \label{app:adaptive1}

The selection of collocation points used to enforce the PDE can be performed in a one-time manner, where a fixed set of training points is used throughout the entire training process, either with or without mini-batching. However, periodically resampling the collocation points during training can serve as a regularization technique, improving the network's ability to generalize to spatiotemporal regions that were not explicitly sampled. Moreover, if the resampling process is not purely random but instead adaptive -- e.g., guided by residuals or other heuristics -- it has been shown to significantly enhance the final accuracy of the trained network \cite{COLLOCS1, COLLOCS2, COLLOCS3}.

For this study, we adopt the residual-based adaptive distribution (RAD) technique \cite{RAD}. In particular, an initial dense pool of $N_\text{pool}$ collocation points, $\left\{\left(t_\text{pool}^i,\mathbf{x}_\text{pool}^i\right)\right\}_{i=1}^{N_\text{pool}}$, is generated from a uniform grid, and training is performed on a dynamically selected subset of $N_\text{pde} \ll N_\text{pool}$ points. These points are periodically resampled from the pool according to the probability density function

\begin{equation}
	p\left(t_\text{pool},\mathbf{x}_\text{pool}\right) =  \frac{\left \lVert \mathcal{R}_\text{pde}\left[u\left(t_\text{pool},\mathbf{x}_\text{pool};\boldsymbol{\theta}\right)\right] \right \rVert_2^\delta}{\frac{1}{N_\text{pool}}\sum_{i=1}^{N_\text{pool}}\left \lVert \mathcal{R}_\text{pde}\left[u\left(t_\text{pool}^i,\mathbf{x}_\text{pool}^i;\boldsymbol{\theta}\right)\right] \right \rVert_2^\delta} + C,  \label{eq18}
\end{equation}

\noindent where $\delta \geq 0$, $C \geq 0$ are hyperparameters of the method. This resampling strategy directs the network’s training toward regions where the PDE residuals are larger, which can be particularly beneficial in scenarios involving discontinuities or sharp gradients in the PDE solution.

\subsection{Global Loss Weighting} \label{app:adaptive2}

When minimizing a loss function composed of multiple terms, a common challenge arises from the fact that different terms may converge at different rates. As a result, selecting appropriate weights for each term is a necessity. For instance, in L2 regularization, the choice of the regularization coefficient significantly impacts training: an excessively large value can impede learning, while a very small value may render the regularization effect negligible. A similar issue occurs in PIML, where there is often a bias toward minimizing certain terms of Eq. \eqref{eq6} while neglecting others.

Among the numerous proposed ways to address this issue \cite{WangNTK, KKANs, AAHWeights}, we choose the learning-rate annealing (LRA) algorithm introduced in \cite{Wang1}. During training, the weight adjustment is guided by the computation

\begin{equation}
	\hat{\lambda}_\xi = \frac{||\nabla_{\boldsymbol{\theta}} \mathcal{L}_\text{pde}\left(\boldsymbol{\theta}\right)||_2 + ||\nabla_{\boldsymbol{\theta}} \mathcal{L}_\text{ic}\left(\boldsymbol{\theta}\right)||_2 + ||\nabla_{\boldsymbol{\theta} }\mathcal{L}_\text{bc}\left(\boldsymbol{\theta}\right)||_2}{||\nabla_{\boldsymbol{\theta}} \mathcal{L}_\xi\left(\boldsymbol{\theta}\right)||_2}, \label{eq19}
\end{equation}

\noindent $\xi$ represents either ``pde'', ``ic'' or ``bc''. The loss weights of Eq. \eqref{eq6} are then updated according to the rule

\begin{equation}
	\lambda_\xi^{\text{new}} = a \lambda_\xi^{\text{old}} + \left(1-a\right)\hat{\lambda}_\xi, \label{eq20}
\end{equation}

\noindent where $a$ is the method's hyperparameter. The initial values are set to $\lambda_\text{pde} = \lambda_\text{ic} = \lambda_\text{bc} = 1$ to ensure equal weighting at the start of training, although in cases where domain knowledge is available, more suitable initializations can be selected. The update of Eq. \eqref{eq20} is performed periodically at a predetermined interval.

\subsection{Causal Training} \label{app:adaptive3}

A common challenge in training neural networks to solve time-dependent PDEs is the violation of causality. Since collocation points are sampled from the entire temporal domain, the network may unintentionally minimize residuals associated with future states before adequately minimizing those corresponding to past states. To mitigate this issue, following \cite{Causal}, we partition the temporal domain into $M$ sequential segments of equal length and introduce temporal weights $\left\{w_i\right\}_{i=1}^{M}$, modifying Eq. \eqref{eq7} as

\begin{equation}
	\mathcal{L}_\text{pde}\left(\boldsymbol{\theta}\right) = \frac{1}{M}\sum_{i=1}^{M}{w_i\mathcal{L}_\text{pde}^i\left(\boldsymbol{\theta}\right)} , \label{eq21}
\end{equation}

\noindent where $\mathcal{L}_\text{pde}^i\left(\boldsymbol{\theta}\right)$ represents the PDE loss computed over collocation points whose temporal coordinates fall within the $i$-th segment. The temporal weights are updated at each training iteration (epoch) according to

\begin{equation}
	w_i = \exp\left(-\epsilon \sum_{j=1}^{i-1}\mathcal{L}_\text{pde}^j\left(\boldsymbol{\theta}\right)\right) , \label{eq22}
\end{equation}

\noindent where $\epsilon \geq 0$ is a hyperparameter controlling the influence of the cumulative loss from the first $i-1$  segments on the weight assigned to the $i$-th segment. In practice, this enforces a time-ordered minimization of the PDE residuals, ensuring that earlier time steps are prioritized before the network attempts to minimize residuals in future states.

\subsection{Residual-Based Attention} \label{app:adaptive4}

The LRA algorithm introduced in Section \ref{app:adaptive2} is essentially a global loss weighting scheme, while causal training corresponds to a batch-wise weighting strategy, where groups of collocation points share the same weights. However, point-wise multipliers have also demonstrated significant success in PIML \cite{SA} and have been integral to works achieving state-of-the-art results \cite{KKANs, BRDR}. In such methods, the individual loss terms in Eqs. \eqref{eq7}-\eqref{eq9} are modified as

\begin{equation}
	\mathcal{L}_\xi\left(\boldsymbol{\theta}\right) = \frac{1}{N_\xi}\sum_{i=1}^{N_\xi}\left \lVert \alpha_{\xi}^{i} \mathcal{R}_\xi\left[u\left(t_\xi^i,\mathbf{x}_\xi^i;\boldsymbol{\theta}\right)\right] \right \rVert_2^2, \label{eq23}
\end{equation}

\noindent where $\xi$ represents either ``pde'', ``ic'' or ``bc'' and $\alpha_\xi^i$ is the local weight assigned to the $i$-th collocation point.

For this study, we adopt the residual-based attention (RBA) method introduced in \cite{RBA}, where all local weights are initially set to 1 and updated at each training iteration according to

\begin{equation}
	\alpha_\xi^{i~\text{(new)}} = \gamma \, \alpha_\xi^{i~\text{(old)}} + \eta \,\frac{\left \lVert \mathcal{R}_\xi\left[u\left(t_\xi^i,\mathbf{x}_\xi^i;\boldsymbol{\theta}\right)\right]\right \rVert_2}{\max_j\left(\left\{\left \lVert \mathcal{R}_\xi\left[u\left(t_\xi^j,\mathbf{x}_\xi^j;\boldsymbol{\theta}\right)\right]\right \rVert_2\right\}_{j=1}^{N_\xi}\right)} \label{eq24}
\end{equation}

\noindent where $\gamma \geq 0$, $\eta \geq 0$ are hyperparameters of the method.

In theory, while RBA is not inherently incompatible with causal training -- as the batch-wise weights can be computed using the modified loss of Eq. \eqref{eq23} -- the same does not hold for RAD as presented in \ref{app:adaptive1}. Since RAD involves resampling collocation points, it removes the one-to-one correspondence between them and their local weights. In order to apply these methods in conjunction, we adopt the following strategy: instead of assigning local weights solely to the currently selected $N_\text{pde}$ training points, we define the RBA weights over the entire pool, i.e., $\alpha_\text{pool}^i = 1$ for $i = 1, \dots, N_\text{pool}$ at initialization. These weights are updated at each training iteration according to Eq. \eqref{eq24}, using the residuals evaluated at the corresponding points. Then, during each application of RAD, the probability density function in Eq. \eqref{eq18} is modified by incorporating the RBA weights as multiplicative factors on the residuals:

\begin{equation}
	p\left(t_\text{pool},\mathbf{x}_\text{pool}\right) =  \frac{\left \lVert \alpha_\text{pool} \mathcal{R}_\text{pde}\left[u\left(t_\text{pool},\mathbf{x}_\text{pool};\boldsymbol{\theta}\right)\right] \right \rVert_2^\delta}{\frac{1}{N_\text{pool}}\sum_{i=1}^{N_\text{pool}}\left \lVert \alpha_\text{pool}^i \mathcal{R}_\text{pde}\left[u\left(t_\text{pool}^i,\mathbf{x}_\text{pool}^i;\boldsymbol{\theta}\right)\right] \right \rVert_2^\delta} + C,  \label{eq25}
\end{equation}

\noindent This modification effectively biases the sampling process toward regions where the weighted residuals are larger, allowing RAD to leverage the localized attention mechanism introduced by RBA. As a result, the method retains the adaptive sampling benefits of RAD while amplifying its focus on regions deemed important by RBA.

\section{Detailed Derivations} \label{app:derivs}

\subsection{Proposed Initialization Scheme} \label{app:derivs1}

Consider a single KAN layer with outputs given by Eq. \eqref{eq26}:

\begin{equation}
	y_j \;=\; \sum_{i=1}^{d_\text{I}} \sum_{m=1}^{D} Z^{(j)}_{im},
	\label{eqA1}
\end{equation}

\noindent where all biases have been set to zero at initialization and we have defined

\begin{equation}
	Z^{(j)}_{im} = w_{jim}\,B_m(x_i).
	\label{eqA2}
\end{equation}

\noindent Since weights are independent of inputs, expectations factor as products. Therefore

\begin{equation}
	\mathbb{E}\left[Z^{(j)}_{im}\right]
	= \mathbb{E}\,\left[w_{jim}B_m(x_i)\right]
	= \mathbb{E}\,\left[w_{jim}\right]\,\mathbb{E}\,\left[B_m(x_i)\right]
	= 0
	\label{eqA3}
\end{equation}

\noindent holds. In addition, considering two pairs of indices $\left(i,m\right) \neq \left(i^\prime,m^\prime\right)$, we find

\begin{align}
	\mathbb{E}\,\left[Z^{(j)}_{im}\,Z^{(j)}_{i^\prime m^\prime}\right]
	&= \mathbb{E}\,\left[w_{jim}w_{ji^\prime m^\prime}\,B_m\left(x_i\right)\,B_{m^\prime}\left(x_{i^\prime}\right) \right] \nonumber\\
	&= \mathbb{E}\,\left[w_{jim}w_{ji^\prime m^\prime}\right]\;
	\mathbb{E}\,\left[B_m\left(x_i\right)\,B_{m^\prime}\left(x_{i^\prime}\right)\right] \nonumber\\
	&= \mathbb{E}\,[w_{jim}]\,\mathbb{E}\,[w_{ji^\prime m^\prime}]\;
	\mathbb{E}\,\left[B_m\left(x_i\right)\,B_{m^\prime}\left(x_{i^\prime}\right)\right]  \nonumber\\
	&= 0,
	\label{eqA4}
\end{align}

\noindent where we have taken into consideration that distinct weights are independent and zero-mean. Therefore, for $(i,m)\neq(i^\prime,m^\prime)$, we arrive at

\begin{align}
	\mathrm{Cov}\,\left(Z^{(j)}_{im},Z^{(j)}_{i^\prime m^\prime}\right)
	&= \mathbb{E}\,\left[Z^{(j)}_{im}Z^{(j)}_{i^\prime m^\prime}\right]
	- \mathbb{E}\left[Z^{(j)}_{im}\right]\,\mathbb{E}\left[Z^{(j)}_{i^\prime m^\prime}\right] \nonumber\\
	&= 0 - 0\cdot 0 \;=\; 0. \label{eqA5}
\end{align}

\noindent Using this result, the variance of the sum in Eq. \eqref{eqA1} reduces to a sum of individual variances,

\begin{equation}
	\mathrm{Var}\left(y_j\right) \;=\; \sum_{i=1}^{d_\text{I}}\sum_{m=1}^{D}\mathrm{Var}\left(Z^{(j)}_{im}\right)
	\;=\; \sum_{i=1}^{d_\text{I}}\sum_{m=1}^{D}\mathrm{Var}\left(w_{jim}B_m\left(x_i\right)\right).
	\label{eqA6}
\end{equation}

\noindent For a single term, the independence of $w_{jim}$ and $x_i$ together with $\mathbb{E}[w_{jim}]=0$ leads to

\begin{align}
	\mathrm{Var}\left[w_{jim} \, B_m\left(x_i\right)\right] &=  \mathbb{E}^2\left[B_m\left(x_i\right)\right]\mathrm{Var}\left(w_{jim}\right) + \mathrm{Var}\left(w_{jim}\right)\mathrm{Var}\left[B_m\left(x_i\right)\right] \nonumber \\
	&= \sigma_m^2 ~\mathbb{E}\left[B_m\left(x_i\right)^2\right]. \label{eqA7}
\end{align}

\noindent Therefore,

\begin{equation}
	\mathrm{Var}(y_j) \;=\; d_\text{I}\sum_{m=1}^{D}\sigma_m^2\,\mu_m^{(0)},
	\label{eqA8}
\end{equation}
and enforcing $\mathrm{Var}(y_j)=\mathrm{Var}(x_i)=1$ leads directly to Eq. \eqref{eq28} of the main text.

As far as the backward pass is concerned, differentiating Eq. \eqref{eqA1} with respect to $x_i$ yields

\begin{equation}
	\frac{\partial y_j}{\partial x_i} \;=\; \sum_{m=1}^{D} w_{jim}\,B_m^\prime\left(x_i\right).
	\label{eqA9}
\end{equation}

\noindent The loss gradient with respect to $x_i$ then becomes

\begin{equation}
	\delta x_i \;=\; \sum_{j=1}^{d_\text{O}} \frac{\partial y_j}{\partial x_i}\,\delta y_j
	= \sum_{j=1}^{d_\text{O}} \sum_{m=1}^{D} w_{jim}B_m^\prime(x_i)\,\delta y_j.
	\label{eqA10}
\end{equation}

\noindent Following the same reasoning as for the forward pass, distinct $(j,m)$ pairs are uncorrelated, so

\begin{equation}
	\mathrm{Var}\left(\delta x_i\right) \;=\; \sum_{j=1}^{d_\text{O}}\sum_{m=1}^{D}
	\mathrm{Var}\left(w_{jim}B_m^\prime\left(x_i\right)\,\delta y_j\right).
	\label{eqA11}
\end{equation}

\noindent Each summand can be evaluated as

\begin{equation}
	\mathrm{Var}\left(w_{jim}B_m^\prime\left(x_i\right)\,\delta y_j\right)
	= \sigma_m^2\,\mathbb{E}[B_m^\prime(x_i)^2]\,\mathbb{E}[\delta y_j^2]
	= \sigma_m^2\,\mu_m^{(1)}\,\mathrm{Var}(\delta y_j).
	\label{eqA12}
\end{equation}

\noindent Thus,

\begin{equation}
	\mathrm{Var}(\delta x_i) \;=\; d_\text{O}\,\mathrm{Var}(\delta y_j)\sum_{m=1}^{D}\sigma_m^2\,\mu_m^{(1)},
	\label{eqA13}
\end{equation}

\noindent and imposing $\mathrm{Var}(\delta x_i)=\mathrm{Var}(\delta y_j)$ yields Eq. \eqref{eq30}.

\subsection{Chebyshev-based KAN Derivative} \label{app:derivs2}

Following \cite{PirateNets}, we consider small activations at initialization and adopt the linear-regime approximation, where

\begin{equation}
	\tanh x \approx x, 
	\qquad 
	\frac{d}{dx}\tanh x \approx 1.
	\label{eqΑ14}
\end{equation}

\noindent In this regime, and recalling from Eq. \eqref{eq16} that $B_m(x) = T_m\left(\tanh x\right)$, we expand the Chebyshev polynomial $T_m(x)$ around zero and retain only the linear term $\mathcal{O}(x)$. The basis functions then simplify to

\begin{equation}
	B_m(x) \approx T_m(0) + T_m^{\prime}(0)\,x + \mathcal{O}(x^2).
	\label{eqA15}
\end{equation}

\noindent Differentiating Eq. \eqref{eqA15} yields

\begin{equation}
	B_m^{\prime}(x) \approx T_m^{\prime}(0) = m\,U_{m-1}(0),
	\label{eqA16}
\end{equation}

\noindent where $U_n\left(\cdot\right)$ denotes the $n$-th Chebyshev polynomial of the second kind and we have used the identity $T_m^{\prime}(z) = m\,U_{m-1}(z)$. Since

\begin{equation}
U_n(0) = \sin\!\left(\frac{(n+1)\pi}{2}\right),
\label{eqA17}
\end{equation}

\noindent it follows that

\begin{equation}
	U_{m-1}(0) = \sin\!\left(\frac{m\pi}{2}\right) =
	\begin{cases}
		0, & m~\text{even},\\[4pt]
		(-1)^{\frac{m-1}{2}}, & m~\text{odd}.
	\end{cases}
	\label{eqA18}
\end{equation}

\noindent Substituting Eq. \eqref{eqA18} into Eq. \eqref{eqA16} gives

\begin{equation}
	B_m^{\prime}(x) \approx 
	\begin{cases}
		0, & m~\text{even},\\[4pt]
		m\,(-1)^{\frac{m-1}{2}}, & m~\text{odd}.
	\end{cases}
	\label{eqA19}
\end{equation}

\noindent Therefore, substituting Eq. \eqref{eqA19} into Eq. \eqref{eq36} from the main text, and using Eq. \eqref{eq38}, we find

\begin{align}
	\frac{\partial u_j^{(l)}(x;\boldsymbol{\theta})}{\partial x}
	\; &\approx \;
	\sum_{i=1}^{d_{l-1}} \frac{\partial u_i^{(l-1)}(x;\boldsymbol{\theta})}{\partial x} \sum_{m\,\text{odd}}^{D}
	m\,(-1)^{\frac{m-1}{2}}\,w_{jim}^{(l)} \nonumber \\
	&= \sum_{i=1}^{d_{l-1}}
	\tilde{w}_{ji}^{(l)}\,
	\frac{\partial u_i^{(l-1)}(x;\boldsymbol{\theta})}{\partial x}.
	\label{eqA20}
\end{align}

\section{Implementation Details} \label{app:impl}

All neural network architectures utilized in this study are implemented in JAX \cite{jax} using the \texttt{jaxKAN} framework \cite{jaxKAN} and trained at the highest precision settings on an NVIDIA GeForce RTX 4090 GPU. Their performance is assessed in terms of the L$^2$ error of the predicted solution, $\mathbf{u}_\text{pred}$, relative to a reference solution, $\mathbf{u}_\text{ref}$, i.e.,

\begin{equation}
	\mathcal{E} = \frac{\left \lVert \mathbf{u}_\text{pred} - \mathbf{u}_\text{ref}\right \rVert_2}{\left \lVert \mathbf{u}_\text{ref}\right \rVert_2}, \label{eqB1}
\end{equation}

\noindent where $\mathcal{E}$ and $\lVert \cdot \rVert_2$ denote the relative $L^2$ error and $L^2$ norm, respectively. In the following, we provide a breakdown of the training settings and hyperparameter configurations used for each family of experiments presented in the main text.

\subsection{Small-Scale Benchmarks} \label{app:impl1}

For the function fitting benchmarks of Section \ref{sec3.2.1}, we generate $4\cdot 10^3$ input–output samples uniformly over the domain $[-1,1]^d$, with $d$ being the dimensionality of the function. Networks are trained for $2\cdot 10^3$ iterations to minimize the $L^2$ loss using the Adam \cite{adam} optimizer with a constant learning rate of $10^{-3}$ in full-batch mode. For the final relative $L^2$ error evaluation, we use 1,000 uniformly spaced points in $[-1,1]$ for the one-dimensional task, a $200\times200$ grid for the two-dimensional tasks, a $30^3$ grid for the three-dimensional Hartmann function, and a $10^5$ grid for the five-dimensional Sobol g-function.

As far as the small-scale PDE benchmarks of Section \ref{sec3.2.2} are concerned, $N_\text{pde} = 2^{12}$ collocation points are used to enforce the differential operator for each PDE. In the Burgers' case, $N_\text{bc} = 2^6$ points are used for each of the two boundary conditions, together with $N_\text{ic} = 2^6$ points for the initial condition. For Helmholtz, $N_\text{bc} = 2^6$ points are used for each of the four boundary conditions. Collocation points are sampled once from a uniform grid and remain fixed throughout training. Training is performed for $5\cdot10^3$ iterations using the Adam optimizer with a constant learning rate of $10^{-3}$ in full-batch mode.

\subsection{Depth-Scaling Experiments} \label{app:impl2}

For the depth-scaling experiments of Section \ref{sec3.3}, we utilize all adaptive methods described in \ref{app:adaptive}. In particular, adaptive collocation-point resampling is performed with hyperparameters $\delta = 1$ and $C = 1$, which is a standard choice in the literature \cite{RAD}. The pool of collocation points to enforce the PDE is constructed using $N_\text{pool} = 400\times 400$ uniformly distributed samples over the spatiotemporal domain $[0,1]\times[-1,1]$. Every $2\cdot 10^3$ iterations, $N_\text{pde} = 2^{12}$ points are resampled for training. Alongside this, we apply the RBA method with hyperparameters $\gamma = 0.999$ and $\eta = 0.01$ as in \cite{RBA}, so that resampling follows the probability density function defined in Eq. \eqref{eq25}. For the initial condition, a fixed set of $N_\text{ic} = 2^6$ points is used, while RBA weighting is still applied. No collocation points are used to enforce boundary conditions, which are instead incorporated directly into the network architecture \cite{ExactBCs}.

In addition, every $10^3$ iterations we apply LRA according to Eqs. \eqref{eq19}-\eqref{eq20}, with decay parameter $a=0.9$. Causal training is also employed by partitioning the temporal domain into $M=32$ segments and using $\epsilon=1.0$ \cite{PirateNets}. Models are trained for $10^5$ iterations using the Adam optimizer with an initial warm-up phase of $10^3$ iterations, reaching a learning rate of $10^{-3}$, followed by exponential decay with a factor of $0.9$ every $2\cdot10^3$ iterations.

\subsection{Final Experimental Results} \label{app:impl3}

For the extensive comparative benchmarks presented in Section \ref{sec5}, the results presented in Sections \ref{sec5.1} -- \ref{sec5.7} correspond to the exact same configuration as detailed in \ref{app:impl2}. For the (2+1)-dimensional heat equation of Section \ref{sec5.8}, the pool of collocation points is constructed using $N_\text{pool} = 50 \times 50 \times 50$ uniformly distributed samples over the spatiotemporal domain, while all other settings remain unchanged. Finally, regarding the Navier-Stokes equation of Section \ref{sec5.9}, the pool of collocation points consists of 32 uniformly distributed points in the temporal domain and a $64 \times 64$ grid in the spatial domain. Furthermore, due to the problem's sensitivity to the initial condition, we increase the number of collocation points used to enforce the initial condition to $N_\text{ic} = 2^{12}$. All other training settings match those described previously.

\section{Function Fitting Benchmarks} \label{app:ff}

In this appendix we provide the analytic definitions of the benchmark functions used in the function-fitting experiments of Section \ref{sec3.2.1}. Each function is defined on the hypercube $[-1,1]^d$, where $d$ denotes the input dimensionality.

\paragraph{One-dimensional oscillatory function}

\begin{equation}
	f_1(x) \;=\; \sin\!\left(2 \pi x\right) \;+\; 3x .
	\label{eqC1}
\end{equation}

\smallskip

\paragraph{Two-dimensional product function}  

\begin{equation}
	f_2(x_1, x_2) \;=\; x_1 \, x_2 .
	\label{eqC2}
\end{equation}

\smallskip

\paragraph{Two-dimensional Bessel-based function}

\begin{equation}
	f_3(x_1, x_2) \;=\; I_1(x_1) \;+\; \exp\!\big(I_1^{(e)}(x_2)\big) \;+\; \sin(x_1 x_2) ,
	\label{eqC3}
\end{equation}

\noindent where $I_1(\cdot)$ denotes the modified Bessel function of the first kind of order~1, and $I_1^{(e)}(\cdot)$ its exponentially scaled version.

\smallskip

\paragraph{Three-dimensional Hartmann function}

\begin{equation}
	f_4(x_1, x_2, x_3) \;=\; - \sum_{k=1}^4 \alpha_k \,
	\exp\!\left(- \sum_{j=1}^3 A_{kj} \,\big(x_j - P_{kj}\big)^2 \right),
	\label{eqC4}
\end{equation}

\noindent where

\begin{equation*}
	\alpha = (1.0,\, 1.2,\, 3.0,\, 3.2),
\end{equation*}

\noindent and 

\begin{equation*}
	A =
	\begin{bmatrix}
		3 & 10 & 30 \\
		0.1 & 10 & 35 \\
		3 & 10 & 30 \\
		0.1 & 10 & 35
	\end{bmatrix}, \quad P = 10^{-4}\!
	\begin{bmatrix}
		3689 & 1170 & 2673 \\
		4699 & 4387 & 7470 \\
		1091 & 8732 & 5547 \\
		381 & 5743 & 8828
	\end{bmatrix}.
\end{equation*}

\smallskip

\paragraph{Five-dimensional Sobol g-function}

\begin{equation}
	f_5(x_1,\dots,x_5) \;=\; \prod_{j=1}^5 \frac{|\,4x_j - 2\,| + a_j}{1 + a_j},
	\label{eqC5}
\end{equation}

\noindent where $a_j = \tfrac{j-2}{2}$ for $j=1,\dots,5$.

\section{Studied Partial Differential Equations} \label{app:pdes}

In this appendix, we present the PDEs studied throughout this work, including their governing equations, boundary and/or initial conditions, and corresponding reference solutions. The equations are listed in the same order in which they appear in Section \ref{sec5} of the main text.

\subsection{Allen--Cahn Equation} \label{app:pdes1}

The (1+1)-dimensional Allen--Cahn equation on the spatiotemporal domain $t \in [0,1]$, $x \in [-1,1]$ is given by

\begin{equation}
	\frac{\partial u}{\partial t} \;-\; D \frac{\partial^2 u}{\partial x^2} \;=\; 5\,(u - u^3),
	\label{eqD1}
\end{equation}

\noindent where $D$ represents the diffusion coefficient. This parameter controls the width of the interfacial transition layers between the stable phases ($u=\pm 1$); specifically, smaller values of $D$ lead to sharper, stiffer interfaces. In this study, we consider the stiff regime with $D = 10^{-4}$. The problem is subject to the initial condition

\begin{equation}
	u(0,x) \;=\; x^2 \cos(\pi x),
	\label{eqD2}
\end{equation}

\noindent and periodic boundary conditions

\begin{equation}
	u(t,-1) \;=\; u(t,1), \qquad 
	\frac{\partial u}{\partial x}(t,-1) \;=\; \frac{\partial u}{\partial x}(t,1).
	\label{eqD3}
\end{equation}

\noindent The reference solution shown in Figure \ref{fig9} of the main text corresponds to the data used in \cite{PirateNets} and accessed from the paper's accompanying GitHub repository \cite{jaxpi}.

\subsection{Burgers' Equation} \label{app:pdes2}

The (1+1)-dimensional viscous Burgers' equation on the spatiotemporal domain $t \in [0,1]$, $x \in [-1,1]$ is given by

\begin{equation}
	\frac{\partial u}{\partial t} + u \frac{\partial u}{\partial x} \;=\; \nu \frac{\partial^2 u}{\partial x^2},
	\label{eqD4}
\end{equation}

\noindent where $\nu$ represents the kinematic viscosity. This parameter governs the balance between nonlinear convection (which tends to steepen gradients into shocks) and diffusion (which smooths them out); lower values of $\nu$ result in sharper shock fronts. In this benchmark, we use $\nu = 1/(100\pi)$. It is considered with initial condition

\begin{equation}
	u(0,x) \;=\; -\sin(\pi x),
	\label{eqD5}
\end{equation}

\noindent and homogeneous Dirichlet boundary conditions

\begin{equation}
	u(t,-1) \;=\; u(t,1) \;=\; 0.
	\label{eqD6}
\end{equation}

\noindent The reference solution shown in Figure \ref{fig10} of the main text corresponds to the data used in \cite{PirateNets} and accessed from the paper's accompanying GitHub repository \cite{jaxpi}.

\subsection{Korteweg--De Vries Equation} \label{app:pdes3}

The (1+1)-dimensional Korteweg--De Vries equation on the spatiotemporal domain $t \in [0,1]$, $x \in [-1,1]$ is given by

\begin{equation}
	\frac{\partial u}{\partial t} \;+\; u\, \frac{\partial u}{\partial x} \;+\; \lambda^2\, \frac{\partial^3 u}{\partial x^3} \;=\; 0,
	\label{eqD7}
\end{equation}

\noindent where $\lambda$ governs the dispersion of the system. This parameter balances the nonlinear convective steepening ($u u_x$) against dispersive spreading ($u_{xxx}$). We consider the low-dispersion regime with $\lambda = 0.022$; low values of $\lambda$ allow nonlinearity to dominate initially, causing the wave to steepen into a train of narrow, high-frequency solitons rather than spreading out smoothly. The problem is considered with initial condition

\begin{equation}
	u(0,x) \;=\; \cos(\pi x),
	\label{eqD8}
\end{equation}

\noindent and periodic boundary conditions

\begin{equation}
	u(t,-1) \;=\; u(t,1).
	\label{eqD9}
\end{equation}

\noindent The reference solution shown in Figure \ref{fig11} of the main text corresponds to the data used in \cite{PirateNets} and accessed from the paper's accompanying GitHub repository \cite{jaxpi}.

\subsection{Sine Gordon Equation} \label{app:pdes4}

The (1+1)-dimensional Sine Gordon equation on the spatiotemporal domain $t \in [0,1]$, $x \in [0,1]$ is given by

\begin{equation}
	\frac{\partial^2 u}{\partial t^2} \;-\; \frac{\partial^2 u}{\partial x^2} \;+\; \sin u \;=\; 0.
	\label{eqD10}
\end{equation}

\noindent This nonlinear hyperbolic PDE is fundamental in relativistic field theory and the modeling of Josephson junctions. We consider the standard dimensionless form, where the characteristic velocity and field mass are normalized to unity. The problem is initialized with

\begin{equation}
	u(0,x) \;=\; \sin(\pi x),
	\label{eqD11}
\end{equation}

\noindent and subject to homogeneous Dirichlet boundary conditions

\begin{equation}
	u(t,0) \;=\; u(t,1) \;=\; 0.
	\label{eqD12}
\end{equation}

\noindent The analytical solution of this equation, depicted in Figure \ref{fig12} of the main text, is given by

\begin{equation}
	u\left(t,x\right) \;=\; \frac{1}{2}\left[\sin\left(\pi\left(x+t\right)\right) + \sin\left(\pi\left(x-t\right)\right)\right].
	\label{eqD13}
\end{equation}

\subsection{Advection Equation} \label{app:pdes5}

The (1+1)-dimensional linear advection equation on the spatiotemporal domain $t \in [0,1]$, $x \in [0,2\pi]$ is given by

\begin{equation}
	\frac{\partial u}{\partial t} \;+\; c\,\frac{\partial u}{\partial x} \;=\; 0,
	\label{eqD14}
\end{equation}

\noindent where $c$ represents the advection velocity, determining the speed at which the initial wave profile travels across the domain. We consider a high-velocity regime with $c = 20$; such high convection speeds typically pose a challenge for physics-informed learning \cite{failmodes}. The problem is initialized with

\begin{equation}
	u(0,x) \;=\; \sin x,
	\label{eqD15}
\end{equation}

\noindent and subject to periodic boundary conditions

\begin{equation}
	u(t,x) \;=\; u(t,x+2\pi).
	\label{eqD16}
\end{equation}

\noindent The analytical solution of this equation, depicted in Figure \ref{fig13} of the main text, is given by

\begin{equation}
	u\left(t,x\right) \;=\; \sin\left(\text{mod}\left(x - c t, 2\pi\right)\right).
	\label{eqD17}
\end{equation}

\subsection{Helmholtz Equation} \label{app:pdes6}

The 2-dimensional Helmholtz equation on the spatial domain $x \in [-1,1]$, $y \in [-1,1]$ is given by

\begin{equation}
	\frac{\partial^2 u}{\partial x^2} \;+\; \frac{\partial^2 u}{\partial y^2} \;+\; k^2 u \;=\; \left[ k^2 - \pi^2 \left(a_1^2 + a_2^2\right) \right] \sin\!\left(a_1 \pi x\right) \sin\!\left(a_2 \pi y\right),
	\label{eqD18}
\end{equation}

\noindent where $k=1$ is the wave number and $a_1, a_2$ represent the mode frequencies along the $x$ and $y$ axes, respectively. These parameters control the spatial oscillation density of the solution. The problem is considered with homogeneous Dirichlet boundary conditions

\begin{equation}
	u(-1,y) \;=\; u(1,y) \;=\; u(x,-1) \;=\; u(x,1) \;=\; 0.
	\label{eqD19}
\end{equation}

\noindent The analytical solution, depicted for $a_1 = 1$ and $a_2 = 4$ in Figure \ref{fig14} of the main text, is given by

\begin{equation}
	u(x,y) \;=\; \sin\!\left(a_1 \pi x\right)\sin\!\left(a_2 \pi y\right).
	\label{eqD20}
\end{equation}

\subsection{Poisson Equation} \label{app:pdes7}

The 2-dimensional Poisson equation on the spatial domain $x \in [-1,1]$, $y \in [-1,1]$ is given by

\begin{equation}
	\frac{\partial^2 u}{\partial x^2} \;+\; \frac{\partial^2 u}{\partial y^2} \;=\;  - 2\pi^2\omega^2 \sin\!\left(\omega \pi x\right) \sin\!\left(\omega \pi y\right),
	\label{eqD21}
\end{equation}

\noindent where $\omega$ governs the spatial frequency of the source term and the solution. Higher values of $\omega$ introduce more rapid oscillations, therefore increasing this parameter allows us to test the network's ability to resolve high-frequency features. We consider three distinct regimes of increasing complexity: $\omega \in \{1, 2, 4\}$. The problem is subject to homogeneous Dirichlet boundary conditions

\begin{equation}
	u(-1,y) \;=\; u(1,y) \;=\; u(x,-1) \;=\; u(x,1) \;=\; 0.
	\label{eqD22}
\end{equation}

\noindent The analytical solution is given by

\begin{equation}
	u(x,y) \;=\; \sin\!\left(\omega \pi x\right)\sin\!\left(\omega \pi y\right),
	\label{eqD23}
\end{equation}

\noindent and is depicted in Figure \ref{fig15} of the main text.

\subsection{Heat Equation} \label{app:pdes8}

The (2+1)-dimensional multi-scale heat equation on the spatiotemporal domain $t \in [0, 1]$, $(x, y) \in [0, 1]^2$ is given by

\begin{equation}
	\frac{\partial u}{\partial t} \;-\; D_x \frac{\partial^2 u}{\partial x^2} \;-\; D_y \frac{\partial^2 u}{\partial y^2} \;=\; 0,
	\label{eqD24}
\end{equation}

\noindent where $D_x$ and $D_y$ are the diffusion coefficients along the spatial axes. These parameters govern the rate at which temperature gradients are smoothed out in their respective directions; a larger coefficient leads to rapid dissipation, while a smaller one results in the persistence of high-frequency features. In this benchmark, we introduce extreme anisotropy by setting $D_x = (500\pi)^{-2}$ and $D_y = \pi^{-2}$. The significant disparity $D_y \gg D_x$ creates a stiff multi-scale problem where dynamics evolve at radically different rates along the $x$ and $y$ axes. The problem is subject to the initial condition

\begin{equation}
	u(0, x, y) \;=\; \sin(20\pi x) \sin(\pi y), \quad \text{for } (x,y) \in \Omega,
	\label{eqD25}
\end{equation}

\noindent and homogeneous Dirichlet boundary conditions on the boundary of the unit square $\partial \Omega$

\begin{equation}
	u(t, x, y) \;=\; 0, \quad \text{for } (x,y) \in \partial \Omega.
	\label{eqD26}
\end{equation}

\noindent The analytical solution, snapshots of which are depicted in Figure \ref{fig16} of the main text, is given by

\begin{equation}
	u(t, x, y) \;=\; \sin(20\pi x) \sin(\pi y) \exp\left(-\frac{626}{625}t\right).
	\label{eqD27}
\end{equation}

\subsection{Navier-Stokes Equation} \label{app:pdes9}

The (2+1)-dimensional incompressible Navier-Stokes equations in the velocity-vorticity formulation on the toroidal domain $\Omega = [0, 2\pi]^2$ for $t \in [0, 0.1]$ are given by

\begin{equation}
	\frac{\partial w}{\partial t} \;+\; \mathbf{u} \cdot \nabla w \;=\; \frac{1}{\text{Re}} \Delta w, \qquad \nabla \cdot \mathbf{u} \;=\; 0,
	\label{eqD28}
\end{equation}

\noindent where $\text{Re}$ is the Reynolds number, $\mathbf{u} = (u, v)$ is the velocity field and $w = \nabla \times \mathbf{u}$ is the vorticity. The Reynolds number characterizes the ratio of inertial forces to viscous forces. Lower values of $\text{Re}$ indicate a viscosity-dominated regime where flow features smooth out rapidly, whereas higher values allow for sustained turbulence and complex small-scale structures. We consider a Reynolds number $\text{Re} = 100$ and periodic boundary conditions in both spatial directions. Following \cite{Causal,ExpertsGuide}, the initial conditions are generated numerically from a random divergence-free velocity field. The reference solution is computed using a high-accuracy pseudo-spectral solver with Crank-Nicolson Runge-Kutta 4 (CN-RK4) time integration on a $64 \times 64$ grid (see Data Availability statement). Figure \ref{fig17} of the main text depicts the resulting reference fields ($u, v$, and $w$) at the final time $t=0.1$.

\section{Neural Tangent Kernel Analysis} \label{app:ntk}

The Neural Tangent Kernel (NTK) framework serves as an important tool to assess the stability of neural networks by examining the evolution of their gradient descent training dynamics \cite{JacotNTK}. The idea was subsequently extended to PINNs in order to investigate their convergence properties and spectral bias \cite{WangNTK}. Leveraging recent work that derived and analyzed the NTK for Chebyshev-based KANs and cPIKANs \cite{ChebNTK}, we utilize this theoretical framework to provide insight into the improved stability offered by our proposed initialization compared to the default scheme.

\subsection{Function Fitting} \label{app:ntk1}

\begin{figure}[b!]
	\centering
	\includegraphics[width=\linewidth]{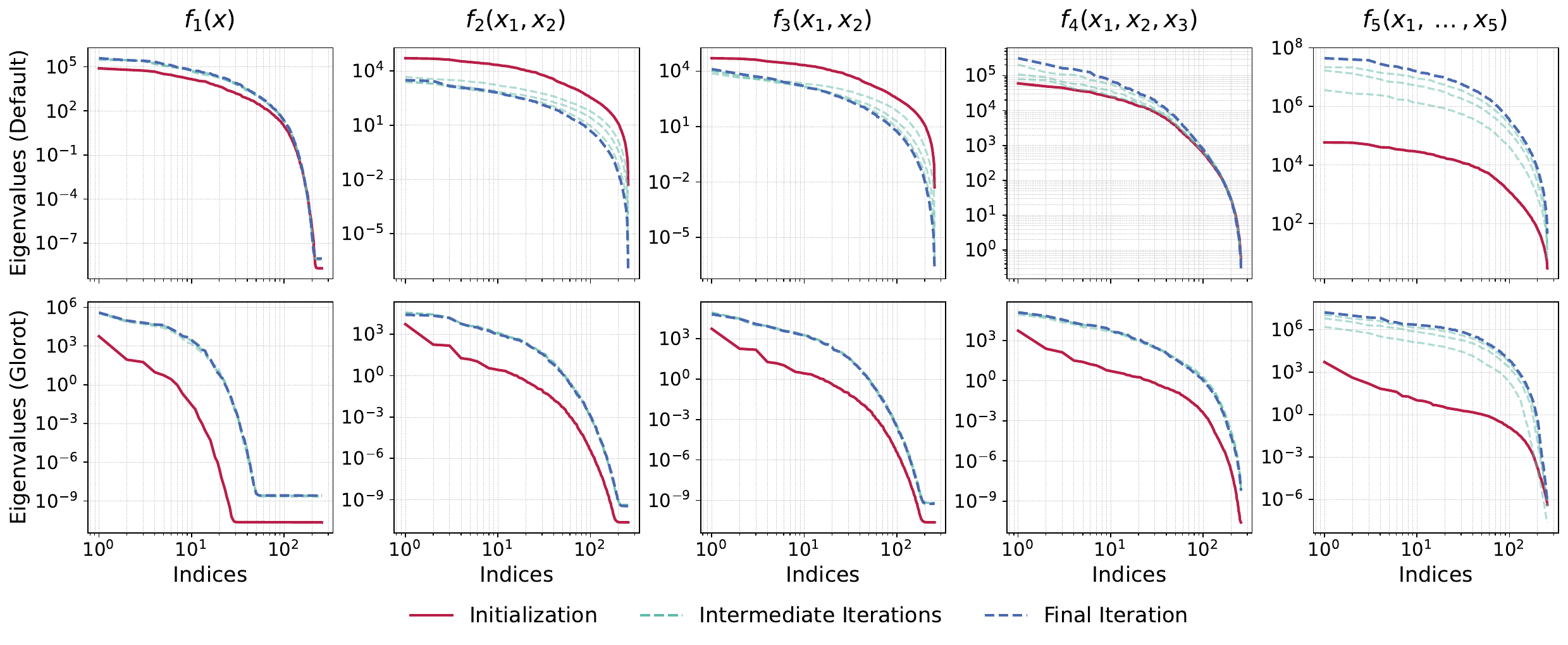}
	\caption{Evolution of the NTK eigenvalue spectra for the small architecture (width 4, depth 3) across the five benchmark functions. The top row displays the spectral evolution under the default initialization, while the bottom row shows the proposed Glorot-like initialization. The solid red line indicates the spectrum at initialization, the green dashed lines represent intermediate checkpoints (every 500 iterations) and the blue dashed line represents the final iteration.}
	\label{figntk1}
\end{figure}

\begin{figure}[b!]
	\centering
	\includegraphics[width=\linewidth]{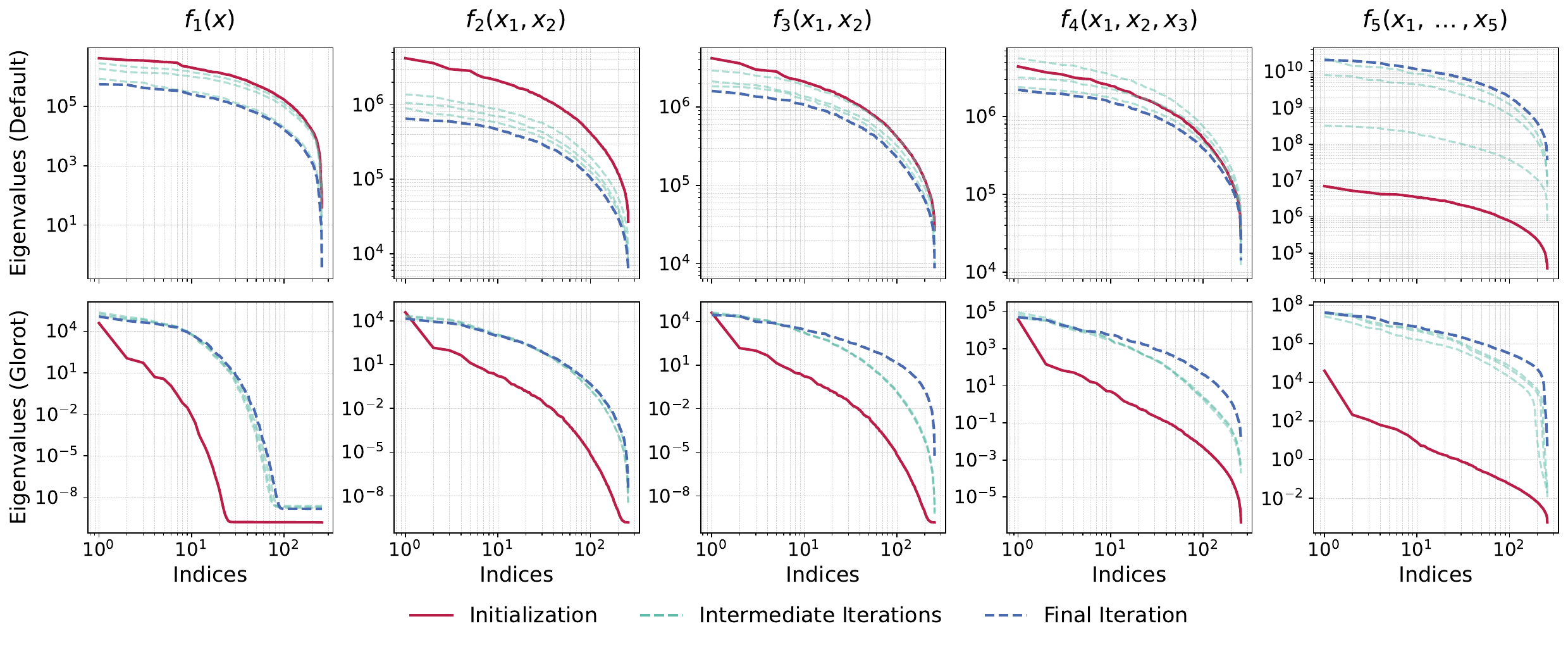}
	\caption{Evolution of the NTK eigenvalue spectra for the big architecture (width 16, depth 5) across the five benchmark functions. The top row displays the spectral evolution under the default initialization, while the bottom row shows the proposed Glorot-like initialization. The solid red line indicates the spectrum at initialization, the green dashed lines represent intermediate checkpoints (every 500 iterations) and the blue dashed line represents the final iteration.}
	\label{figntk2}
\end{figure}

For our NTK-centered analysis, we first use the experimental settings from the function fitting benchmarks in Section \ref{sec3.2.1}. To present the most competitive comparison, we select the training runs where the default initialization achieved its lowest final loss. For these runs, we extract the NTK spectra at initialization and subsequently every 500 iterations until the end of training. The spectra are computed using a randomly selected subset of $2^8$ points from the training data. The evolution of the NTK eigenvalues is visualized in Figure \ref{figntk1} for the small architecture and in Figure \ref{figntk2} for the large architecture. In both figures, the top row displays the spectra under the default initialization, while the bottom row corresponds to the proposed Glorot-like scheme.

Across both small and large architectures, the proposed initialization demonstrates superior spectral properties compared to the default scheme. With the exception of $f_5$, where the benchmark's complexity limits the impact of initialization for the studied networks, a consistent pattern emerges: spectra under the default initialization tend to systematically decrease or collapse during training. In contrast, the proposed scheme exhibits the opposite, favorable evolution, expanding to a robust profile. Notably, this stabilization often occurs rapidly; the significant overlap between intermediate and final spectral states (green and blue lines, respectively) indicates that the proposed initialization allows the network to quickly locate a stable optimization path. The distinction in stabilization between the two initialization schemes is particularly evident in the large architecture, where the default initialization yields spectra dominated almost exclusively by very large eigenvalues. Conversely, the NTK for the proposed initialization evolves into a characteristic power-law distribution that spans several orders of magnitude -- retaining high eigenvalues for dominant features while also accessing very low values near zero -- thereby allowing effective learning across multiple scales.

These observations align closely with the final relative $L^2$ error for these runs reported in Table \ref{tabntk1}. The correspondence is most striking in the large architecture configurations for functions $f_2$ and $f_4$: under the default initialization, the large networks fail to converge, yielding relative errors exceeding unity. This divergence is a direct consequence of the ill-conditioned spectra observed in the top row of Figure \ref{figntk2}, where the dominance of excessively large eigenvalues hinders effective gradient descent. In stark contrast, the proposed initialization not only prevents this divergence but achieves relative errors of $\mathcal{O}(10^{-2})$ or lower, confirming that the induced spectral stability is a prerequisite for accurate function approximation in deeper KAN architectures.

\begin{table}[h!]
	\centering
	\renewcommand{\arraystretch}{1.2}
	\setlength{\tabcolsep}{4pt}
	\small
	\begin{tabular}{c|cc||cc}
		\hline
		\multirow{2}{*}{\textbf{Function}} & \multicolumn{2}{c||}{\textbf{Width 4, Depth 3}} & \multicolumn{2}{c}{\textbf{Width 16, Depth 5}} \\
		& \textbf{Default} & \textbf{Proposed} & \textbf{Default} & \textbf{Proposed} \\
		\hline
		$f_1(x)$ & $5.98\times 10^{-2}$ & $\mathbf{6.45\times 10^{-3}}$ & $2.01\times 10^{-2}$ & $\mathbf{1.37\times 10^{-3}}$ \\
		$f_2(x_1,x_2)$ & $7.34\times 10^{-2}$ & $\mathbf{1.06\times 10^{-2}}$ & $1.29\times 10^{0}$  & $\mathbf{1.36\times 10^{-2}}$ \\
		$f_3(x_1,x_2)$ & $2.19\times 10^{-2}$ & $\mathbf{5.17\times 10^{-3}}$ & $5.75\times 10^{-1}$ & $\mathbf{4.06\times 10^{-3}}$ \\
		$f_4(x_1,x_2,x_3)$ & $9.99\times 10^{-1}$ & $\mathbf{5.17\times 10^{-2}}$ & $1.37\times 10^{0}$  & $\mathbf{4.17\times 10^{-2}}$ \\
		$f_5(x_1,\dots,x_5)$ & $9.47\times 10^{-1}$ & $\mathbf{9.08\times 10^{-1}}$ & $9.45\times 10^{-1}$ & $\mathbf{7.50\times 10^{-1}}$ \\
		\hline
	\end{tabular}
	\caption{Final relative $L^2$ errors for the representative training runs used in the NTK analysis for function fitting. The proposed initialization achieves the lowest error across all cases (highlighted in bold).}
	\label{tabntk1}
\end{table}

\subsection{Forward PDE Problems} \label{app:ntk2}

\begin{figure}[t!]
	\centering
	\includegraphics[width=\linewidth]{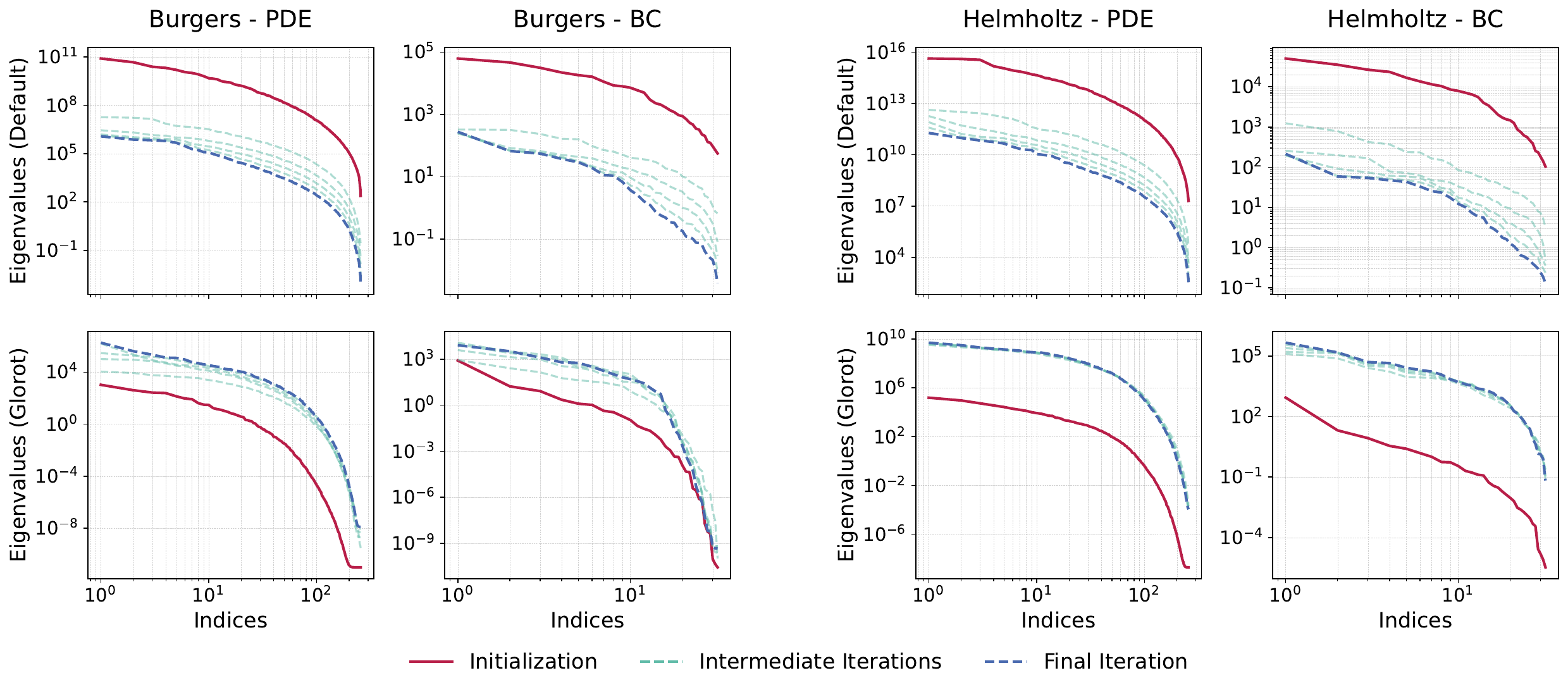}
	\caption{Evolution of the NTK eigenvalue spectra for the small architecture (width 4, depth 3) on PDE benchmarks. The top row displays the spectral evolution under the default initialization, while the bottom row shows the proposed Glorot-like initialization. The solid red line indicates the spectrum at initialization, the green dashed lines represent intermediate checkpoints (every 1000 iterations) and the blue dashed line represents the final iteration.}
	\label{figntk3}
\end{figure}

\begin{figure}[b!]
	\centering
	\includegraphics[width=\linewidth]{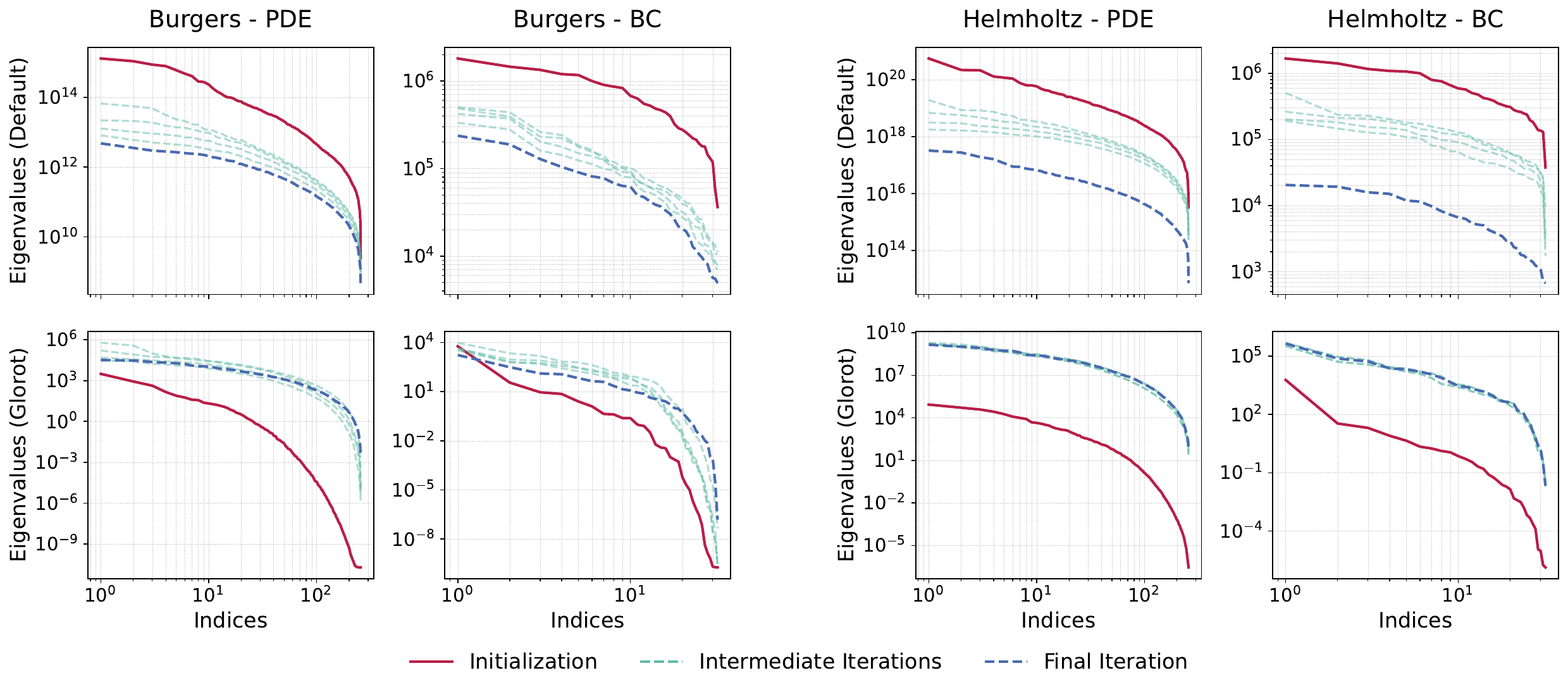}
	\caption{Evolution of the NTK eigenvalue spectra for the large architecture (width 16, depth 5) on PDE benchmarks. The top row displays the spectral evolution under the default initialization, while the bottom row shows the proposed Glorot-like initialization. The solid red line indicates the spectrum at initialization, the green dashed lines represent intermediate checkpoints (every 1000 iterations) and the blue dashed line represents the final iteration.}
	\label{figntk4}
\end{figure}

We continue our analysis with the PDE benchmarks of Section \ref{sec3.2.2}. Unlike the function fitting tasks, the loss function here is a composite of multiple terms, therefore we compute separate NTKs for the residuals associated with the PDE operator and the boundary conditions \cite{WangNTK}. As before, we analyze the training runs where the default initialization yielded the lowest final loss and extract the NTK spectra at initialization and subsequently every 1000 iterations until the end of training. The spectra are computed using a subset of $2^8$ points for the PDE term and $2^5$ points for the boundary condition term. Figure \ref{figntk3} and Figure \ref{figntk4} visualize the spectral evolution for the small (width 4, depth 3) and large (width 16, depth 5) architectures, respectively.

The results for the large architecture offer the most critical insight into the stability gap discussed in Section \ref{sec3.2.2}. Under the default initialization (top row), the eigenvalues at initialization are pathologically large. This is most extreme for the Helmholtz PDE term, where the maximum eigenvalue exceeds $10^{20}$. Such excessive magnitudes at initialization force the optimizer to drastically compress the spectrum to fit the target zero residuals, resulting in the observed spectral collapse (the blue line falls far below the red line). This behavior indicates that the network is primarily struggling to suppress its own initial scale rather than learning the solution structure. In contrast, the proposed initialization (bottom row) starts optimization in a much more favorable regime, with maximum eigenvalues typically between $10^2$ and $10^5$. Similar to the function fitting observations, the spectra under the proposed scheme tend to expand or stabilize rather than collapse. Furthermore, particularly for the Helmholtz equation, the proposed initialization leads to rapid convergence to a stable spectral state, as evidenced by the significant overlap between the intermediate (green dashed) and final (blue dashed) spectra.

Complementing the spectral analysis, Table \ref{tabntk2} presents the final relative $L^2$ errors for these representative runs. In every case, regardless of architecture size or PDE type, the difference between the two schemes is effectively the difference between divergence and convergence. Models trained under the default initialization systematically fail to learn, yielding relative errors near unity or higher, which aligns perfectly with the spectral collapse observed in the NTK spectra. Conversely, the proposed initialization consistently enables convergence, leading to lower relative $L^2$ errors.

\begin{table}[h!]
	\centering
	\renewcommand{\arraystretch}{1.2}
	\setlength{\tabcolsep}{4pt}
	\small
	\begin{tabular}{c|cc||cc}
		\hline
		\multirow{2}{*}{\textbf{PDE}} & \multicolumn{2}{c||}{\textbf{Width 4, Depth 3}} & \multicolumn{2}{c}{\textbf{Width 16, Depth 5}} \\
		& \textbf{Default} & \textbf{Proposed} & \textbf{Default} & \textbf{Proposed} \\
		\hline
		Burgers & $1.00\times 10^{0}$ & $\mathbf{4.51\times 10^{-1}}$ & $1.05\times 10^{0}$ & $\mathbf{3.57\times 10^{-1}}$ \\
		Helmholtz & $1.03\times 10^{0}$ & $\mathbf{4.77\times 10^{-1}}$ & $1.14\times 10^{0}$  & $\mathbf{1.21\times 10^{-1}}$ \\
		\hline
	\end{tabular}
	\caption{Final relative $L^2$ errors for the representative training runs used in the NTK analysis for forward PDEs. The proposed initialization is the only one achieving convergence across all cases (highlighted in bold).}
	\label{tabntk2}
\end{table}

\section{PirateNet Architecture} \label{app:pirate}

In this appendix, we provide a detailed description of the PirateNet architecture \cite{PirateNets} employed in the benchmarks of Section \ref{sec5}, together with an explicit parameter count. As in Section \ref{sec4.2}, we consider a single input sample $\mathbf{x} \in \mathbb{R}^{d_\text{I}}$, where $d_\text{I}$ is the number of input coordinates. If periodic boundary conditions are present, they are first enforced through the embedding of Eq. \eqref{eq39} in the main text, resulting in the transformed input $\tilde{\mathbf{x}} \in \mathbb{R}^{\tilde{d}_\text{I}}$.

The embedded coordinates are then passed through a RFF embedding layer. A trainable kernel $\mathbf{B} \in \mathbb{R}^{\tilde{d}_\text{I} \times 0.5 d_\text{H}}$ is initialized from a Gaussian distribution $\mathcal{N}(0, s^2)$ with $s > 0$ ($s = 1$ is used for the benchmarks presented in this work), and the embedding is defined as

\begin{equation}
	\Phi_j =
	\begin{bmatrix}
		\cos\left(\sum_{i=1}^{\tilde{d}_\text{I}} B_{ji} \tilde{x}_i\right) \\[4pt]
		\sin\left(\sum_{i=1}^{\tilde{d}_\text{I}} B_{ji} \tilde{x}_i\right)
	\end{bmatrix},
	\label{eqE1}
\end{equation}

\noindent where $\boldsymbol{\Phi} \in \mathbb{R}^{d_\text{H}}$. The resulting features are then processed through two MLP gates that generate the vectors $\mathbf{U}, \mathbf{V} \in \mathbb{R}^{d_\text{H}}$:

\begin{equation}
	U_j 
	= 
	\tanh\left(\sum_{i=1}^{d_\text{H}} w^u_{ji} \phi_i + b^u_j\right),
	\qquad
	V_j 
	= 
	\tanh\left(\sum_{i=1}^{d_\text{H}} w^v_{ji} \phi_i + b^v_j\right).
	\label{eqE2}
\end{equation}

\noindent All MLP layers, including the two of Eq. \eqref{eqE2}, follow the Random Weight Factorization (RWF) formulation \cite{RWF}, with weights initialized using the standard Glorot scheme \cite{Glorot} and biases initialized at zero.

The adaptive skip connection is introduced through $N$ identical blocks, each consisting of three MLP layers and a single gating parameter $\alpha$ (initialized at zero). Denoting the input to the $l$-th block by $\mathbf{x}^{(l)}$, with $\mathbf{x}^{(1)} = \boldsymbol{\Phi}$, the forward pass is given by

\begin{align}
	f_j^{(l)} &= \tanh\left(\sum_{i=1}^{d_\text{H}} w^{(l)}_{1,ji}\, x_i^{(l)} + b^{(l)}_{1,j}\right), \label{eqE3} \\
	z_{1,j}^{(l)} &= f_j^{(l)}\, U_j + \bigl(1 - f_j^{(l)}\bigr) V_j, \label{eqE4} \\
	g_j^{(l)} &= \tanh\left(\sum_{i=1}^{d_\text{H}} w^{(l)}_{2,ji}\, z_{1,i}^{(l)} + b^{(l)}_{2,j}\right), \label{eqE5}  \\
	z_{2,j}^{(l)} &= g_j^{(l)}\, U_j + \bigl(1 - g_j^{(l)}\bigr) V_j, \label{eqE6} \\
	h_j^{(l)} &= \tanh\left(\sum_{i=1}^{d_\text{H}} w^{(l)}_{3,ji}\, z_{2,i}^{(l)} + b^{(l)}_{3,j}\right), \label{eqE7}
\end{align}

\begin{align}
	x_j^{(l+1)} &= \alpha\, h_j^{(l)} + (1 - \alpha)\, x_j^{(l)}.
	\label{eqE8}
\end{align}

The output of the final PirateNet block, $\mathbf{x}^{(N+1)} \in \mathbb{R}^{d_\text{H}}$, is mapped to the network output through a linear layer

\begin{equation}
	u_j 
	= 
	\sum_{i=1}^{d_\text{H}}
	w^{o}_{ji}\,
	x^{(N+1)}_i,
	\label{eqE9}
\end{equation}

\noindent where $\mathbf{u} \in \mathbb{R}^{d_\text{O}}$. This final layer is initialized using the same physics-informed least-squares procedure described in Section \ref{sec4.2}, but since this is a standard linear transformation, no re-indexing is required. If non-periodic boundary conditions are present, they are directly enforced at this stage by multiplying the network output with suitable boundary-shaping functions.

The total number of trainable parameters of the above architecture is

\begin{align}
	\left|\boldsymbol{\theta}\right|
	&= \underbrace{0.5\,d_\text{H}\tilde{d}_\text{I}}_{\text{RFF Embeddings}}
	+ \overbrace{2d_\text{H}\bigl(d_\text{H}+2\bigr)}^{\text{MLP Gates}}
	+ \underbrace{N\bigl[3d_\text{H}\bigl(d_\text{H}+2\bigr) + 1\bigr]}_{\text{PirateNet Blocks}}
	+ \overbrace{d_\text{O}d_\text{H}}^{\text{Output Layer}} \nonumber \\[4pt]
	&= d_\text{H}\left[0.5\,\tilde{d}_\text{I} + d_\text{O} + \bigl(d_\text{H}+2\bigr)(3N+2)\right] + N.
	\label{eqE10}
\end{align}

\noindent Note that, due to the RWF formulation of the layers, each MLP block with input dimension $n_\text{in}$ and output dimension $n_\text{out}$ contains $n_\text{out}(n_\text{in}+2)$ trainable parameters, rather than $n_\text{out}(n_\text{in}+1)$ as in standard MLPs.

\section*{CRediT authorship contribution statement}

\textbf{Spyros Rigas:} Conceptualization, Data curation, Formal analysis, Investigation, Methodology, Software, Validation, Visualization, Writing – original draft, Writing - Review \& Editing. \textbf{Fotios Anagnostopoulos:} Data curation, Visualization, Writing – original draft. \textbf{Michalis Papachristou:} Data curation, Visualization, Writing – original draft. \textbf{Georgios Alexandridis:} Project administration, Supervision. 

\section*{Declaration of competing interest}

The authors declare that they have no known competing financial interests or personal relationships that could have appeared to influence the work reported in this paper.

\section*{Data Availability}

All data and source code used to produce the experimental results reported in this work are openly accessible at https://github.com/srigas/RGA-KANs. %This repository contains all necessary scripts, configurations, and instructions to reproduce the experiments and figures presented in the paper.

%All results presented in this paper were generated using openly available code, which can be accessed at [GitHub repository link]. No additional data were used in this study.

\iffalse
\section*{Acknowledgments}

This research did not receive any specific grant from funding agencies in the public, commercial, or not-for-profit sectors.
\fi

\bibliography{references}

\end{document}